\documentclass[a4paper]{article}
\usepackage{amsmath}
\usepackage{setspace}

\usepackage{accents,amssymb,amsthm,color,enumerate,fancyhdr,geometry,hyperref}
\allowdisplaybreaks[4]
\usepackage[all,pdf]{xy}
\usepackage{pifont}
\usepackage[perpage,symbol*]{footmisc}
\usepackage{cases}
\usepackage{graphicx}
\usepackage{subcaption}
\usepackage{appendix}
\usepackage{bbm}
\usepackage{bm}
\usepackage{lineno} 
\usepackage{changepage}

\allowdisplaybreaks[4]

\usepackage[T1]{fontenc}
\usepackage{authblk}
\usepackage{natbib}
\setcitestyle{authoryear,round}

\usepackage{amsfonts}
\usepackage{float}
\usepackage{abstract}

\usepackage{microtype}
\usepackage{booktabs} % for professional tables
\usepackage{multirow}
\usepackage{makecell}
\usepackage[figuresright]{rotating}
\usepackage{array}
\usepackage[export]{adjustbox}
\usepackage{anyfontsize}
\usepackage{enumitem}
\usepackage{algorithm}
\usepackage{algorithmic}
\usepackage{setspace}

\usepackage{hyperref}
\hypersetup{hypertex = true,
colorlinks = true,
linkcolor = blue,
anchorcolor = blue,
citecolor = blue}

\fontsize{12pt}{\baselineskip}\selectfont

\title{\textbf{Simultaneous Identification of  Sparse Structures and Communities in Heterogeneous Graphical Models}}
\author[1]{Dapeng Shi}
\author[1]{Tiandong Wang}
\author[2]{Zhiliang Ying}

\affil[1]{Shanghai Center for Mathematical Sciences, Fudan University}
\affil[2]{Department of Statistics, Columbia University}

\date{}

\newcommand{\Pen}{\textbf{Pen}}

\newcommand{\prox}{\operatorname{prox}}
\newcommand{\Se}{S^+}
\newcommand{\maT}{\mathcal{T}}
\newcommand{\maV}{\mathcal{V}}
\newcommand{\bt}{\boldsymbol{t}}
\newcommand{\argmin}{\operatorname*{argmin}} % define the argmin operator
\newcommand{\argmax}{\operatorname*{argmax}} % define the argmin operator

\newtheorem{remark}{Remark}
\newtheorem{assumption}{Assumption}
\newtheorem{lemma}{Lemma}
\newtheorem{theorem}{Theorem}
\newtheorem{corollary}{Corollary}
\newtheorem{proposition}{Proposition}

\newcommand\keywords[1]{\textbf{Keywords}: #1}

\begin{document}
%\pagewiselinenumbers	

\begin{spacing}{1.5}%%行间距调整
\maketitle

\begin{abstract}	
     Exploring and detecting community structures hold significant importance in genetics, social sciences, neuroscience, and finance. Especially in graphical models, community detection can encourage the exploration of sets of variables with group-like properties. In this paper, within the framework of Gaussian graphical models, we introduce a novel decomposition of the underlying graphical structure into a sparse part and low-rank diagonal blocks (non-overlapped communities). 
	%Specifically, when the primary factors influencing the overall variables are eliminated, we hypothesize that the graph structure of the remaining residue is composed of non-overlapped communities (diagonal-block and low-rank) and sparse structures. 
    We illustrate the significance of this decomposition through two modeling perspectives and propose a three-stage estimation procedure with a fast and efficient algorithm for the identification of the sparse structure and communities. Also on the theoretical front, we establish conditions for local identifiability and extend the traditional irrepresentability condition to an adaptive form by constructing an effective norm, which ensures the consistency of model selection for the adaptive $\ell_1$ penalized estimator in the second stage. Moreover, we also provide the clustering error bound for the $K$-means procedure in the third stage. Extensive numerical experiments are conducted to demonstrate the superiority of the proposed method over existing approaches in estimating graph structures. Furthermore, we apply our method to the stock return data, revealing its capability to accurately identify non-overlapped community structures.
\end{abstract}

\section{Introduction}

Probabilistic graphical models are widely applied in various domains such as gene regulatory networks \citep{li2012sparse,fan2009}, social networks \citep{guo2015estimating, tan2014learning}, brain connectivity networks \citep{li2018nonparametric, pircalabelu2020community, eisenach2020high}, and business networks \citep{zhao2023identifiability}. In graphical models, nodes represent random variables, while edges denote conditional independence/dependence among different random variables. 
In the current study, we focus on the Gaussian graphical models and delve into the graph structure decomposition method to identify sparse structures and non-overlapped communities.

\par The problem of structure estimation in the Gaussian graphical models has been extensively investigated in the literature under the assumption of sparsity \citep{friedman2007,yuan2007model,tan2014learning, Ravikumar2011,meinshausen2006high, fan2009}. However, recent attention has shifted towards addressing the estimation challenges posed by heterogeneous graph structures. Two primary types of heterogeneity characterize these structures. The first arises from the hierarchical structure of the population, leading to partially different graph structures in distinct subpopulations \citep{guo2011joint, danaher2014joint, hao2018simultaneous}. Additionally, graph structures may depend on individual characteristics, leading to subject-specific graphical models. To address such heterogeneity, decomposition procedures of graph structures have been introduced, wherein each substructure represents the subgraph structure generated by different covariates \citep{zhang2022high, wang2023, niu2023}.

\par 
Another form of heterogeneity arises from the community attributes of nodes. Nodes within the same community tend to be densely connected, whereas nodes in different communities exhibit sparse connections. Consequently, the groups or node communities demonstrate more homogeneity within than between communities. Community detection, a well-explored task in network data analysis \citep{holland1983stochastic, amini2013pseudo,amini2018semidefinite}, focuses on identifying these groups of nodes. However, a fundamental distinction between probabilistic graphical models and complex network models lies in the fact that, in the latter, the graph structure is predetermined at the outset of the analysis. In contrast, our objective here is to estimate the graph structure based on the data.
 
 \par Consider node heterogeneity, and we utilize graph structure decomposition methods to identify communities and characterize the remaining sparse structures that contain information both between and within communities. When hub nodes with large degrees emerge within a community, \cite{tan2014learning} and \cite{tarzanagh2018estimation} proposed to use the $\ell_2$ group Lasso penalty \citep{yuan2006model} and the $\ell_1$ penalty \citep{tibshirani1996regression} to identify hubs and the remaining sparse structure, respectively. It is noteworthy that the sparse structure may also provide important insights into the underlying model structure. For example, identifying the remaining sparse structure proves to be beneficial for distinguishing subtypes when analyzing the differential gene expressions \citep{danaher2014joint,hao2018simultaneous}.

\par 
To the best of our knowledge, there has been limited exploration to simultaneously detect communities with comparable degrees and identify sparse structures through the decomposition of graph structures. The majority of the existing literature concentrates on estimating the overall graph structure without leveraging the additional information available through decomposition. For example, \cite{tan2015cluster} proposed a two-stage method that initially clusters nodes and then assumes the overall graph structure to be diagonal-block. \cite{wilms2022tree} introduced a tree-based node-aggregation method that encodes side information to construct relationships within communities, even in scenarios where overlapping communities are allowed. These approaches, however, only facilitate the estimation of the subgraph structure corresponding to each community but overlook the relationship between communities. In addition, \cite{pircalabelu2020community} conducted community detection by constructing a new pseudo-covariance matrix through the perturbation of community information to the observed covariance matrix. \cite{eisenach2020high} employed the factor model to incorporate community information and developed inference tools for the graph structure of the factors. Moreover, \cite{ma2021inter} and \cite{yuan2014partial} explored partial graphical models in a similar spirit. Although these methods consider sparse connections between communities, they are ineffective in extracting information within communities. 

\par In this paper, we introduce a three-stage estimation {procedure} aiming at simultaneously identifying both sparse structures and communities by properly decomposing the graph structure. Our key contributions are summarized as follows:
\begin{enumerate}
    \item We present a novel decomposition of the overall graph structure. Specifically, we decompose the graph structure of the regression residue into two components. One component consists of low-rank diagonal blocks, capturing the non-overlapped community structure. The other component is sparse, revealing the underlying
    connectivity within and between communities. %We also give two modeling perspectives of this decomposition for a comprehensive understanding.
    \item {We also provide a three-stage estimation procedure with efficient algorithms to recover the sparse structure and communities. Especially after removing globally influential factors, we propose an adaptive $\ell_1$ penalized estimation method to obtain a low-rank matrix with certain diagonal-block sparsity, aiming to capture the initial structure of the communities. Furthermore, we apply the $K$-means clustering method based on the resulting estimator to obtain specific membership labels. Besides, we develop a fast and efficient algorithm to implement the proposed graph decomposition procedure.}
    \item To ensure the identifiability of the sparse and community parts, we extend the classical low-rank manifold and its corresponding tangent space to accommodate {low-rank diagonal blocks}. Building upon this extension, we introduce a practical norm that generalizes the traditional irrepresentability condition and establish the model selection consistency {for the adaptive $\ell_1$ penalized estimation in the second stage.  We further establish a clustering error bound for  the $K$-means procedure in the third stage.}
    \item Finally, we apply the proposed method to the synthetic and real datasets. The results further validate that the proposed method can simultaneously identify both the communities and the sparse structures with good interpretation. Meanwhile, in terms of uncovering sparse structures, the proposed method performs favorably against
     some well-established methods. 
\end{enumerate}
%These contributions collectively establish the proposed method as a robust and effective approach for the simultaneous exploration of sparse structures and non-overlapped communities within the graph structure decomposition paradigm.
\par The remainder of the paper is organized as follows. Section \ref{Sec.GraphModel} gives a general framework for the decomposition of graph structure, which consists of two modeling strategies. One is inspired by the stochastic block model widely used in network analysis, while the other arises when there exist grouped latent variables. Section~\ref{Sec.ProMethod} gives a detailed statement of the proposed adaptive $\ell_1$ penalized estimation method, including the efficient algorithm as well as weights assignment and choice of tuning parameters. We then develop theoretical results on identifiability and asymptotic properties in Subsections \ref{Sec.Identi} and \ref{Sec.Asymp}, respectively. In Sections \ref{Sec.Simulation} and \ref{Sec.Realdata}, we illustrate the proposed framework on several synthetic and real datasets. Concluding remarks are collected in Section \ref{Sec.Conclusion}.

\subsection{Notation} We first introduce some notation to be used throughout the rest. Let $\mathbb{R}^p$ be the $p$-dimensional vector space, and $\mathbb{R}^{p\times q}$, $\mathbb{S}^{p\times p}$ be the sets of $p\times q$ matrices and symmetric $p\times p$ matrices, respectively. For any positive integer $m$, we let $[m]\triangleq\{1, \ldots, m\}$. For two nonnegative sequences $a_n$ and $b_n$, $a_n \lesssim b_n$ means there exist a constant $c>0$ and an integer $n_0>0$ such that $a_n \leq c b_n$ for all $n\ge n_0$, while $a_n \asymp b_n$ represents there exist a constant $d>1$ and an integer $n_1>0$ such that $\frac{1}{d}b_n\leq a_n\leq d b_n$ for all $n\ge n_1$.  Also, for two sequences of nonnegative random variables  $x_n, y_n$, $x_n \lesssim_P y_n$ means there exists a constant $c>0$ such that $\mathbb{P}\left(a_n \leq c b_n\right) \rightarrow 1$ as $n\to \infty$. 
\par For a matrix $A=\left(a_{i j}\right)_{p \times q}$, the sub-matrix corresponding to some rectangle set $S$ is denoted by $A_{S}\triangleq\left(a_{ij}\right)_{(i,j) \in S}$. The support of $A$ is defined as $\operatorname{Supp}(A) = \{(i,j)|a_{ij}\not= 0\}$ and the rank of $A$ is $\operatorname{rank}(A)$. We let $\mathbf{0}_{p\times q}$ be the matrix of all zeros in $\mathbb{R}^{p\times q}$. We also denote the following norms for a matrix $A$:
\begin{align*}
\|A\|_0 &\triangleq \#\{(i,j)|a_{ij}\not= 0 \},\qquad \|A\|_{\text {off},1 } \triangleq\sum_{i \neq j}\left|a_{i j}\right|,\\
\|A\|_{\infty }
&\triangleq\max _{i j}\left|a_{i j}\right|,\qquad \|A\|_F \triangleq \sqrt{\sum_{ij}|a_{ij}|^2}.
\end{align*}
We use $v(A) \in \mathbb{R}^{p^2}$ to denote the vectorization of $A$ and $\operatorname{diag}(A)\in\mathbb{R}^{p\times p}$ to denote the diagonal matrix of $A\in\mathbb{R}^{p\times p}$. We use $A\succeq 0$ if $A$ is positive semidefinite and $A\succ 0$ if $A$ is positive definite. In addition, for $A\succeq 0$, we let $\|A\|_2$ and $\|A\|_*$ be its respective spectral norm and nuclear norm.
\par Finally, for a linear space $\maT$, let $\maT^\perp$ be the orthogonal complement space of $\maT$, and $\maT_1\oplus \maT_2$ be the orthogonal direct sum space of two linear spaces $\maT_1, \maT_2$.

\section{Graphical models with community-based decomposition}
\label{Sec.GraphModel}
We now introduce a decomposition for the underlying graph structure, aiming at revealing some community structure with a focus on the case of non-overlapped communities. We start by assuming that the observed $p$-dimensional variables $\boldsymbol{X} = (X_1, \cdots, X_p)^T$ and $q$-dimensional covariates $\boldsymbol{C} = (C_1, \cdots, C_q)^T$ follow the regression model:
\begin{equation}
	\begin{aligned}
		\boldsymbol{X}|\boldsymbol{C} &= B^T\boldsymbol{C} + \boldsymbol{R}, \quad \boldsymbol{R} \sim  N(0, \Sigma),		
	\end{aligned}
\end{equation}
where $B\in\mathbb{R}^{q\times p}$ represents the regression coefficient matrix. Each component in $\boldsymbol{X}$ is associated with a node in an undirected graph $G(E, V)$, where $V=\{1,\cdots, p\}$ denotes the set of nodes, and $E = \left\{\left\{a,b\right\}|a\not= b \in V\right\}$ the set of undirected edges. Let $\widetilde{\Theta} \triangleq (B\operatorname{Var}(\boldsymbol{C})B^T +\Sigma)^{-1} = (\tilde{\theta}_{ij})_{p\times p}$ be the precision matrix of $\boldsymbol{X}$, with $\tilde{\theta}_{ij} = 0$ equivalent to $\{i,j\}\notin E$ under the Gaussian assumption for both $\boldsymbol{C}$ and $\boldsymbol{R} $ \citep{lauritzen1996graphical}.   Although the precision matrix $\widetilde{\Theta}$ generally does not exhibit community structures, the precision matrix of the residual
$\boldsymbol{R}$ may show some community structures after properly
accounting for the covariates with global effects. In particular, we decompose the precision matrix $\Theta \triangleq \Sigma^{-1}$ of $\boldsymbol{R}$ as:
\begin{equation}\label{Model.decomposition}
	\Theta = L + S, \quad  L\succeq 0 \ \ \ (\text{or }- L\succeq 0),
\end{equation}
where $L\in\mathbb{S}^{p\times p}$ consists of low-rank diagonal blocks up to row or column permutations to capture the non-overlapped  communities,  and 
$S\in\mathbb{S}^{p\times p}$ is sparse (detailed description of the assumption see Assumption \ref{A.non-overlapped} and \ref{A.transver}).   Notice that when there is only one block, such decomposition is the same as the setup in \cite{chandrasekaran2010latent}. Such sparse plus low-rank decomposition can be locally identifiable under mild conditions, which has been carefully studied \citep{chandrasekaran2010SIAM,candes2011robust}. Additionally, it is worth noting that we do not assume that $S$ has the sparse structure with the same diagonal blocks as $L$. This implies that within the off-diagonal blocks of $L$, which are entirely composed of zeros, $S$ does not necessarily have to be zero. In what follows, we illustrate the significance of such a decomposition procedure from two different perspectives.

\subsection{ Latent community graphical models}\label{Sec.GMLC}

The first perspective is motivated by the stochastic block model (SBM) in network analysis \citep{amini2018semidefinite,li2021convex,lei2015consistency}.  Assume there exist $m$ non-overlapped latent communities and for each node $i$, we assume it belongs to the $k$-th latent community, denoted by $\boldsymbol{a}_i := (0, \cdots, a_{ik},\cdots,0)^T\in\mathbb{R}^r$. Here, we relax the constraint $a_{ik}\in \{0,1\}$ in SBM to $a_{ik}\in\mathbb{R}$, representing the effect of the $k$-th community imposed on the node $i$. We also allow the effect of the $k$-th community to have more than one dimension, i.e., $\boldsymbol{a}_i = (0, \cdots, a_{i1}, \cdots, a_{ir_{k}},\cdots,0)^T$. For nodes $i$ and $j$ belonging to different communities, it is satisfied that $\operatorname{Supp}(\boldsymbol{a}_i)\cap \operatorname{Supp}(\boldsymbol{a}_j) = \emptyset$, and $\sum_{k=1}^mr_k = r$. Hence, we model the precision matrix $\Theta = (\theta_{ij})_{p\times p}$ of $\boldsymbol{R}$ as:
\begin{equation}\label{Model.GMLC}
\theta_{ij} = (\boldsymbol{a}_i)^T\boldsymbol{a}_j + s_{ij},
\end{equation}
where the appearance of one edge in $E$ mainly depends on two effects after removing the effects of $\boldsymbol{C}$: one comes from the community effect $(\boldsymbol{a}_i)^T\boldsymbol{a}_j$, and the other from the remaining sparse effect $s_{ij}$. Denote $A = (\boldsymbol{a}_1^T, \cdots, \boldsymbol{a}_p^T)^T$ and let $L = AA^T$, then model \eqref{Model.GMLC} is exactly the model \eqref{Model.decomposition} with $L\succeq 0$, and we refer to such models as latent  community graphical models. 
\subsection{Grouped latent variable graphical models} \label{Sec.GMGLV}
The second perspective is an extension of the classical latent variable graphical model \citep{chandrasekaran2010latent}. Suppose that there exist $m$ grouped latent variables (unobserved variables)  $\boldsymbol{Z}_1 = (Z_{1,1},\cdots, Z_{1,r_1}  )^T\in\mathbb{R}^{r_1}, \cdots, \boldsymbol{Z}_m =(Z_{m,1},\cdots, Z_{m,r_m})^T\in\mathbb{R}^{r_m}$ such that $(\boldsymbol{X}^T, \boldsymbol{Z}_1^T,\cdots, \boldsymbol{Z}_m^T)^T|\boldsymbol{C} = B_{(OH)}^T\boldsymbol{C} +\boldsymbol{R}_{(OH)}, \boldsymbol{R}_{(OH)} \sim N(0, \Theta_{(OH)}^{-1}) $, where $\Theta_{(OH)}$ is the precision matrix of the joint errors. We in addition let the submatrices $\Theta_O, \Theta_{O, H}, \Theta_H$ specify the dependency among the observed variables, between the observed and latent variables, and among the latent variables, respectively. In addition, $B_{(OH)}\in\mathbb{R}^{q\times(p+r)}$ denotes the joint regression coefficients of $(\boldsymbol{X}^T, \boldsymbol{Z}_1^T,\cdots, \boldsymbol{Z}_m^T)^T$ on $\boldsymbol{C}$. Then the regression for the observed data $\boldsymbol{X}$ conditional on $\boldsymbol{C}$ is 
\begin{equation}\label{Model.GLGM}
	\boldsymbol{X}|\boldsymbol{C}  = B_O^T\boldsymbol{C} + \boldsymbol{R}_O, \quad \boldsymbol{R}_O \sim N(0, \Theta^{-1}),\quad 
    \Theta = \Theta_O - \Theta_{O,H}(\Theta_H)^{-1}\Theta_{O,H}^T,  
\end{equation} 
where $B_O\in\mathbb{R}^{q\times p}$ is the first $p$ columns of $B_{(OH)}$ associated with the regression coefficients of observed variables $\boldsymbol{X}$ on $\boldsymbol{C}$, and $\boldsymbol{R}_O$ is the first $p$ components of the error term $\boldsymbol{R}_{(OH)}$.  Now we assume $\boldsymbol{X}$ admits the partition $\{\boldsymbol{X}_{G_1}, \cdots, \boldsymbol{X}_{G_m} \}$ such that  
\begin{equation}\label{A.Condidepen}
	(\boldsymbol{X}_{G_i}, \boldsymbol{Z}_i)\perp \boldsymbol{Z}_{-i}|(\boldsymbol{C}, \boldsymbol{X}_{-G_i}), \forall i\in [m] ,
\end{equation}
 where $\boldsymbol{Z}_{-i} = (\boldsymbol{Z}_1,\cdots, \boldsymbol{Z}_{i-1}, \boldsymbol{Z}_{i+1}, \cdots, \boldsymbol{Z}_m)$, and $\boldsymbol{X}_{-G_i} = (\boldsymbol{X}_{G_1}, \cdots, \boldsymbol{X}_{G_{i-1}}, \boldsymbol{X}_{G_{i+1}}, \cdots, \boldsymbol{X}_{G_m})$.
Then $ \Theta_{O, H}(\Theta_H)^{-1}\Theta_{O, H}^T \succeq 0$ is low-rank diagonal blocks up to column or row permutations. This assumption is a generalization of \cite{chandrasekaran2010latent} suggesting that important unobserved latent variables exist in various communities. The observed variables in each community are only connected to the latent variables present within that community. Sparse connections between the communities are established based solely on the observed variables. Under the sparsity assumption for $\Theta_{O}$, we find that the decomposition \eqref{Model.GLGM} is also a case of the model \eqref{Model.decomposition}. 

In the sequel, our analysis is solely based on the assumption that $L\succeq 0$, though it can be easily extended to $-L\succeq 0$.

\section{Proposed method}\label{Sec.ProMethod}
\subsection{A three-stage estimation procedure}
We propose a three-stage estimation procedure to estimate the parameters $(B, S, L)$ and community memberships. In the first stage, we estimate $B$ using the least squares method, provided that $\widetilde{\boldsymbol{C}}^T\widetilde{\boldsymbol{C}}$ is nonsingular,
\[
\widehat{B} = (\widetilde{\boldsymbol{C}}^T\widetilde{\boldsymbol{C}})^{-1}\widetilde{\boldsymbol{C}}^T\widetilde{\boldsymbol{X}},
\]
where $\widetilde{\boldsymbol{C}}= (\tilde{\boldsymbol{c}}_1, \cdots, \tilde{\boldsymbol{c}}_n)^T\in\mathbb{R}^{n\times q}, \tilde{\boldsymbol{c}}_i = (\tilde{c}_{i1}, \cdots, \tilde{c}_{iq})^T, i = 1,\cdots, n$ and $\widetilde{\boldsymbol{X}} = (\tilde{\boldsymbol{x}}_1, \cdots, \tilde{\boldsymbol{x}}_n)^T \in \mathbb{R}^{n\times p}, \tilde{\boldsymbol{x}}_i = (\tilde{x}_{i1}, \cdots, \tilde{x}_{ip})^T, i = 1,\cdots, n$. Then we obtain a residual estimate by the plug-in method, i.e., $\widehat{\boldsymbol{R}} = (\boldsymbol{I}_n - \widetilde{\boldsymbol{C}}(\widetilde{\boldsymbol{C}}^T\widetilde{\boldsymbol{C}})^{-1}\widetilde{\boldsymbol{C}}^T)\widetilde{\boldsymbol{X}}$ with $\boldsymbol{I}_n$ being the $n$ dimensional identity matrix. 
\par In the second stage, we focus on the consistent estimation of $(S, L)$. Let $\widehat{\Sigma} = \frac{1}{n}\widehat{\boldsymbol{R}}^T\widehat{\boldsymbol{R}}$ be the empirical estimator of $\Sigma$, and we consider the following constrained maximum likelihood estimation problem: 
\begin{align}
			(  \widehat{S},\widehat{L}) =& \argmax_{( S,L)}\log\det(S+L  ) - \text{tr}\left(\widehat{\Sigma}(S+L ) \right)\nonumber\\
		s.t.\quad
		&\begin{cases}
			L + S\succ 0,\quad L\succeq 0,\\
			\Vert L\Vert_0 \leq l_0,\quad \Vert S\Vert_0 \leq s_0,\\
			\operatorname{rank}(L)\leq r, 
		\end{cases}\label{eq:constraints}
\end{align}
where the loss function is taken as the likelihood function for the zero-mean multivariate normal distribution. This non-convex optimization problem is computationally challenging due to the use of the $\|\cdot\|_0$ norm and the rank constraint. However, imposing the constraint $\Vert L\Vert_0 \leq l_0$ alone does not lead to a diagonal-block solution. If we instead relax the second constraint in \eqref{eq:constraints} to a combined penalty function for $L$ with respect to $\ell_1$ norm and nuclear norm $\|\cdot\|_*$, \cite{tan2022sparse}, \cite{ richard2012estimation} and \cite{oymak2015simultaneously} show that this combined penalty will generate the diagonal-block phenomenon
 up to column/row permutations. Therefore, we consider the following adaptive $\ell_1$ penalized  estimation problem,
\begin{equation}\label{penobj}
	\begin{aligned}
		(\widehat{S},\widehat{L}) =& \argmin_{( S,L)}  \ell_n(S,L) + \Pen(S, L)\\
		s.t.\quad & L + S \succ 0,\quad L\succeq 0,
	\end{aligned}
\end{equation} 
where 
$
\ell_n(S,L) = -\log\det(S +L)+\text{tr}\left(\widehat{\Sigma}(S+L) \right)
$  and 
\begin{equation}\label{Penalty}
	\Pen(S, L) =\gamma_n\|S\|_{\text{off},1}  + \delta_n\Vert L\Vert_{*}+  \tau_n\sum_{i,j = 1}^pw_{ij}| L_{ij}| ,
\end{equation}
and $w_{ij}$ are pre-specified data-driven weights.  Different from existing methods in \cite{tan2022sparse, richard2012estimation}, we use an adaptive $\ell_1$ penalty form introduced by \cite{zou2006adaptive, huang2008adaptive}, upon which we conduct the consistency analysis for model selection. We will discuss this further in Section \ref{Sec.Asymp}. 
%\sd{ Since $A$ is rotation invariant, we simply set $\widehat{A} = \widehat{U}_1\widehat{\Sigma}_1^{\frac{1}{2}}$ to capture community characteristics under balanced low-rank factor assumption (described in Assumption \ref{A.non-overlapped}) with $\widehat{U}_1\widehat{\Sigma}_1\widehat{U}_1^T$ being the eigendecomposition of $\widehat{L}$. }
\par {
	In the third stage, to uncover the membership labels, we apply the
	%identify the structure  of each community, i.e., specific members of each community,  we explore the application of 
	$K$-means clustering algorithm \citep{hastie2009elements} to $\widehat{L}$. Specifically, let $\{\hat{l}_i \}_{i=1}^p$ represent the row vectors of $\widehat{L}$, and then we seek to minimize the following loss function:
	\begin{align}\label{eq.cluster}
		\left\{\widehat{\mathcal{C}_i}\right\}_{i=1}^m& = \argmin_{\mathcal{C}_1,\cdots, \mathcal{C}_m}\sum_{i=1}^m \sum_{k,j\in\mathcal{C}_i}\|\hat{l}_k - \hat{l}_j\|^2,
	\end{align}
	where $(\mathcal{C}_1,\cdots, \mathcal{C}_m)$ represents a partition of $[p]$, and $\|\cdot\|$ here denotes the Euclidean norm of vectors. In addition, when $m = r$ can be considered as a prior, we also apply the $K$-means clustering to a specifically transformed counterpart of $\widehat{L}$. For instance, we define the correlation matrix associated with $\{\hat{l}_i\}$ as $\operatorname{Cor}(\operatorname{abs}(\widehat{L})) \triangleq (\operatorname{cor}(\operatorname{abs}(\hat{l}_i), \operatorname{abs}(\hat{l}_j)))_{p\times p}$, where $\operatorname{cor}(\operatorname{abs}(\hat{l}_i), \operatorname{abs}(\hat{l}_j))$ is the Pearson correlation between the absolute values of $\hat{l}_i$ and $\hat{l}_j$. %The rationale behind this transformation can be seen from Theorem \ref{Thm.CorHamming}. 
	
	Note that here we do not use the spectral clustering approach \citep{lei2015consistency, jin2015fast}, due to its dependence on the accuracy of rank estimation in the second stage. Also, in the sequel, we assume the number of communities (clusters) $m$ is known when clustering.
	}
	
%	The rationale behind  this transformation lies in the assumption that for the noiseless community part $L^*$, the row vectors $\{l_i^*\}_{i=1}^p$ form $m$ clusters, and the vectors in different clusters are completely orthogonal, resulting in a correlation coefficient of zero between them

\subsection{Algorithm}  

\begin{algorithm}[htp!]
	\caption{ADMM Algorithm for Solving (\ref{blockobj}).}
	\label{Algori}
	\setstretch{1.3}
	%\textsl{}\setstretch{1.8}
	\renewcommand{\algorithmicrequire}{\textbf{Input:}}
	\renewcommand{\algorithmicensure}{\textbf{Output:}}	
	\begin{algorithmic}[1]
		\renewcommand{\algorithmicrequire}{\textbf{Input:}}
		\REQUIRE The residue $\widehat{\mathbf{R}}$ computed in the first stage. Tuning parameters $(\mu, \gamma_n, \delta_n , \tau_n)>0$.  Suitable weights $w_{ij}>0$. Stopping error $\varepsilon>0 $.
		\STATE \textbf{Initialization:}
		\STATE Let primal variables  $S^0$, $L_1^0=L_2^0$ all be identity matrix, and $\Theta^0= S^0+L_1^0$. 
		\STATE Set $Y_1^0 = Y_2^0 = (\Theta^0, S^0, L^0_1, L^0_2)$. 
		\STATE Let dual variables $\Gamma_0$ be zero matrix.
		\WHILE {the stopping criterion $\|Y_1^{k+1} - Y_1^k\|_F^2/\|Y_1^k\|_F^2 \leq \varepsilon$ is met }
		\STATE For the $k$-th iteration, denote $Y_2^k = (\widetilde{\Theta}^k, \widetilde{S}^k,\widetilde{L}_1^k, \widetilde{L}_2^k)$, $\Gamma^k = (\Gamma_\Theta^k, \Gamma_S^k, \Gamma_{L_1}^k, \Gamma_{L_2}^k ) $. 
		\STATE Update $\Theta^{k+1}$:
		$$
		\Theta^{k+1} = U\operatorname{diag}(\xi)U^T,\quad \xi_i = \frac{\sigma_i + \sqrt{\sigma_i^2 + 4\mu}}{2},
		$$
		where $U\operatorname{diag}(\sigma)U^T$is the eigendecomposition for $\widetilde{\Theta}^k - \mu\widehat{\Sigma} - \Gamma_{\Theta}^k$.  
		\STATE Update $S^{k+1}$:
		$$
		\left(S^{k+1}\right)_{i j}
		= \left\{\begin{array}{l}
			\operatorname{prox}_{\mu\gamma_n ,\|\cdot\|_1}\left(\widetilde{S}^{k}_{ij}- (\Gamma_{S}^{k})_{ij}\right), i \neq j \\
			\widetilde{S}_{i j}^{k}-\left(\Gamma_{S}^{k}\right)_{i j}, \quad i = j .
		\end{array}\right. 
		$$
		\STATE Update $L_1^{k+1}$:
		$$
		L_1^{k+1} = \mathcal{P}_{\Se}\left( \left(\prox_{\mu \tau_n w_{ij}, \|\cdot\|_1}(\widetilde{L}_1^k - \Gamma_{L_1}^k) \right)_{ij}  \right),
		$$
		where $S^+$ denotes the semidefinite positive cone in $\mathbb{S}^{p\times p}$. 
		\STATE Update $L_2^{k+1}$:
		$$
		L_2^{k+1}= \prox_{\mu \delta_n, \|\cdot\|_*}(\widetilde{L}_2^k - \Gamma_{L_2}^k).
		$$
		\STATE Let $Y_1^{k+1} = (\Theta^{k+1}, S^{k+1},L_1^{k+1}, L_2^{k+1})$ and partition the matrix $T^k \triangleq Y_1^{k+1} + \Gamma^k = (T^k_{\Theta}, T^k_{S}, T^k_{L_1}, T^k_{L_2})$
		into four blocks in the same form as $Y_2^k$. \STATE Update $( \widetilde{\Theta}^{k+1},\widetilde{S}^{k+1}, 	\widetilde{L}_1^{k+1}, 	\widetilde{L}_2^{k+1})$:
		$$
		\left\{\begin{array}{l}
			\widetilde{\Theta}^{k+1}=  \frac{3}{5}T^k_{\Theta}+ \frac{2}{5}T^k_{S} + \frac{1}{5}T^k_{L_1} + \frac{1}{5}T^k_{L_2} \\
			\widetilde{S}^{k+1}= \frac{2}{5}T^k_{\Theta}+\frac{3}{5}T^k_{S}		 - \frac{1}{5}T^k_{L_1}-\frac{1}{5}T^k_{L_2}\\
			\widetilde{L}_1^{k+1} = \frac{1}{5}T^k_{\Theta} -\frac{1}{5}T^k_{S}+ \frac{2}{5}T^k_{L_1}+\frac{2}{5}T^k_{L_2}\\
			\widetilde{L}_2^{k+1}= \frac{1}{5}T^k_{\Theta}-\frac{1}{5}T^k_{S} + \frac{2}{5}T^k_{L_1}+\frac{2}{5}T^k_{L_2}    \end{array}\right. .
		$$
		\STATE Denote $Y_2^{k+1} = ( \widetilde{\Theta}^{k+1},\widetilde{S}^{k+1}, 	\widetilde{L}_1^{k+1}, 	\widetilde{L}_2^{k+1}) $. Update $\Gamma^{k+1}$:
		$$
		\Gamma^{k+1}=\Gamma^k+\left(Y_1^{k+1}-Y_2^{k+1}\right).
		$$
		\STATE Update the iteration step: $k \gets k+1.$ 
		\ENDWHILE
		\STATE  Collect the last iteration value denoted by $Y_1^{\text{final}} = (\Theta^{\text{final}}, S^{\text{final}}, L_1^{\text{final}}, L_2^{\text{final}}) $.
		\ENSURE The final result $( \Theta^{\text{final}}, S^{\text{final}}, L_2^{\text{final}} ) $.
	\end{algorithmic}  
\end{algorithm}
%$(X^{\text{final}}, Y^{\text{final}}, \Gamma^{\text{final}})$
Since $\widehat{B}$ is computationally efficient due to its closed form and equation \eqref{eq.cluster} has well-established algorithms for solving it \citep{hastie2009elements}, here our main focus is to develop an efficient alternative direction method of multipliers algorithm (ADMM, \cite{boyd2011distributed}) to solve the convex optimization problem in \eqref{penobj}. Since the global convergence of ADMM  with three blocks (or more) of variables is ambiguous in general \citep{ma2013alternating}, we rewrite the problem \eqref{penobj} as a convex minimization problem with two blocks of variables and two separable functions as follows \citep{richard2012estimation,ma2013alternating, tan2022sparse}. This form can be viewed as a special case of
the consensus problem discussed in \cite{boyd2011distributed}, whose  global convergence  has been well studied. Specifically,
we rewrite the problem in \eqref{penobj} as
\begin{equation}\label{penobjsepa}
	\begin{aligned}
		(\widehat{\Theta}, \widehat{S}, \widehat{L}_1, \widehat{L}_2) = &\argmin_{(\Theta, S, L_1, L_2)} \ell_n(\Theta) + \mathbf{Pen}(S,L) ,\\
		s.t.\quad& \begin{cases}
			\Theta = L_1 + S, \quad L_1 = L_2, \\
			\Theta \succ 0,\quad L_1, L_2\succeq 0,
		\end{cases}	
	\end{aligned}
\end{equation} 
where 
$
\ell_n(\Theta) = -\log\det(\Theta) + \text{tr}\left(\widehat{\Sigma}\Theta  \right)
$ and $\mathbf{Pen}(S,L)$ is defined as in \eqref{Penalty}.
Then denote $Y_1=(\Theta, S,  L_1, L_2)\in\mathbb{R}^{p\times4p}$, $Y_2=(\widetilde{\Theta}, \widetilde{S},  \widetilde{L}_1,\widetilde{L}_2)\in\mathbb{R}^{p\times4p}, $ the optimization problem in \eqref{penobjsepa} is equivalent to 
\begin{equation}\label{blockobj}
	\begin{aligned}
		\min\quad &f(Y_1)+\psi(Y_2)\\
		s.t. \quad& Y_1-Y_2 = 0,
	\end{aligned}
\end{equation}
where 
$$
f(Y_1) \triangleq \ell_n(\Theta) + \mathbf{Pen}(S,L),
 $$ 
$$
\psi(Y_2) \triangleq 
	 \mathcal{I}\{\widetilde{\Theta}-\widetilde{L}_1-\widetilde{S}=0\} + 
	 \mathcal{I}\{\widetilde{L}_1-\widetilde{L}_2=0\},
$$ 
and the indicator function is defined as 
$$
\begin{aligned}
	\mathcal{I}(X\in \mathcal{X})= \begin{cases}0, & \text { if } X\in \mathcal{X} \\ +\infty, & \text { otherwise }\end{cases}.
\end{aligned}
$$

Define the augmented Lagrangian function associated with \eqref{blockobj} as 
$$
\mathcal{L}_{\mu}(Y_1, Y_2,\Gamma)=f(Y_1)+\psi(Y_2)+\Gamma\cdot (Y_1-Y_2)+\frac{1}{2\mu}\|Y_1-Y_2\|_F^2,
$$
where 
$$
A \cdot B\triangleq\sum_{i=1}^p \sum_{j=1}^p A_{i j} B_{i j}=\operatorname{Tr}\left(A ^TB\right),
$$
and  $\Gamma\in\mathbb{R}^{p\times4p}$ is the Lagrange multiplier. Here $\mu>0$ is the penalty parameter. Then we divide \eqref{blockobj} into solvable subproblems through the scaled ADMM algorithm:
$$
\left\{\begin{aligned}
	Y_1^{k+1} &= \argmin_{Y_1}\mathcal{L}_{\mu}(Y_1, Y_2^k,\Gamma^k) =\argmin_{Y_1} f(Y_1) + \frac{1}{2 \mu}\left\|Y_1-Y_2^k + \Gamma^k\right\|_F^2 \\
	Y_2^{k+1} & =\argmin_{Y_2}\mathcal{L}_{\mu}(Y_1^{k+1}, Y_2,\Gamma^k) =\argmin_{Y_2} \psi(Y_2) + \frac{1}{2 \mu}\left\|Y_1^{k+1}-Y_2 + \Gamma^k\right\|_F^2, \\
	\Gamma^{k+1} & =\Gamma^k+\left(Y_1^{k+1}-Y_2^{k+1}\right).
\end{aligned}\right.
$$
We also introduce some useful proximal mappings (see Appendix \ref{Appen.Algorithm} for details) to design the algorithm. The proximal mapping for $p(X) = \|X\|_1$ is 
$$
\begin{aligned}
	\operatorname{prox}_{\lambda, \|\cdot\|_1}(Z) 
	&\triangleq\operatorname{sign}(Z)(Z-\lambda\textbf{1}_p\textbf{1}_p^T)_+,
\end{aligned}
$$
where $(\cdot)_+ = \max(\cdot, 0)$ and $\mathbf{1}_p=(1,\cdots, 1)^T\in\mathbb{R}^p$. 
 The proximal mapping for $p(X) = \|X\|_*, X\succeq 0$ is 
$$
\begin{aligned}
	\operatorname{prox}_{\lambda, \|\cdot\|_*}(Z)
	&\triangleq U\widetilde{\Sigma} U^T,
\end{aligned}
$$
where $Z = U\Sigma U^T, \Sigma = \operatorname{diag}(\sigma_1, \dots, \sigma_p)$ and $\widetilde{\Sigma} = \operatorname{diag}\left((\sigma_1 -\lambda)_+,\dots, (\sigma_p -\lambda)_+ \right)$. 
In addition, the proximal mapping for $p(X) = I(X\in \mathcal{X})$ is the projection mapping
 $\mathcal{P}_{ \mathcal{X}}(Z)$ denoting the projection of $Z$ on set $\mathcal{X}$ with respect to the inner product defined above. The complete algorithm is summarized in the Algorithm~\ref{Algori}, and a detailed description of this algorithm is provided in Appendix \ref{Appen.Algorithm}.

\subsection{Weight assignment and choices of tuning parameters}\label{Sec.Weight}
According to \cite{huang2008adaptive} and \cite{zou2006adaptive}, compared to the standard $\ell_1$ penalty, incorporating a high penalty for zero elements and a low penalty for nonzero elements in the true values of $L$ within the adaptive $\ell_1$ penalty reduces the estimation bias and improves the accuracy of variable selection. We start by implementing the method given in \cite{chandrasekaran2010latent} to determine an initial weight assignment:
\begin{equation}\label{eq.InitialEsti}
	\begin{aligned}
		( \bar{S},\bar{L} ) &= \argmin_{(\Theta, S, L)} \ell_n(S,L)+   \lambda_n\|S\|_{\operatorname{off},1}+\gamma_n\Vert L\Vert_{*}\\
		s.t.\quad & \Theta = L + S, \quad \Theta\succ 0,\quad L\succeq 0.
	\end{aligned}
\end{equation} 
Denote $\bar{L} = (\bar{l}_{ij})_{p\times p}$, then the initial weight estimate $\widetilde{W}_n $ is set as $\tilde{w}_{ij} \triangleq \bar{l}_{ij}^a$, for $i, j = 1,\ldots, p$, and we assume $a>0$ which is typically set as 1 or 2. 
{Subsequently, the adaptive weight $W_n = (w_{ij})_{p\times p}$ assigned in \eqref{Penalty} is:}
\begin{equation}\label{eq.Weight} 
	w_{ij} = \frac{1}{\tilde{w}_{ij}}.
\end{equation}

Moreover, 
we rely on the
cross-validation (CV) method to choose tuning parameters, which in general is more accurate than using BIC but with higher computational costs. We randomly split the rows of residue $\boldsymbol{\widehat{R}}$ into $M$ segments of equal size, and denote the sample covariance matrix using the data in the $t$-th segment
$(t = 1,..., T)$ by $\widehat{\Sigma}^t$. Let $(\widehat{S}_{\Xi}^{-t}, \widehat{L}^{-t}_\Xi)$ be the estimator obtained by solving \eqref{penobj} but using all data excluding
 the $t$-th segment with the tuning parameters $\Xi \equiv (\gamma_n, \delta_n, \tau_n)$. 
We then choose $\widehat{\Xi}$ that minimizes the average predictive negative log-likelihood:
$$
\operatorname{CV}(\Xi)=\sum_{t=1}^T\left[\operatorname{tr}\left(\widehat{\Sigma}^{t} (\widehat{S}_\Xi^{-t} + \widehat{L}_\Xi^{-t})\right)-\log \left(\operatorname{det}(\widehat{S}_\Xi^{-t} + \widehat{L}_\Xi^{-t})\right)\right],
$$
i.e. we set tuning parameters as $\widehat{\Xi} = \operatorname{argmin}_{\Xi\in \mathbb{R}_+^3}\operatorname{CV}(\Xi)$.

\section{Theoretical properties}\label{Sec.Theory}

In this section, we will discuss the identifiability issues in the model (\ref{Model.decomposition}) and provide theoretical guarantees for the three-stage estimation procedure. Throughout the discussion, we will always denote the true parameters by ($B^*, S^*, L^*$).
\subsection{Identifiability} \label{Sec.Identi}
The identifiability for $B^*$ can be ensured by assuming that the fixed design matrix $\widetilde{\boldsymbol{C}}$ is of column full rank.  We mainly present conditions for identifiability of ($S^*, L^*$) by resorting to the tools developed by
\cite{chandrasekaran2010SIAM,chandrasekaran2010latent}. To proceed, we will describe two submanifolds of $\mathbb{S}^{p\times p}$ and their tangent spaces. 
\par Denote $s_0 = \|\mathbf{O}(M)\|_0$ with $\mathbf{O}(M) \triangleq M - \operatorname{diag}(M)$. Consider the set of sparse matrices with $s_0$ nonzero elements except the diagonal, let
\[\mathcal{S}(s_0) = \left\{ M \in \mathbb{S}^{p\times p}: \|\operatorname{Supp}(\mathbf{O}(M))\|_0 \leq s_0 \right\},\]
 It is a subspace of $\mathbb{S}^{p\times p}$ and thus can also be viewed as a submanifold of $\mathbb{S}^{p\times p}$. The tangent space at any smooth point $S\in \mathcal{S}(s_0)$ around $S^*$ is all given by 
\[ \Omega(S) =  \{N: \operatorname{Supp}(\mathbf{O}(N))\subseteq \operatorname{Supp}(\mathbf{O}(S^*))\}. \]

To describe the tangent space of the low-rank diagonal blocks $L$ up to row and column permutations near $L^*$, we make the first assumption for $L^*$. 

\begin{assumption}\label{A.non-overlapped}
	Denote the eigendecomposition of $L^*$ by $L^* = U_1^*\Sigma_1^*U_1^{*T}$, where $\Sigma_1^* = \operatorname{diag}(\sigma^*_1,\cdots, \sigma^*_r)$, $U_1^* \in\mathbb{R}^{p\times r}, U_1^{*T}U_1^* = I_r$. Assume that 
	\begin{enumerate}[label = {(\arabic*)}]
		\setlength{\itemsep}{0pt}
		\setlength{\topsep}{0pt}
		\setlength{\parskip}{1pt}
		\item (Eigengap) $\sigma_1^*>\cdots>\sigma^*_r>0$.
	    \item (Diagonal-block) $L^*$ has a diagonal-block structure up to column and row permutations. Without loss of generality, let $L^* = \operatorname{diag}(L_1^*,\cdots, L_m^*), L^*_i\in\mathbb{S}^{d_i\times d_i}, \sum_{i=1}^m d_i=p$ and $\forall i\in [r]$, $L_i^*\not=0$ in the entry-wise sense. In addition, denote $\operatorname{rank}(L_i^*) = r_i$, $\sum_{i=1}^mr_i =r$.
%	    \item (Balanced low-rank factor) 
%             $A^* = U_1^*\Sigma_1^{*\frac{1}{2}}$.
	\end{enumerate}
\end{assumption}
We make some remarks on this assumption. Part (1) is the eigengap assumption commonly imposed in the literature of matrix perturbation \citep{Yu2014,fan2018eigenvector}, which is necessary to identify the eigenspace of $L^*$.  Part (2) states that $L^*$ has $m$ diagonal blocks up to row and column permutations and each block is low-rank. To fix the idea, we directly assume that $L^*$ is diagonal and this condition emphasizes that the sparsity of the truth $L^*$ only comes from off-diagonal blocks. To clarify this point,  define   $G^*_{ij} = \left\{(k,l)|k\in \{d_{i-1}+1,\cdots, d_{i-1} + d_i\}, l\in\{r_{j-1}+1,\cdots, r_{j-1} + r_j\}  \right\}$ for $i,j\in [m]$ and  define $s_0 = 0, r_0 =0$. Then $\{G_{ij}^*\}_{(i,j)\in [m]\times [m]}$ forms a partition of $[p]\times [r]$ and $(U_1^*)_{G_{ij}^*} =\mathbf{0}_{d_i\times r_j} $ for $i\not=j\in [m]$. It is easy to see that when $r = m$, part (2) will be automatically satisfied, which means that each block has at most one dimension, i.e.,  each row of $U_1^*$ has at most one nonzero element.  Essentially, part (2)  requires that the sparsity of the eigenspace is mainly used to derive the sparsity of the off-diagonal-blocks of $L^*$ and the eigenspace corresponding to the diagonal-blocks is as non-sparse as possible. Furthermore, in the context of this paper, different diagonal blocks represented by $L_i$ signify distinct communities (clusters). To more accurately identify the members of each community, we define the true labels of different communities as $\mathcal{C}_i^* = \{d_{i-1}+1, \cdots, d_{i-1} + d_i\}, i = 1, \cdots, m$  and $\{\mathcal{C}_i^*\}_{i=1}^m$ forms a partition of $[p]$.
\par Now we consider the set of diagonal-block matrices with the rank of each block less than $r_i$. Let
\begin{equation*}
\begin{aligned}
	\mathcal{LS}(m, r) = \bigg\{L = \operatorname{diag}(L_1,\cdots, L_m): \operatorname{rank}(L_i)\leq r_i,L_i \in\mathbb{S}^{d_i\times d_i} \bigg\} .
\end{aligned}
\end{equation*} 
When $m = 1$, this set is the classical  algebraic variety of low-rank matrices $ \mathcal{L}(r_i)\triangleq \mathcal{LS}({1,r_i})$ with tangent space $\maT_1(L_i) = \{U_{1i}Y_{1i}^T + Y_{1i}U_{1i}^T: Y_{1i}\in\mathbb{R}^{d_i\times r_i}\}$, where $U_{1i} $ is the $r_i$ leading eigenvectors of $L_i$ \citep{chandrasekaran2010SIAM} (when we say “leading eigenvectors”, we are comparing
the magnitudes of the eigenvalues by neglecting the signs). Under Assumption \ref{A.non-overlapped}, all the smooth points $L \in \mathcal{LS}(m,r)$ form a submanifold of $\mathbb{S}^{p\times p}$ satisfying 
$\operatorname{rank}(L_i) = \operatorname{rank}(L_i^*) = r_i$, and $\operatorname{Supp}(L_i) = \operatorname{Supp}(L_i^*) $, which is a simple application of property of product manifold (details see Appendix \ref{Appen.ProProof.Tangent}). The tangent space at any smooth point of this submanifold can be viewed as an intersection of two spaces. We summarize this property in the following proposition.
\begin{proposition}\label{P.TangentSpace}
	The tangent space  at any smooth point $L= U_1\Sigma U_1^T\in\mathcal{LS}(m, r)$ is given by 
	\begin{equation*}
		\begin{aligned}
		\mathcal{T}(L)&  = \bigg\{ N = \operatorname{diag}(N_1,\cdots, N_m):   N_i\in\mathbb{S}^{d_i\times d_i}, N_i\in \maT_1(L_i) \bigg\} = \maT_1(L)\cap \maT_2(L^*) ,
		\end{aligned}
	\end{equation*}	
where $ \maT_1(L) = \left\{U_1Y^T + YU_1^T: Y\in\mathbb{R}^{p\times r} \right\},   \maT_2(L^*) = \{ N = \operatorname{diag}(N_1,\cdots, N_m): \operatorname{Supp}(N_i)\subseteq \operatorname{Supp}(L_i) \}$. Moreover,
\begin{equation*}
	\maT^\perp(L) = \maT_1^\perp(L) + \maT_2^\perp(L^*) = \maV(L) \oplus \maT_2^\perp(L^*) ,
\end{equation*} 
with $\maV(L) = \left\{ N = \operatorname{diag}(N_1,\cdots, N_m): N_i\in\mathbb{S}^{d_i\times d_i},  N_i\in \maT^\perp_1(L_i)  \right\}$.
\end{proposition}
\par Denote the spaces evaluated at the true value by $\Omega(S^*) = \Omega^*$, $\maV(L^*) = \maV^*$, $\maT_2(L^*) = \maT^*_2$, $\maT(L^*) = \maT^*$ and $\maT_2^\perp(L^*) = \maT_2^{*\perp}$. In addition, for each $i$, let $S_i^* \triangleq S^*_{\mathcal{C}_i^*\times \mathcal{C}_i^*}$ be the submatrix in the diagonal of $S^*$ exhibiting the same block size as $L_i^*$.  For identifiability, we need the following transversality condition \citep{chandrasekaran2010latent,chen2016fused,zhao2023identifiability,hsu2011robust}. 

\begin{assumption}\label{A.transver}
$\Omega^* \cap \maT^* =\mathbf{0}_{p\times p}$.
\end{assumption}
According to Proposition \ref{P.TangentSpace}, this assumption is equalivent to $ \Omega(S^*_{i}) \cap \maT_1(L_i^*) = \mathbf{0}_{d_i\times d_i}, i = 1, \cdots, m$, which says identifiability of $(S^*, L^*)$ can be determined by that of $(S_i^*, L_i^*)$ in the diagonal blocks. Note that this assumption allows for the possibility that $S$ can be non-zero in the off-diagonal block positions of $L$. Now we state that the non-overlapped community part $L$ and the sparse structure $S$ can be locally identified from the data. 
 \begin{theorem}\label{Thm.IdentiforL}
	 Under Assumptions \ref{A.non-overlapped} (2) and \ref{A.transver}, $ (S^*, L^*) $ is locally identifiable with respect to $\mathcal{S}(s_0)\times \mathcal{LS}(m, r)$. Moreover, if Assumption \ref{A.non-overlapped} (1) also holds, $U_1^*$ is also locally identifiable up to sign flips.
\end{theorem}
\begin{remark}
For convenience, let $\mathcal{E}(r) $ be the set of  $r\times r$ sign flip matrices, i.e., for any $M\in\mathbb{R}^{p\times r}$ and $P\in\mathcal{E}(r)$, and $MP$ is a matrix whose $k$-th column equals to the $k$-th column of $M$ multiplied by $1$  or $-1$. Theorem \ref{Thm.IdentiforL} claims that there exists a neighborhood $\mathcal{N}(S^*, L^*)$ of $(S^*,L^*)$ in $\mathcal{S}(s_0)\times \mathcal{LS}(m, r)$ such that if it holds that $S+L = S^*  +L^*$ for $(S,L)\in\mathcal{N}(S^*, L^*)$, then $S = S^*, L_i=L_i^*, U_{1i} = U_{1i}^*P_i $ for some $P_i\in \mathcal{E}(r_i)$, where $L = \operatorname{diag}(L_1, \cdots, L_m)$ and $U_{1i},U_{1i}^*$ are the $r_i$ leading eigenvectors of $L_i, L_i^*$ respectively.
\end{remark}
%\begin{theorem}\label{Thm.IdentiforL}
%	 Under Assumptions \ref{A.non-overlapped} (1) (2) and \ref{A.transver}, there exists $\varepsilon >0$ such that for any smooth point $S\in \mathcal{S}^*$, $L \in\mathcal{LS}(L^*)$  satisfying
%	  $\|S - S^*\|_{\infty}\leq \varepsilon,   \|L- L^*\|_{\infty}\leq\varepsilon,  $
%	  if
%	 $$
%	        S+L = S^* + L^*,
%	 $$
%	 it holds that
%	$$ (S,L) = (S^*, L^*) .$$ Moreover,  denote the eigendecomposition $L= U_1\Sigma_1 U_1^T$ and $A = U_1\Sigma_1^{\frac{1}{2}}$. If Assumption \ref{A.non-overlapped} (3) also holds, then there exists $R\in \mathcal{E}(r)$ such that $AR = A^*$.
%\end{theorem}

\subsection{Asymptotic analysis}\label{Sec.Asymp}
\par  In this section, we provide some theoretical guarantees for the proposed three-stage estimation procedure for  $(\widehat{B}, \widehat{S}, \widehat{L})$ and $\{\widehat{\mathcal{C}}_i\}_{i=1}^m$ in the asymptotic regime, where we fix $p, q$ and let $n\to \infty$ only. For $\widehat{B}$ in the first stage, we make the following commonly used assumption, which ensures that $\widehat{B}$ is $\sqrt{n}$-consistent for $B^*$ from the classical linear regression theory.
\begin{assumption}\label{A.Bhat}
	$\frac{1}{n}\boldsymbol{\widetilde{C}}^T\boldsymbol{\widetilde{C}} \to C^*$ for the positive definite matrix $C^*$.
\end{assumption} 
When it comes to  $\delta\|L\|_* + \|L\|_1$ type penalty in the second stage, the existing studies mainly focus on the estimation error problem \citep{tan2022sparse,richard2012estimation} or signal processing problem \citep{oymak2015simultaneously}. This work will establish the model selection consistency and provide a convergence rate under the adaptive type penalty \eqref{Penalty}. 
\par  The key point depends on the discussion about $L$. Therefore, we temporarily ignore the part of $S$ and denote $\tilde{\ell}(L) = \ell_n(S,L)$. According to the optimality condition \citep{boyd2004convex} to the problem \eqref{penobj}, we want to solve the following equation for smooth points $L\in\mathcal{LS}(L^*) $ with $L\succeq 0$,
\begin{equation*}
     -\nabla\tilde{\ell}(L)\in\partial(\delta_n\|L\|_* +\tau_n\|W_n\odot L\|_1),
\end{equation*}
where $\odot$ is the Hadamard product. It is equivalent to solving the equations (see Lemma \ref{L.LSubgradient} in Appendix \ref{Appen.Lemmas} for the expression of this subgradient)
\begin{equation}\label{Pro.Projequations}
    \begin{aligned}
 \mathcal{P}_{\maT(L)}( -\nabla\tilde{\ell}(L)) &= \delta_n UU^T + \tau_n\mathcal{P}_{\maT(L)}(W_n\odot\operatorname{sign}(L^*)),\\
 \mathcal{P}_{\maT^\perp(L)}( -\nabla\tilde{\ell}(L)) &= \tau_n\mathcal{P}_{\maT^\perp(L)}(W_n\odot\operatorname{sign}(L^*)) + \delta_n F_1 + \tau_n W_n\odot F_2,\\
\end{aligned}
\end{equation}
where $F_1\in \maT^\perp_1(L), F_2\in \maT_2^{*\perp}, \|F_1\|_{\infty}\leq 1, \|F_2\|_2\leq 1$. Second, to solve these two equations, we resort to the primal-dual witness technique \citep{Ravikumar2011,chandrasekaran2010latent,chen2016fused}, i.e., we obtain a solution $\tilde{L}$ from the first equation. Then we substitute this solution into the second equation to construct suitable dual elements $F_1$ and $F_2$. Finally, we establish the uniqueness of this solution. This procedure encounters two primary challenges.  The first one is related to the extra terms $\tau_n\mathcal{P}_{\maT(L)}(W_n\odot\operatorname{sign}(L^*))$ and $\tau_n\mathcal{P}_{\maT^\perp(L)}(W_n\odot\operatorname{sign}(L^*))$, which can affect the existence of the solution. The second difficulty lies in identifying dual elements $F_1$ and $F_2$ that satisfy the second equation through some suitable norm. Unlike the existing literature that mainly focuses on the decomposable penalty function \citep{Sahand2012, candes2013simple}, the penalty involved in this problem does not satisfy the decomposable condition, making it challenging to directly use the dual norm of this penalty to identify the dual elements.

\par The first difficulty can be solved by appropriate weight selection, and the extra terms will vanish with rate $\gamma_n$ in probability. In fact, we need the weight matrix $W_n$ in \eqref{Penalty} to satisfy the following assumption \citep{zou2006adaptive, huang2008adaptive}:
\begin{assumption}\label{A.Wn}
       Let $J_{1}$ be the support set of $L^*$. The initial estimator $\widetilde{W}_n = (\tilde{w}_{ij})_{p\times p}$ defined in \eqref{eq.Weight} is $r_n$-consistent 
	\[
	r_n\max_{i,j}\left|(\widetilde{W}_{n})_{ij} - (\Lambda_{n})_{ij}\right| = O_P(1), \quad r_n \to \infty,
	\] 
	where $\Lambda_{n}$ are unknown nonrandom constants satisfying 
	\[
	\max_{(i,j)\not\in J_{1}}\left|(\Lambda_{n})_{ij}\right|\leq M_{n2} = O(\frac{1}{r_n^{1-\alpha}}),\quad \max_{(i,j)\in J_1} \frac{1}{(\Lambda_{n})_{ij}}\leq M_{n1} = o(r_n^{1- \alpha}),
	\]
	for sufficiently small $\alpha>0$.
\end{assumption}
\begin{remark}
This assumption is a special case of \cite{huang2008adaptive}. It assumes that the initial estimator $(\widetilde{W}_n)_{ij}$ actually estimates some proxy $(\Lambda_{n})_{ij}$ of $(L_{n}^*)_{ij}$,
so that the weight $(W_n)_{ij} = |(\widetilde{W}_{n})_{ij}|^{-1}$
is not too large for $(L_{n}^*)_{ij}\not= 0$ and not too small for
 $(L_{n}^*)_{ij} = 0$. According to Theorem 4.1 in \cite{chandrasekaran2010latent}, the choice of weights in section \ref{Sec.Weight} satisfies this assumption by taking $r_n = \sqrt{n} $ for fixed $p$ and $\Lambda_n = L^*$. By reparameterization, we first denote $\delta_n = \rho_1\gamma_n, \tau_n = \frac{\rho_2\gamma_n}{r_n^{1-\alpha}}$. Then according to Lemma \ref{L.WeightOrder} in Appendix \ref{Appen.Lemmas}, we find $\tau_n\mathcal{P}_{\maT(L)}(W_n\odot\operatorname{sign}(L^*))$ and $\tau_n\mathcal{P}_{\maT^\perp(L)}(W_n\odot\operatorname{sign}(L^*))$ are all $o_P(\gamma_n)$.
\end{remark}
\par  To overcome the second difficulty, we construct the following useful norm, which is only defined on $\Omega^{*\perp}\times \maT^{\perp}(L)$ with $ \maT^\perp(L) =\mathcal{ V}(L) \oplus \maT_2^{*\perp}$, 
\begin{equation}\label{eq.Dualuppernorm}
h_{\rho_1,\rho_2, L, \widetilde{W}_n}(S,L') = \max\bigg\{ \|S\|_{\infty}, \frac{\|L_1\|_2}{\rho_1}, \frac{r_n^{1-\alpha}\|\widetilde{W}_n\odot L_2\|_{\infty}}{\rho_2} \bigg\},
\end{equation}
where $L' = L_1 \oplus L_2$ is the orthogonal direct sum decomposition of $L'$ restricted to $  \maV(L) \oplus \maT_2^{*\perp}$, i.e., $L_1 = \mathcal{P}_{\maV(L)}(L') , L_2 = \mathcal{P}_{\maT_2^{*\perp}}(L') $. This norm is well-defined for fixed $L$ due to the uniqueness of the decomposition and $\maV(L)$ has an invariant dimension for any smooth point $L\in\mathcal{LS}(m,r)$ according to Proposition \ref{P.TangentSpace}. It can be verified that with probability tending to 1, combing $ \tau_n\mathcal{P}_{\maT^\perp(L)}(W_n\odot\operatorname{sign}(L^*))= o_P(\gamma_n)$ and
$$
 h_{\rho_1,\rho_2,L, \Lambda_n}(0,  \mathcal{P}_{\maT^\perp(L)}( -\nabla\tilde{\ell}(L))) < \gamma_n 
$$ 
imply the existence of dual elements $F_1, F_2$ (see Lemma \ref{L.LSubgradient} in Appendix \ref{Appen.Lemmas}).

\par After completing the above preparations, we present the adaptive irrepresentability condition that is important for the consistency of the adaptive $\ell_1$ penalized estimator defined in equation \eqref{penobj}. The core quantity involved here is the positive definite Fisher information matrix evaluated at the true parameter, i.e., 
$$\mathcal{I}^* \triangleq \mathcal{I}(\Theta^*) = \Theta^{*-1}\otimes\Theta^{*-1} = (S^* + L^*)^{-1}\otimes(S^* + L^*)^{-1},$$
where $\otimes$ represents the matrix Kronecker product and $\mathcal{I}^*\in\mathbb{R}^{p^2\times p^2}$. We also treat $\mathcal{I}^*$ as a tensor product operator, i.e., $\mathcal{I}^*: \mathbb{R}^{p\times p} \to \mathbb{R}^{p\times p}$. To abbreviate notation, we will abuse the meaning of $\mathcal{I}^*(M)$ and $\mathcal{I}^*\cdot v(M)$$\left( = v(\mathcal{I}^*(M))\right)$, where the latter one denotes matrix-vector multiplication.
Define two linear operators $\mathbf{G}: \Omega^* \times \maT^*\rightarrow \Omega^* \times \maT^*$ and $\mathbf{G}^{\perp}: \Omega^* \times \maT^* \rightarrow \Omega^{* \perp} \times \maT^{* \perp}$,
$$
\mathbf{G}(S, L)=\left(\mathcal{P}_{\Omega^*}\left\{\mathcal{I}^*(S+L)\right\}, \mathcal{P}_{\maT^*}\left\{\mathcal{I}^*(S+L)\right\}\right),
$$
$$
\mathbf{G}^{\perp}\left(S^{\prime}, L^{\prime}\right)=\left(\mathcal{P}_{\Omega^{* \perp}}\left\{\mathcal{I}^*\left(S^{\prime}+L^{\prime}\right)\right\}, \mathcal{P}_{\maT^{*\perp}}\left\{\mathcal{I}^*\left(S^{\prime}+L^{\prime}\right)\right\}\right).
$$
\begin{assumption}\label{A.irrepresentable}
	The adaptive irrepresentability condition is, as $n\to\infty$,
	\[ h_{\rho_1,\rho_2, L^*, \Lambda_n}\left(\mathbf{G}^{\perp}\mathbf{G}^{-1}(\operatorname{sign}(\mathbf{O}(S^*)), \rho_1 U_1^*U_1^{*T})\right) <1,\]
	where recall $\mathbf{O}(S^*)= S^* - \operatorname{diag}(S^*)$. The invertibility of $\mathbf{G}$ is ensured by Lemma \ref{L.FInvertible} in  Appendix \ref{Appen.Lemmas}.
\end{assumption}
\begin{remark}
The adaptive irrepresentability assumption generalizes the strong representability condition proposed by \cite{zhao2006model} and \cite{Ravikumar2011} for the pursuit of not only rank consistency but also diagonal-block structure consistency of $L^*$.  If $\operatorname{sign}(\Lambda_{n}) = \operatorname{sign}(L_{n}^*)$, we say that the initial estimates are zero-consistent with rate $r_n$. In this
case, $M_{n2} = 0$ and Assumption \ref{A.irrepresentable} will degenerate to assumption A4 in \cite{chen2016fused}. 
%In addition,  when $L^*$ is taken as a low-rank graph with diagonal blocks and $S^*$ is taken as a bounded degree graph, it is sufficient to make the Assumption \ref{A.transver} and \ref{A.irrepresentable} hold using the language developed by \cite{chandrasekaran2010latent,chandrasekaran2010SIAM} (details see Appendix \ref{Appen.Conditioncheck}).
\end{remark}
  
\par Now, we state the main theorem about the adaptive $\ell_1$ penalized estimator in the second stage.
\begin{theorem}\label{Thm.Main}
	Under Assumptions \ref{A.non-overlapped}-\ref{A.irrepresentable}, let tuning parameter $\gamma_n\asymp n^{-\frac{1}{2}+\eta}$ for some sufficiently small positive constant $\eta$, and $\delta_n=\rho_1 \gamma_n, \tau_n = \frac{\rho_2\gamma_n}{r_n^{1-\alpha}}$ with $\rho_1,\rho_2$ satisfying Assumption \ref{A.irrepresentable}. Then, with probability tending to one, the optimization problem \eqref{penobj} obtains a unique solution $(\widehat{S}, \widehat{L})$ that is both estimation consistent and selection consistent in the following sense,
	$$
	\begin{aligned}
	    &\|\widehat{S}- S^*\|_{\infty}\lesssim_P n^{-\frac{1}{2} + 2\eta}, \quad \|\widehat{L}- L^*\|_{\infty}\lesssim_P n^{-\frac{1}{2} + 2\eta},	    
	\end{aligned}
	$$	
and
	$$
	\lim_{n\to \infty}\mathbb{P}\left(\operatorname{sign}(\widehat{S}) = \operatorname{sign}(S^*)\right) =1,
	$$
	$$
     \lim _{n \rightarrow \infty} \mathbb{P}\left(\operatorname{sign}(\widehat{L})=\operatorname{sign}\left(L^*\right), \operatorname{rank}(\widehat{L})=\operatorname{rank}\left(L^*\right)\right)=1 .
	$$
\end{theorem}
This theorem shows that the adaptive $\ell_1$ penalized estimator in the second stage achieves both the rank consistency and diagonal-block structure consistency for the non-overlapped community part $L^*$, and sparsity consistency for the sparse structure $S^*$ with high probability. From the proof of Theorem \ref{Thm.Main}, we can also get some consistency results for the $r$ leading eigenvectors $ \widehat{U}_1$ of $\widehat{L}$.
\begin{theorem}\label{Thm.LatentA}
	Suppose all the conditions in Theorem \ref{Thm.Main} are satisfied, there exists $P_{n}\in \mathcal{E}(r)$ such that
	$$
	\|\widehat{U}_1P_{n} - U_1^*\|_\infty \lesssim_P n^{-\frac{1}{2} + \eta}, \quad \lim_{n\to\infty}\mathbb{P}\left( \left(\widehat{U}_1P_{n}\right)_{G^*_{ij}} = 0, \forall i\not=j  \right) = 1.
	$$
\end{theorem}
%\begin{theorem}\label{Thm.LatentA}
%	Suppose all the conditions in Theorem \ref{Thm.Main} are satisfied, it holds that 
%	\[
%	\min_{O\in\mathcal{O}^{r\times r}} \|\widehat{U}O - U^*\|_2\lesssim_Pn^{-\frac{1}{2} + 2\eta},
%	\]
%	where $\mathcal{O}^{r\times r}$ collects all the orthogonal matrices in $\mathbb{R}^{r\times r}$.  Moreover, if Assumption \ref{A.non-overlapped} (3) also holds,
%	then there exists $P_{n}\in \mathcal{E}(r)$ such that
%	$$
%	\|\widehat{A}P_{n} - A^*\|_\infty \lesssim_P n^{-\frac{1}{2} + 2\eta}, \text{ and }\lim_{n\to\infty}\mathbb{P}\left( \left(\widehat{A}P_{n}\right)_{G^*_{ij}} = 0, \forall i\not=j  \right) = 1,
%	$$
%\end{theorem}
\begin{remark}
Under suitable assumptions, this theorem states that with high probability we can identify all the zero blocks $(U^*_1)_{G^*_{ij}}, i\not=j$ in the eigenspace corresponding to the off-diagonal zero blocks of $L^*$. In contrast, the sparsity within the eigenspace $(U_1^*)_{G^*_{ii}}$ corresponding to the diagonal blocks of $L^*$ cannot be identified in general. Note that this theorem is stronger than Theorem \ref{Thm.Main}, it shows that we can even obtain the rank consistency for each diagonal block $L_i,i\in [m].$
\end{remark}
When $m = r$, the sparsity in the eigenspace $U_1^*$  is equivalent to the sparsity in $L^*$. This is a direct corollary of Theorem \ref{Thm.LatentA}. 
\begin{corollary}\label{Coroll.SimPerfect}
	Suppose $m = r$, then under Assumptions \ref{A.non-overlapped} - \ref{A.irrepresentable}, there exists  $P_{n}\in \mathcal{E}(r)$ such that
	$$
	\lim_{n\to\infty}\mathbb{P}\left( \operatorname{sign}(\widehat{U}_1P_{n} )= \operatorname{sign}(U_1^*) \right) = 1,
	$$
	for the $r$ leading eigenvectors $\widehat{U}_1$ of $\widehat{L}$.
\end{corollary}

In the third stage, we performed $K$-means clustering on the (transformed) row vectors of $\widehat{L}$. Recall that $[p] = \mathcal{C}_1^* \cup\cdots \cup \mathcal{C}_m^*$ is the true community partition. We introduce the $p\times 1$ vector $ \boldsymbol{t}$ of true labels such that
$$
\boldsymbol{t}(i) = k\quad  \text{if and only if}\quad i\in \mathcal{C}_k^*, \quad 1\leq i \leq p, 1\leq k\leq m.
$$
For the clustering procedure \eqref{eq.cluster}, there is a (disjoint) partition  $[p] = \widehat{\mathcal{C}}_1 \cup\cdots \cup \widehat{\mathcal{C}}_m$ with estimated labels 
$$
\hat{\boldsymbol{t}}(i) = k\quad  \text{if and only if}\quad i\in \widehat{\mathcal{C}}_k, \quad 1\leq i \leq p, 1\leq k\leq m.
$$
Then the proportion of mis-clustered
nodes (also called Hamming error rate) which has been previously used as a loss function for community detection in
blockmodels \citep{ma2020universal, jin2015fast}  is defined as  
$$
H(\hat{\boldsymbol{t}}, \pi(\boldsymbol{t})) \triangleq \argmin_{\pi\in S_m}\frac{1}{p} \sum_{i=1}^p\mathbbm{1}({\hat{\boldsymbol{t}}(i)\not=\boldsymbol{t}(\pi(i))}),
$$
where $
S_m = \{\pi: \pi \text{ is a permutation of the set $[m]$}\}$. The indicator function $\mathbbm{1}({x\in \mathcal{X}})$ equals 1 if $x\in\mathcal{X}$, and equals 0 otherwise.

\begin{theorem}\label{Thm.Hammingerror}
Denote the rows of $L^*$ by $\{l_i^*\}_{i=1}^p$, and define $\mu_j = \frac{1}{d_j}\sum_{i:\boldsymbol{t}(i) = j}l_i^* $ as the center of the $j$-th community. Suppose that $c \triangleq\min_{j \not= j'}\frac{1}{2}\|\mu_j - \mu_{j'}\| > 0$ and  $d_{j} \in [\frac{p}{\omega m}, \frac{\omega p}{m}], j=1, \cdots, m $, then for any $\widehat{L}$ and the respective estimated labels $\hat{\boldsymbol{t}}$, we have
deterministically that
$$
H(\hat{\boldsymbol{t}}, \pi(\boldsymbol{t})) \leq \frac{C\omega }{pc^2}\left( \|\widehat{L} - L^*\|_F^2 + \sum_{i=1}^p\|l_i^* - \mu^*_{\boldsymbol{t}(i)}\|^2 \right),
$$
 where $C>0$ is an absolute constant and $\omega \ge 1$ regularizes community size
 variability. 
\end{theorem}
\begin{remark}
	This bound for Hamming error rate is similar to Proposition 19 in \cite{ma2020universal} involving two components.  The latter component does not depend on
	data and hence can be viewed as an oracle clustering error term that originates from the
	true latent cluster only. In contrast, the former component depends on the estimation error
	of $L$ and hence can be viewed as a noise-induced error term. Both terms are dependent on $c$ and $\omega$, which respectively quantify the degree of separation between the true cluster centers and the balance between communities.  Under the conditions of Theorem \ref{Thm.Main}, we can further obtain that 	$
	H(\hat{\boldsymbol{t}}, \pi(\boldsymbol{t})) \leq \frac{C\omega}{pc^2}\sum_{i=1}^p\|l_i^* - \mu^*_{\boldsymbol{t}(i)}\|^2 + O_P(n^{-1 + 4\eta})
	$. When $\{l_i^*\}$ has small variations within the same community, the Hamming error rate will be zero exactly with high probability.
\end{remark}
In general, the oracle clustering error term cannot be eliminated. However, when $m = r$ can be regarded as a prior, i.e., each community has one dimension, performing $K$-means on $\operatorname{Cor}(\operatorname{abs}(\widehat{L}))$ will be more suitable than on $\widehat{L}$ directly. Denote the  labels estimated from the clustering procedure \eqref{eq.cluster} based on $\operatorname{Cor}(\operatorname{abs}(\widehat{L}))$ by $\tilde{\bt}$. The following corollary reveals the rationale behind this transformation.
\begin{corollary}\label{Cor.CorHamming}
    When $m = r$, denote $U_{1i}^* = \left(u_{i1}, \cdots, u_{id_i}\right)$ as the eigenvector of $L_i^*$.  For any $i\in\mathcal{C}^*_k, j\in \mathcal{C}^*_l, k,l = 1,\cdots, m$, we have 
\begin{equation*}
		\operatorname{cor}(\operatorname{abs}(l_i^*), \operatorname{abs}(l_j^*))=
	\begin{cases} 
		 \frac{1}{d_k} - \bar{u}_k^2 & \text{if } k = l \\
		0 & \text{if } k \not= l
	\end{cases}
\end{equation*}
with $\bar{u}_k = \frac{1}{d_k}\sum_{h = 1}^{d_k}|u_{ih}|$. Furthermore, if $\min_{k\not= l}\|\frac{1}{d_k}  - \frac{1}{d_l} - \bar{u}_k^2  + \bar{u}_l^2\| >0$ and $d_{j} \in [\frac{p}{\omega m}, \frac{\omega p}{m}], j=1, \cdots, m $, then under conditions of Theorem \ref{Thm.Main}, 
	$$
    \lim_{n\to\infty}\mathbb{P}\left( H(\tilde{\boldsymbol{t}}, \pi(\boldsymbol{t})) = 0 \right)= 1.
    $$
\end{corollary}
This corollary is a direct application of Theorem \ref{Thm.Hammingerror} and Theorem \ref{Thm.Main} by noting that the oracle clustering error term is zero.

\section{Simulation study}\label{Sec.Simulation}
We conduct simulation studies to investigate the performance of the proposed method. We present the results for the adaptive $\ell_1$ penalized estimation in the second stage and $K$-means clustering in the third stage, respectively.
\subsection{Second stage: adaptive $\ell_1$ penalized estimation}\label{Sec.Simu_secstage}
We investigate the performance of the adaptive $\ell_1$ penalized estimation method in the second stage from two perspectives described in Section \ref{Sec.GraphModel}. Throughout this section, we fix $p = 45, r = 3$ and take $n = 1000, 2000, 4000, 8000$. For both graphical models with non-overlapped latent communities and grouped latent variables, we first generate two covariates $C_1, C_2$ ($q = 2$) with $C_1, C_2\sim N(0,1)$ independently and let $B_{ij} \sim \operatorname{Uniform}(0.5, 1)$, for $i = 1, \cdots, q$, and $j = 1,\cdots, p$. Then for the precision matrix of $R$, we consider the following two scenarios separately. 
\begin{enumerate}
    \item  \textbf{Latent community graphical models}: 
    For non-overlapped community graph $L$ in Section \ref{Sec.GMLC}, we take $m =2$, i.e., there exist two latent communities: the first community is of rank 2 and the second is of rank 1. Let nodes $N_1 = \{1, 2, \cdots, 25\}$  and $N_2 = \{26, \cdots, 45\}$ belong to the first community and the second community respectively, i.e., we suppose $\boldsymbol{a}_{i} = (\boldsymbol{a}_{i, N_1}, \boldsymbol{a}_{i, N_2}), i = 1, 2, 3$. In this example, we set $\boldsymbol{a}_{1,N_1} =  3\boldsymbol{u}_1, \boldsymbol{a}_{2, N_1} = 2.5\boldsymbol{u}_2, a_{3,N_1}= 0 $  and $\boldsymbol{a}_{1,N_2}= 0, \boldsymbol{a}_{2,N_2} =0, \boldsymbol{a}_{3,N_2} = 2\boldsymbol{u}_3$, where $\boldsymbol{u}_1, \boldsymbol{u}_2\in\mathbb{R}^{25}, \boldsymbol{u}_3 \in \mathbb{R}^{20}$ are unit vectors and $\boldsymbol{u}_1,\boldsymbol{u}_2$ are orthogonal to each other. For the sparse structure $S$, we set $s_{ii} = 5$, for $i = 1,\cdots p$. To record the sparse connections between the two communities, we set 
    $$s_{ji} = s_{ij} \sim \operatorname{Bernoulli}(\pi)*\operatorname{Uniform}\left((-2,-1.5)\cup(1.5, 2)\right),$$ 
    independently, where $\pi = 0.01$, for $i\in N_1, j\in N_2$. In addition, to {record} information within communities, we set $s_{(i+2), i} = s_{i,(i+2)} \sim  \operatorname{Uniform}\left((-2,-1.5)\cup(1.5, 2)\right), i = \{26, \cdots, 35\}$.
    \item \textbf{Grouped latent variable graphical models}: Suppose $m=3$, i.e. there are three groups of latent variables, each of which has only one latent variable $H_i$, for $i=1,2,3$. To meet the conditional independence assumption  \eqref{A.Condidepen}, we let nodes $N_1 = \{1, 2, \cdots, 15\}$, $N_2 = \{16, \cdots,30\} $  and $N_3 = \{31, \cdots, 45\}$, and suppose $(\Theta_{O,H_1})_{N_1} = 3.5\boldsymbol{u}_{1}, (\Theta_{O,H_2})_{N_2} = 3\boldsymbol{u}_{2},  (\Theta_{O,H_3})_{N_2} = 2.5\boldsymbol{u}_{3}$ and  $(\Theta_{O,H_1})_{N_2\cup N_3} = 0, (\Theta_{O,H_2})_{N_1\cup N_3} = 0, (\Theta_{O,H_3})_{N_1\cup N_2} = 0$, where $\boldsymbol{u}_{1} \in \mathbb{R}^{15}, \boldsymbol{u}_{2} \in \mathbb{R}^{15}, \boldsymbol{u}_{3} \in \mathbb{R}^{15}$ are three unit vectors. Denote $\Theta_O = (\theta^O_{ij})$, and $\Theta_H = (\theta^H_{ij})$. For $\Theta_O$, we generate $\theta^O_{i, (i+2)} = \theta^O_{(i+2),i } \sim \operatorname{Uniform}\left((-2,-1.5)\cup(1.5, 2)\right)$  independently within $N_1$, $i = 1,\cdots, 15 $. For the diagonal, we set $\theta_{ii}^O = 5, i = 1, \cdots p$ and $\theta^H_{jj} = 3$, $j= 1,\cdots, r$.
\end{enumerate}

\par The proposed method will be compared to three other competing methods. 
First, we obtain the estimator by solving \eqref{eq.InitialEsti}, where we adopt the BIC criterion used in \cite{chen2016fused} to select the tuning parameters. 
In what follows, we refer to it as the estimator from the latent variable Gaussian graphical model (LVGGM). 
%{The other two alternative methods are provided based on the LVGGM method and proposed method respectively.} 
Next, we extend the LVGGM method by adopting the hard thresholding criterion $\widehat{L}^{\operatorname{HT-LVGGM}} = \widehat{L}^{\operatorname{LVGGM}}\cdot1_{\{|\widehat{L}^{\operatorname{LVGGM}}|>a_n\}}$, and call it as the estimator from the hard-thresholding latent variable Gaussian graphical model
(HT-LVGGM). Note that the hard thresholding step will produce sparsity (we will set $a_n \asymp \frac{1}{\sqrt{n}}$ in practice), and we also set $\widehat{S}^{\operatorname{HT-LVGGM}}$ same as $\widehat{S}^{\operatorname{LVGGM}}$. Third, the nonadaptive penalized maximum likelihood estimation method (NonAPMLE) is a special case of the proposed adaptive $\ell_1$ penalized estimation, where we take $W_n = \textbf{1}_p\textbf{1}_p^T$.

 We evaluate the CV-based tuning parameter selection via five criteria. Let $(\widehat{S}, \widehat{L})$ be the estimates of the selected model defined as in \eqref{penobj}. Here $\operatorname{TR}_L$ stands for the proportion of true rank estimation of $L$ (equal to one is well),
$$
\operatorname{TR}_L = 1_{\left\{\operatorname{rank}(\widehat{L})=\operatorname{rank}(L^*)\right\}} .
$$
The criterion $\operatorname{TP}_L$ evaluates the proportion of true discoveries of non-zero elements in the network structure estimation of $L$ (the closer to one the better),
$$
 \operatorname{TP}_L= \frac{ |\{ (k,l): k\leq l, \hat{l}_{kl} \not = 0, \text{ and } l^*_{kl} \not= 0 \}|    }{|\{  (k,l): k\leq l, l^*_{kl} \not= 0 \} |}.
$$
We use $\operatorname{FP}_L$ to represent the proportion of false discoveries of zeros in off-diagonal blocks of $L$ (the closer to zero the better),
$$
\operatorname{FP}_L = \frac{ |\{ (k,l): k\leq l, \hat{l}_{kl} \not = 0, \text{ and } l^*_{kl} = 0 \}|    }{|\{  (k,l): k\leq l, l^*_{kl} = 0 \} |}.
$$
Similarly, $\operatorname{TP}_S$ evaluates the proportion of true discoveries of non-zero elements in the network structure estimation of $S$  (the closer to one the better),
$$
 \operatorname{TP}_S = \frac{\mid\left\{(i, j): i<j, \hat{s}_{i j} \neq 0, \text { and } s_{i j}^* \neq 0\right\} \mid}{\left|\left\{(i, j): i<j, s_{i j}^* \neq 0\right\}\right|} .
$$
The criterion $\operatorname{FP}_S$ evaluates the proportion of false discoveries of zeros in $S$ (the closer to zero the better),
$$
\operatorname{FP}_S = \frac{\mid\left\{(i, j): i<j, \hat{s}_{i j} \neq 0, \text { and } s_{i j}^*=0\right\} \mid}{\left|\left\{(i, j): i<j, s_{i j}^*=0\right\}\right|} .
$$

For each sample size, we generate 100 replications to obtain the mean and standard deviation of the five evaluation criteria above for LVGGM, HT-LVGGM, NonAPMLE as well as the proposed method. The results for graphical models with non-overlapped latent communities and grouped latent variables are presented in Tables \ref{Table.Latentcommunity} and \ref{Table.Latentvariable} respectively. 

From the numerical results, we see that for sufficiently large sample sizes, LVGGM, NonAPMLE, and the proposed method can accurately identify the rank of $L$. 
However, for the small sample size, NonAPMLE fails to identify rank, while the proposed method performs comparably well as LVGGM. Also, the proposed method efficiently identifies both non-zero blocks and zero blocks in $L$, indicating the community structure in the graph. In comparison, LVGGM and NonAPMLE struggle to recognize zero blocks. Though HT-LVGGM can identify the community structure, its accuracy in identifying the rank of $L$ is almost zero. Moreover, all four methods correctly identify the zero and non-zero elements in $S$ asymptotically, demonstrating their ability to correctly identify the sparse structure. When the sparse graph contains edges both between and within communities, the proposed method performs slightly worse compared to other methods, necessitating more samples to improve accuracy (cf. Table \ref{Table.Latentcommunity}). However, when the sparse graph has edges only within communities,
all four methods demonstrate nearly identical performance in accurately identifying all zero and non-zero elements (cf. Table \ref{Table.Latentvariable}).

\subsection{Third stage: $K$-means clustering}\label{Sec.Simu_thirdstage}

\par
\begin{figure}[htbp]
\begin{adjustwidth}{-0.8cm}{+1cm} % 调整页面左右边距
	\begin{subfigure}[b]{0.4\textwidth}
		\centering
		\includegraphics[width=3.5in]{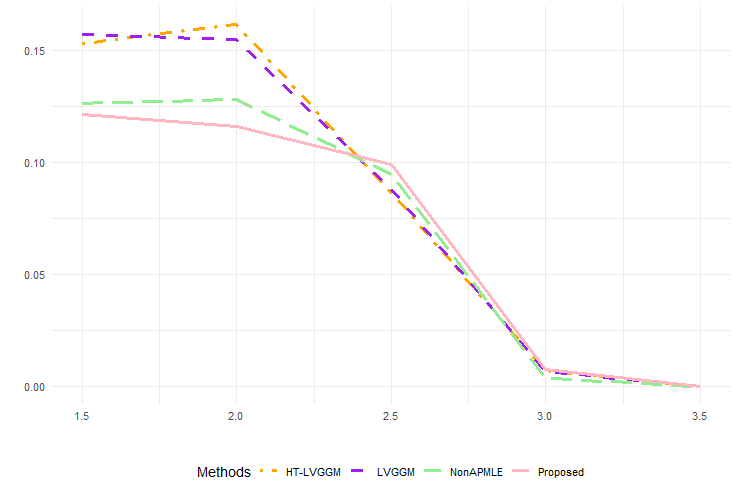}
	\end{subfigure}
	\hfill
	\begin{subfigure}[b]{0.4\textwidth}
		\centering
		\includegraphics[width=3.5in]{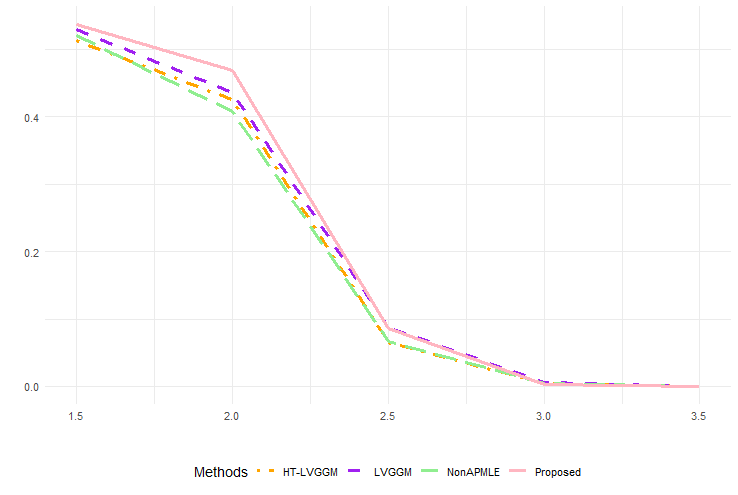}
	\end{subfigure}
\end{adjustwidth}
\caption{Comparison of Hamming error rates under different values of $a$ (representing the strengths of eigenvalues of latent community part $L$). Left panel: clustering based on $\widehat{L}$. Right panel: clustering based on $\operatorname{Cor}(\operatorname{abs}(\widehat{L}))$. $x$-axis: different values of a. $y$-axis: Hamming error rates.}
\label{Fig.Hammerror_a}
\end{figure}

\begin{figure}[htbp]
	\begin{adjustwidth}{-0.8cm}{+1cm} % 调整页面左右边距
		\begin{subfigure}[b]{0.4\textwidth}
			\centering
			\includegraphics[width=3.5in]{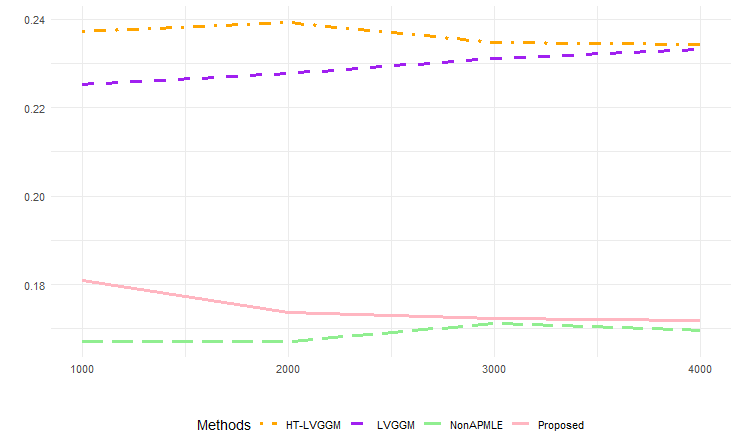}
		\end{subfigure}
		\hfill
		\begin{subfigure}[b]{0.4\textwidth}
			\centering
			\includegraphics[width=3.5in]{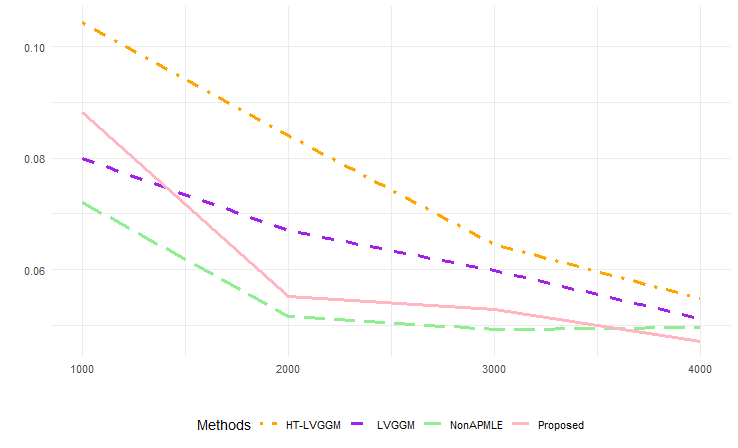}
		\end{subfigure}
	\end{adjustwidth}
	\caption{Comparison of Hamming error rates under different sample sizes. Left panel: clustering based on $\widehat{L}$. Right panel: clustering based on $\operatorname{Cor}(\operatorname{abs}(\widehat{L}))$. $x$-axis: different values of sample size. $y$-axis: Hamming error rates.}
	\label{Fig.Hammerror_samplesize}
\end{figure}

 Now we design experiments to evaluate the Hamming error rates of $K$-means procedure based on both $\widehat{L}$ and $\operatorname{Cor}(\operatorname{abs}(\widehat{L}))$ estimated from the four methods in the second stage. For illustration, we adopt the similar data generation mechanism designed in Section \ref{Sec.Simu_secstage} for grouped latent variable graphical models among two scenarios with the following specifications:
\begin{enumerate}
	\item  $(\Theta_{O,H_1})_{N_1} = a\boldsymbol{u}_{1}, (\Theta_{O,H_2})_{N_2} = (a-0.1)\boldsymbol{u}_{2},  (\Theta_{O,H_3})_{N_2} = (a - 0.2)\boldsymbol{u}_{3}$, where $u_{1i}, u_{2i}, u_{3i}\sim \operatorname{Uniform}(1,2) $ independently, and for comparison $a$ is taken as $ 1.5, 2.0, 2.5, 3.0, 3.5$ respectively. 
	These configurations reflect the maximum eigenvalues of the community part $L$ when the variation within each community is small. In this set of experiments, there are three clusters (communities), namely $N_1, N_2, N_3$. Moreover, for each value of $a$, 100 replications are generated to obtain the mean and standard deviations of Hamming error rates of $K$-means based on four methods.
	\item $(\Theta_{O,H_1})_{N_1} = 3.6\boldsymbol{u}_{1}, (\Theta_{O,H_2})_{N_2} = 3.3\boldsymbol{u}_{2},  (\Theta_{O,H_3})_{N_2} = 3\boldsymbol{u}_{3}$, $u_{1i}\sim \operatorname{Normal}(1,1), u_{2i}\sim \operatorname{Normal}(2,1), u_{3i}\sim \operatorname{Normal}(-1,1) $ independently. In this scenario, the variation within each community is larger than in the first scenario, and we set sample sizes as $1000, 2000, 3000, and 4000$ respectively to evaluate the performance of different clustering methods. For each sample size, 100 replications are generated to obtain the mean and standard deviations of Hamming error rates of $K$-means based on four methods.
\end{enumerate}
 To facilitate a consistent comparison across methods, we systematically excise the zero rows obtained from the proposed adaptive $\ell_1$ penalized estimation process before executing clustering for all four methods. The empirical findings for each of these methods, employing either $\widehat{L}$ or $\operatorname{Cor}(\operatorname{abs}(\widehat{L}))$, are comprehensively depicted in Tables \ref{Table.Hammerror_a}-\ref{Table.Hammerror_samplesize} and Figures \ref{Fig.Hammerror_a}-\ref{Fig.Hammerror_samplesize} for both scenarios.
\par In the first scenario, based on both $\widehat{L}$ and $\operatorname{Cor}(\operatorname{abs}(\widehat{L}))$, when the strength of the maximum eigenvalue of the community part $L$ is strong and the variation within each community is small, $K$-means under all four methods can correctly identify members within each community.  Moreover, as the parameter $a$ decreases, $K$-means clustering leveraging $\widehat{L}$ estimated by the proposed method attains the lowest Hamming error rate, thereby surpassing the alternative methods (cf. Table \ref{Table.Hammerror_a} and Figure \ref{Fig.Hammerror_a}). In contrast, the performance of $K$-means clustering remains largely consistent across all four methods when utilizing $\operatorname{Cor}(\operatorname{abs}(\widehat{L}))$, irrespective of the eigenvalue magnitudes of $L$. It is important to note, however, that in situations where the maximum eigenvalue of $L$ is relatively weak, the resultant clustering error tied to $\operatorname{Cor}(\operatorname{abs}(\widehat{L}))$ markedly exceeds that associated with $\widehat{L}$. This discrepancy can be attributed to the heightened sensitivity of $K$-means clustering to significant estimation errors in $\widehat{L}$ estimated from the second stage when $a$ is small.

\par In the second scenario, $K$-means based on $\widehat{L}$ estimated from the adaptive $\ell_1$ penalized estimation method, consistently yields reduced error margins with increasing sample sizes compared with the LVGGM method. Nevertheless, due to variations within each community, the oracle error delineated in Theorem \ref{Thm.Hammingerror} remains extant. In the implementation of the $K$-means algorithm that utilizes $\operatorname{Cor}(\operatorname{abs}(\widehat{L}))$, it is observed that the Hamming error rates for all investigated methods converge to zero asymptotically. Furthermore, the application of adaptive $\ell_1$ penalized estimation within this context continues to demonstrate diminished error rates concerning the LVGGM method (cf. Table \ref{Table.Hammerror_samplesize} and Figure \ref{Fig.Hammerror_samplesize}).

%Denote the clustering labels based on the four methods in the second stage as $\hat{\boldsymbol{t}}^{LVGGM}$, $\hat{\boldsymbol{t}}^{HT-LVGGM}$, $\hat{\boldsymbol{t}}^{NonAPMLE}$, $\hat{\boldsymbol{t}}^{Proposed}$. Then we compute the respective Hamming distance under 100 replications.

\newpage
\begin{table}[H]
\centering
%\captionsetup{font={\footnotesize}} % 设置标题字体为footnotesize
\setlength{\baselineskip}{\normalbaselineskip} % 使标题内部行间距与正文一致
\caption{The average and standard deviation of different evaluation criteria are reported for latent community graphical models. Two covariates ($q = 2$) are incorporated and the sample size is $n = 1000, 2000, 4000, 8000$ respectively. The rank $r$, number of communities $m$, and number of nodes $p$ are fixed as 3, 2, and 45 respectively. A 5-fold CV is adopted to select tuning parameters for NonAPMLE and the proposed method.  For each sample size, $100$ replications are performed. Criterion
 $\operatorname{TR}_L$ (the closer to one the better), $\operatorname{TP}_L$ (the closer to one the better), $\operatorname{FP}_L$ (the closer to zero the better), $\operatorname{TP}_S$ (the closer to one the better) and $\operatorname{FP}_S$ (the closer to zero the better) are used.   } 
\label{Table.Latentcommunity}
\vskip 0.15in
\renewcommand{\arraystretch}{0.90}
\begin{center}
\begin{small}
%	\begin{sc}
\begin{tabular}{cccccc}
	\toprule
	%\multicolumn{1}{c}{Sample Size}&\multicolumn{1}{c}{Eva}& \multicolumn{3}{c}{Methods} \\
	%\cmidrule(lr){1-1}
      %   \cmidrule(lr){2-2}
	%\cmidrule(lr){3-5} 
	Sample Size & Criteria  & LVGGM & HT-LVGGM & NonAPMLE & Proposed \\
	
	\midrule
	\multirow{6}*{\makecell[c]{$n = 1000$}}  
	~    & $\operatorname{TR}_L$ & 0.960 (0.197)  & 0.000 (0.000)  & 0.010 (0.100) & 0.960 (0.197) \\
	\cmidrule(lr){2-6}
	~    &$\operatorname{TP}_L$ & 1.000 (0.000) & 0.884 (0.036) & 1.000 (0.000) &   0.921 (0.040) \\
	\cmidrule(lr){2-6}
	~    &$\operatorname{FP}_L$ & 1.000 (0.000) & 0.018 (0.016) & 0.999 (0.002) &  0.053 (0.025) \\
	\cmidrule(lr){2-6}
	~	 & $\operatorname{TP}_S$ & 1.000 (0.000) & 1.000 (0.000) & 1.000 (0.000) &  1.000 (0.000)  \\
	\cmidrule(lr){2-6}
	~    & $\operatorname{FP}_S$ & 0.003 (0.003) & 0.003 (0.003) & 0.027 (0.005) & 0.045 (0.006)   \\
	\cmidrule(lr){1-6}
	
	\multirow{6}*{\makecell[c]{$n = 2000$}}  
	~    & $\operatorname{TR}_L$ &1.000 (0.000) & 0.000 (0.000) & 1.000 (0.000) & 1.000 (0.000) \\
	\cmidrule(lr){2-6}
	~    &$\operatorname{TP}_L$ & 1.000 (0.018) & 0.937 (0.000) & 1.000 (0.000)  &  0.930 (0.030)  \\
	\cmidrule(lr){2-6}
	~    &$\operatorname{FP}_L$ & 1.000 (0.000) & 0.011 (0.010) & 0.996 (0.003) &  0.020 (0.012) \\
	\cmidrule(lr){2-6}
	~	 &$\operatorname{TP}_S$ & 1.000 (0.000)  & 1.000 (0.000) & 1.000 (0.000)  &  1.000 (0.000) \\
	\cmidrule(lr){2-6}
	~    &$\operatorname{FP}_S$ & 0.000 (0.001) & 0.000 (0.001)  & 0.005 (0.002) & 0.014 (0.003) \\
	\cmidrule(lr){1-6}
	
    \multirow{6}*{\makecell[c]{$n = 4000$}}  
~    & $\operatorname{TR}_L$ &1.000 (0.000) & 0.000 (0.000) & 1.000 (0.000) & 1.000 (0.000) \\
\cmidrule(lr){2-6}
~    &$\operatorname{TP}_L$ & 1.000 (0.000) & 0.954 (0.010) & 1.000 (0.000)  &  0.965 (0.014)  \\
\cmidrule(lr){2-6}
~    &$\operatorname{FP}_L$ & 1.000 (0.000) & 0.003 (0.005) & 0.996 (0.003) &  0.015 (0.011) \\
\cmidrule(lr){2-6}
~	 &$\operatorname{TP}_S$ & 1.000 (0.000)  & 1.000 (0.000) & 1.000 (0.000)  &  1.000 (0.000) \\
\cmidrule(lr){2-6}
~    &$\operatorname{FP}_S$ & 0.000 (0.000) & 0.000 (0.000)  & 0.001 (0.001) & 0.003 (0.001) \\
\cmidrule(lr){1-6}
	
    \multirow{6}*{\makecell[c]{$n = 8000$}}  
~    & $\operatorname{TR}_L$ &1.000 (0.000) & 0.000 (0.000) & 1.000 (0.000) & 1.000 (0.000) \\
\cmidrule(lr){2-6}
~    &$\operatorname{TP}_L$ & 1.000 (0.000) & 0.960 (0.010) & 1.000 (0.000)  &  0.984 (0.011)  \\
\cmidrule(lr){2-6}
~    &$\operatorname{FP}_L$ & 1.000 (0.000) & 0.001 (0.002) & 0.997 (0.003) &  0.014 (0.010) \\
\cmidrule(lr){2-6}
~	 &$\operatorname{TP}_S$ & 1.000 (0.000)  & 1.000 (0.000) & 1.000 (0.000)  &  1.000 (0.000) \\
\cmidrule(lr){2-6}
~    &$\operatorname{FP}_S$ & 0.000 (0.000) & 0.000 (0.000)  & 0.001 (0.001) & 0.001 (0.001) \\
\cmidrule(lr){1-6}
	
	\bottomrule
\end{tabular}
%	\end{sc}
\end{small}
\end{center}
\vskip -0.1in
\end{table}

\newpage

\begin{table}[H]
\centering
\caption{The average and standard deviation of different evaluation criteria are reported for the grouped latent variable graphical models. Two covariates ($q = 2$) are incorporated and the sample size is $n = 1000, 2000, 4000, 8000$ respectively. The number of latent variables $r$, number of communities $m$, and number of observed variables $p$ are fixed as 3, 3, and 45 respectively. A 5-fold CV is adopted to select tuning parameters for NonAPMLE and the proposed method.  For each sample size, $100$ replications are performed. Criterion
 $\operatorname{TR}_L$ (the closer to one the better), $\operatorname{TP}_L$ (the closer to one the better), $\operatorname{FP}_L$ (the closer to zero the better), $\operatorname{TP}_S$ (the closer to one the better) and $\operatorname{FP}_S$ (the closer to zero the better) are used.  } 
\label{Table.Latentvariable}
\vskip 0.15in
\renewcommand{\arraystretch}{0.90}
\begin{center}
\begin{small}
%	\begin{sc}
\begin{tabular}{cccccc}
	\toprule
	%\multicolumn{1}{c}{Sample Size}&\multicolumn{1}{c}{Eva}& \multicolumn{3}{c}{Methods} \\
	%\cmidrule(lr){1-1}
      %   \cmidrule(lr){2-2}
	%\cmidrule(lr){3-5} 
	Sample Size & Criteria  & LVGGM & HT-LVGGM & NonAPMLE & Proposed \\
	
	\midrule
	\multirow{6}*{\makecell[c]{$n = 1000$}}  
	~    &$\operatorname{TR}_L$ & 1.000 (0.000)  & 0.000 (0.000)  & 0.040 (0.197) & 0.910 (0.288) \\
	\cmidrule(lr){2-6}
	~    &$\operatorname{TP}_L$ & 1.000 (0.000) & 0.805 (0.043) & 1.000 (0.001) &   0.884 (0.057) \\
	\cmidrule(lr){2-6}
	~    &$\operatorname{FP}_L$ & 1.000 (0.000) & 0.021 (0.012) & 0.995 (0.003) &  0.086 (0.055) \\
	\cmidrule(lr){2-6}
	~	 &$\operatorname{TP}_S$ & 0.999 (0.013) & 0.999 (0.013) & 1.000 (0.000) &  1.000 (0.000)  \\
	\cmidrule(lr){2-6}
	~    &$\operatorname{FP}_S$ & 0.001 (0.004) & 0.001 (0.004) & 0.000 (0.000) & 0.000 (0.000)   \\
	\cmidrule(lr){1-6}
	
	\multirow{6}*{\makecell[c]{$n = 2000$}}  
	~    &$\operatorname{TR}_L$ &1.000 (0.000) & 0.000 (0.000) & 0.440 (0.499) & 0.910 (0.288) \\
	\cmidrule(lr){2-6}
	~    &$\operatorname{TP}_L$ & 1.000 (0.000) & 0.825  (0.041) & 1.000 (0.001)  &  0.928 (0.038)  \\
	\cmidrule(lr){2-6}
	~    &$\operatorname{FP}_L$ & 1.000 (0.000) & 0.011 (0.001) & 0.996 (0.002) &  0.078 (0.041) \\
	\cmidrule(lr){2-6}
	~	 &$\operatorname{TP}_S$ & 1.000 (0.000)  & 1.000 (0.000) & 1.000 (0.000)  &  1.000 (0.000) \\
	\cmidrule(lr){2-6}
	~    &$\operatorname{FP}_S$ & 0.000 (0.000) & 0.000 (0.000)  & 0.000 (0.000) & 0.000 (0.000) \\
	\cmidrule(lr){1-6}
	
	\multirow{6}*{\makecell[c]{$n = 4000$}}  
	~    &$\operatorname{TR}_L$ &1.000 (0.000) & 0.000 (0.000) & 1.000 (0.000) & 0.990 (0.100) \\
	\cmidrule(lr){2-6}
	~    &$\operatorname{TP}_L$ & 1.000 (0.000) & 0.849  (0.031) & 1.000 (0.000)  &  0.952 (0.031)  \\
	\cmidrule(lr){2-6}
	~    &$\operatorname{FP}_L$ & 1.000 (0.000) & 0.008 (0.006) & 0.997 (0.002) &  0.068 (0.029) \\
	\cmidrule(lr){2-6}
	~	 &$\operatorname{TP}_S$ & 1.000 (0.000)  & 1.000 (0.000) & 1.000 (0.000)  &  1.000 (0.000) \\
	\cmidrule(lr){2-6}
	~    &$\operatorname{FP}_S$ & 0.000 (0.000) & 0.000 (0.000)  & 0.000 (0.000) & 0.000 (0.000) \\
	\cmidrule(lr){1-6}
	
	\multirow{6}*{\makecell[c]{$n = 8000$}}  
	~    &$\operatorname{TR}_L$ &1.000 (0.000) & 0.000 (0.000) & 1.000 (0.000) & 1.000 (0.000) \\
	\cmidrule(lr){2-6}
	~    &$\operatorname{TP}_L$ & 1.000 (0.000) & 0.864 (0.024) & 1.000 (0.000)  &  0.963 (0.028)  \\
	\cmidrule(lr){2-6}
	~    &$\operatorname{FP}_L$ & 1.000 (0.000) & 0.006 (0.004) & 0.996 (0.003) &  0.057 (0.016) \\
	\cmidrule(lr){2-6}
	~	 &$\operatorname{TP}_S$ & 1.000 (0.000)  & 1.000 (0.000) & 1.000 (0.000)  &  1.000 (0.000) \\
	\cmidrule(lr){2-6}
	~    &$\operatorname{FP}_S$ & 0.000 (0.000) & 0.000 (0.000)  & 0.000 (0.000) & 0.000 (0.000) \\
	\cmidrule(lr){1-6}
	\bottomrule
\end{tabular}
%	\end{sc}
\end{small}
\end{center}
\vskip -0.1in
\end{table}

%\newpage
%\begin{table}[ht]
%	\centering
%	\caption{Hamming error rates and standard deviations for different values of $a$ (representing the strengths of eigenvalues of latent community part $L$) are reported for the graphical models with non-overlapped grouped latent variables. Two covariates are incorporated and the sample size is fixed as $n = 1000$. The number of latent variables $r$, number of communities $m$, and number of observed variables $p$ are fixed as 3, 3, and 45 respectively.  For each value of $a$, 100 replications are performed.}
%	\label{tab:mytable}
%	\vskip 0.15in
%	\renewcommand{\arraystretch}{1.5}
%	\begin{tabular}{ccccc}
%		\hline
%		$a$ & LVGGM & HT-LVGGM & NonAPMLE & Proposed \\
%		\hline
%	       1.500   & 0.149 (0.071) & 0.181 (0.067) & 0.116 (0.038) & 0.121 (0.029) \\
%		\hline
%		2.000 & 0.148 (0.070) & 0.178 (0.076) & 0.126 (0.045) & 0.116 (0.028)\\
%		\hline
%	    2.500 & 0.090 (0.050)& 0.091 (0.060) & 0.095 (0.043) & 0.099 (0.042) \\
%		\hline
%		3.000& 0.007 (0.015) & 0.009 (0.016) & 0.004 (0.010) & 0.008 (0.016) \\
%		\hline
%		3.500 & 0.000 (0.000)& 0.000 (0.000) & 0.000 (0.000) &0.000 (0.000)\\
%		\hline
%		\bottomrule
%	\end{tabular}
%\end{table}
%
%
%\begin{figure}[H]
%	\centering
%	\includegraphics[width=5.5in]{hammerror.png}
%	\caption{ Comparison of Hamming errors. $x$-axis: $a$ representing the strengths of eigenvalues of latent community part $L$. $y$-axis: Hamming errors.}
%	\label{Figure.Hammingerror}
%\end{figure}

\newpage
\begin{table}[H]
	\centering
	%\captionsetup{font={\footnotesize}} % 设置标题字体为footnotesize
	\setlength{\baselineskip}{\normalbaselineskip} % 使标题内部行间距与正文一致
	\caption{When clustering based on $\widehat{L}$ and $\operatorname{Cor}(\operatorname{abs}(\widehat{L}))$, the mean and standard deviation for Hamming error rates under different values of $a$ (representing the strengths of eigenvalues of latent community part $L$) are reported for the grouped latent variable graphical models. Two covariates are incorporated and the sample size is fixed as $n = 1000$. The number of latent variables $r$, number of communities $m$, and number of observed variables $p$ are fixed as 3, 3, and 45 respectively.  For each value of $a$, 100 replications are performed.  } 
	\label{Table.Hammerror_a}
	\vskip 0.15in
	\renewcommand{\arraystretch}{0.90}
	\begin{center}
		\begin{small}
			%	\begin{sc}
				\begin{tabular}{cccccc}
					\toprule
					%\multicolumn{1}{c}{Sample Size}&\multicolumn{1}{c}{Eva}& \multicolumn{3}{c}{Methods} \\
					%\cmidrule(lr){1-1}
					%   \cmidrule(lr){2-2}
					%\cmidrule(lr){3-5} 
					Clustering based	 & a & LVGGM & HT-LVGGM & NonAPMLE & Proposed \\
					
					\midrule
					\multirow{6}*{\makecell[c]{$\widehat{L}$}}  
					~& 1.5   & 0.157 (0.071) & 0.153 (0.064) & 0.126 (0.043) & 0.121 (0.029) \\
				\cmidrule(lr){2-6}
					~& 2.0 & 0.155 (0.071) & 0.161 (0.071) & 0.128 (0.045) & 0.116 (0.028)\\
				\cmidrule(lr){2-6}
					~& 2.5 & 0.088 (0.049)& 0.086 (0.058) & 0.095 (0.041) & 0.099 (0.042) \\
					\cmidrule(lr){2-6}
					~& 3.0& 0.007 (0.015) & 0.007 (0.015) & 0.004 (0.010) & 0.008 (0.016) \\
				\cmidrule(lr){2-6}
					~& 3.5 & 0.000 (0.000)& 0.000 (0.000) & 0.000 (0.000) &0.000 (0.000)\\
					\cmidrule(lr){1-6}
					
					\multirow{6}*{\makecell[c]{$\operatorname{Cor}(\operatorname{abs}(\widehat{L}))$}}  
				   ~& 1.5   & 0.530 (0.062) & 0.513 (0.059) & 0.521 (0.058) & 0.537 (0.060) \\
			    \cmidrule(lr){2-6}
				    ~& 2.0 & 0.436 (0.088) & 0.426 (0.096) & 0.408 (0.091) & 0.469 (0.078)\\
				  \cmidrule(lr){2-6}
				   ~&  2.5 & 0.087 (0.080)& 0.065 (0.078) & 0.067 (0.044) & 0.086 (0.075) \\
				   \cmidrule(lr){2-6}
				   ~&  3.0& 0.005 (0.011) & 0.004 (0.010) & 0.004 (0.009) & 0.003 (0.007) \\
				  \cmidrule(lr){2-6}
				   ~&  3.5 & 0.000 (0.000)& 0.000 (0.000) & 0.000 (0.000) &0.000 (0.000)\\
					\cmidrule(lr){1-6}				
					\bottomrule
				\end{tabular}
				%	\end{sc}
		\end{small}
	\end{center}
	\vskip -0.1in
\end{table}

\newpage
\begin{table}[H]
	\centering
	%\captionsetup{font={\footnotesize}} % 设置标题字体为footnotesize
	\setlength{\baselineskip}{\normalbaselineskip} % 使标题内部行间距与正文一致
	\caption{When clustering based on $\widehat{L}$ and $\operatorname{Cor}(\operatorname{abs}(\widehat{L}))$, the mean and standard deviation for Hamming error rates under different sample size are reported for the grouped latent variable graphical models. Two covariates are incorporated and the sample size is set as $ 1000, 2000, 3000, 4000$ respectively. The number of latent variables $r$, number of communities $m$, and number of observed variables $p$ are fixed as 3, 3, and 45 respectively.  For each sample size, 100 replications are performed.   } 
	\label{Table.Hammerror_samplesize}
	\vskip 0.15in
	\renewcommand{\arraystretch}{0.90}
	\begin{center}
		\begin{small}
			%	\begin{sc}
				\begin{tabular}{cccccc}
					\toprule
					%\multicolumn{1}{c}{Sample Size}&\multicolumn{1}{c}{Eva}& \multicolumn{3}{c}{Methods} \\
					%\cmidrule(lr){1-1}
					%   \cmidrule(lr){2-2}
					%\cmidrule(lr){3-5} 
				 Clustering based	 & Sample size $n$  & LVGGM & HT-LVGGM & NonAPMLE & Proposed \\
					
					\midrule
					\multirow{6}*{\makecell[c]{$\widehat{L}$}}  
                    ~&	$1000$   & 0.225 (0.027) & 0.237 (0.020) & 0.167 (0.031) & 0.181 (0.031) \\
                	\cmidrule(lr){2-6}
                   ~&$2000$ & 0.228 (0.022) & 0.239 (0.014) & 0.167 (0.029) & 0.174 (0.033)\\
	                \cmidrule(lr){2-6}
                    ~&$3000$ & 0.231 (0.017)& 0.235 (0.015) & 0.171 (0.025) & 0.172 (0.025) \\
                	\cmidrule(lr){2-6}
                   ~&$4000$& 0.233 (0.015) & 0.234 (0.014) & 0.170 (0.021) & 0.172 (0.023) \\
					\cmidrule(lr){1-6}
					
					\multirow{6}*{\makecell[c]{$\operatorname{Cor}(\operatorname{abs}(\widehat{L}))$}}  
					~& $1000$   & 0.080 (0.032) & 0.104 (0.033) & 0.072 (0.034) & 0.088 (0.044) \\
	     	        \cmidrule(lr){2-6}
		            ~&	$2000$ & 0.067 (0.030) & 0.084 (0.030) & 0.052 (0.027) & 0.055 (0.028)\\
		            \cmidrule(lr){2-6}
	             	~&	$3000$ & 0.060 (0.030)& 0.064 (0.030) & 0.049 (0.025) & 0.053 (0.028) \\
	            	\cmidrule(lr){2-6}
	             	~&	$4000$& 0.051 (0.029) & 0.055 (0.028)  & 0.050 (0.023) & 0.047 (0.024) \\
			     	\cmidrule(lr){1-6}					
					\bottomrule
				\end{tabular}
				%	\end{sc}
		\end{small}
	\end{center}
	\vskip -0.1in
\end{table}

\newpage
\section{Real data application}\label{Sec.Realdata}
We apply the proposed method to stock return data. 
%Exploring the underlying community structure for stocks contributes to a better understanding of the sources of stock price correlations, thereby assisting users in effectively mitigating risks.
To proceed, we have selected 45 stocks from the energy, financial, and healthcare sectors, each with a substantial market size (15 stocks for each sector). This choice is informed by their associations with public health and safety, environmental protection, and societal stability. We collect their adjusted closing prices between January 2014 and December 2023. These data are publicly available on the Yahoo Finance website and are bundled in standard software packages such as the R package \verb6quantmod6. %For each period, we initially calculate the daily returns for each stock based on the adjusted closing prices. 
To eliminate possible serial dependence, we compute the stock returns based on prices from alternating days, which lead to a sample of $n = 1253$ observations for each stock. In addition, to model the market effect, we use the S\&P 500 index as our covariate. Then in the second stage, we apply the proposed adaptive $\ell_1$ penalized estimation method to the grouped latent variable graphical models defined in Section \ref{Sec.GMGLV} to learn the sparse graph $S$ and the non-overlapped latent community graph $L$. 
For comparison, we use LVGGM as the benchmark model.

\par
\begin{figure}[H]

	\begin{minipage}{\textwidth}
		\centering		
		\begin{subfigure}[b]{0.8\textwidth}
			\includegraphics[width=4.5in]{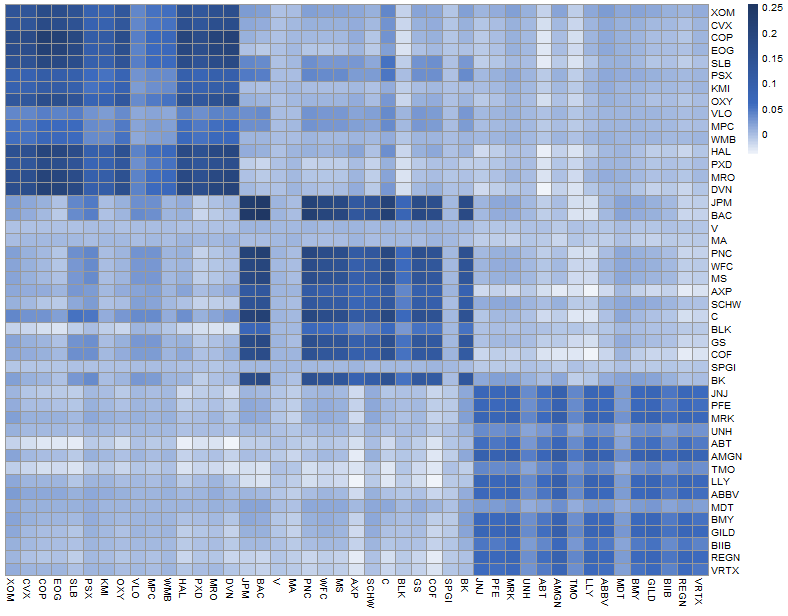}
			\caption{Heatmap of latent community graph estimated by LVGGM method}
		\end{subfigure}
		
		\vspace{1.5cm} % 调整两个子图之间的垂直间距
		\begin{subfigure}[b]{0.8\textwidth}
			\includegraphics[width=4.5in]{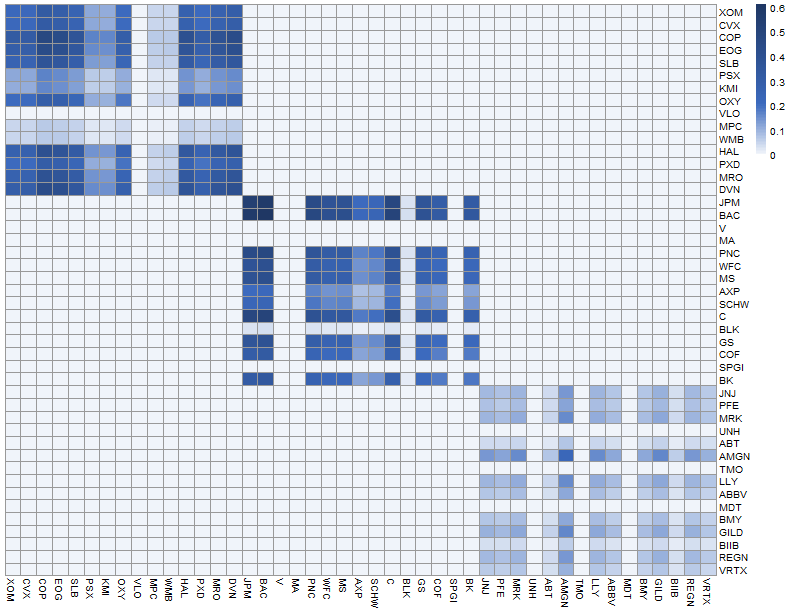}
			\caption{Heatmap of latent community graph estimated by proposed method}
		\end{subfigure}
	\end{minipage}	
	\caption{Heatmaps of latent community graph estimated by LVGGM method and proposed method respectively. The top 15 companies on the vertical axis are from the energy sector, the middle 15 are from the financial sector, and the bottom 15 are from the healthcare sector.}
	\label{Figure.Heatmap.L}
\end{figure}

 
After accounting for the market effect revealed by the S\&P 500 index, both LVGGM and the proposed method estimate the rank of the non-overlapped latent community graph $L$ as $r = 3$, indicating three latent variables. This coincides with the three different sectors we have determined beforehand. Additionally, Figure \ref{Figure.Heatmap.L} depicts the heatmaps of $L$ estimated by both methods. From the perspective of community detection, the LVGGM method struggles to accurately identify the number of communities, indicating that it does not fully reveal the inter-dependencies among different sectors. In contrast, the proposed method significantly identifies the independent relationships among the three sectors. 

\begin{figure}[H]
	\begin{minipage}{\textwidth}
		\centering	
		\begin{subfigure}[b]{1\textwidth}
			\includegraphics[width=5in]{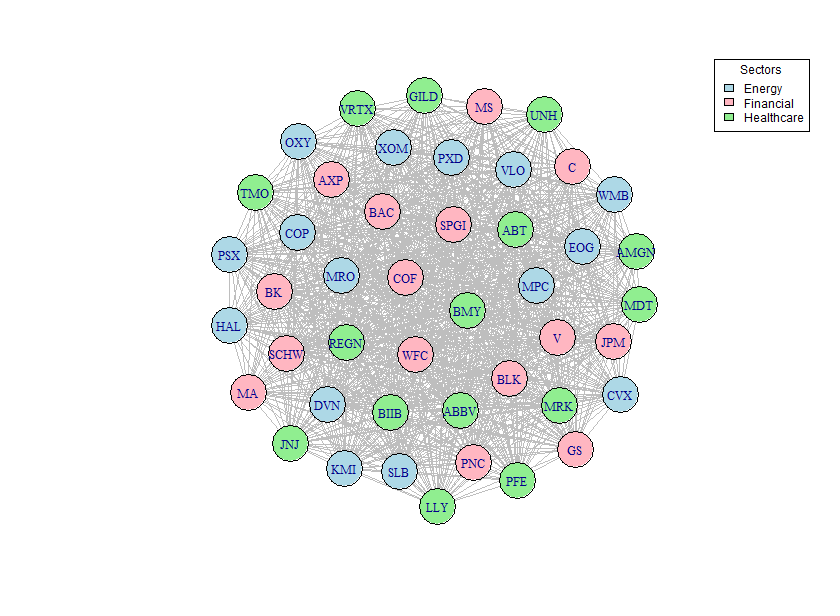}
			\caption{Latent community graph estimated by LVGGM method}
		\end{subfigure}
		
		\vspace{0.5cm} % 调整两个子图之间的垂直间距
		\begin{subfigure}[b]{1\textwidth}
			\includegraphics[width=5in]{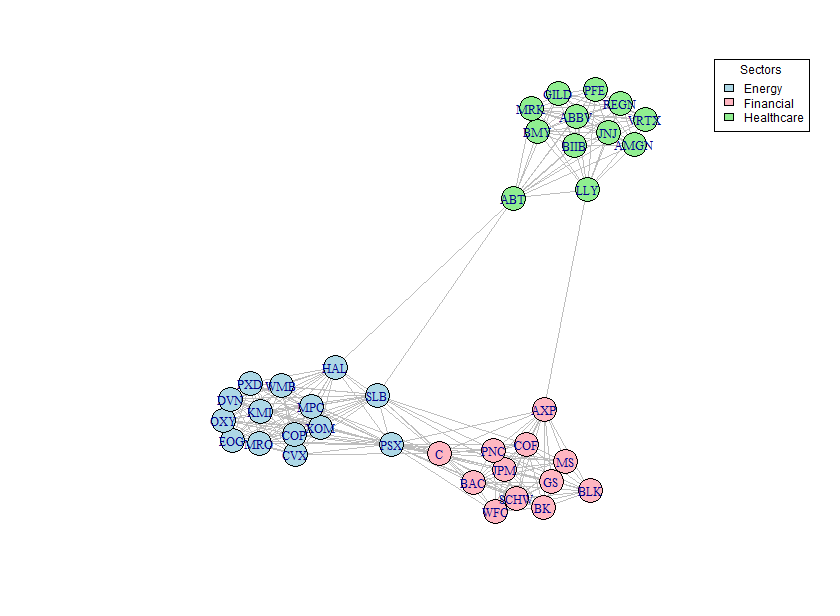}
			\caption{Latent community graph estimated by proposed method}
		\end{subfigure}
	\end{minipage}
	
	\caption{Latent community graphs estimated by LVGGM method and proposed method respectively. Green represents the healthcare sector, pink represents the financial sector, and blue represents the energy sector.}
		\label{Figure.Connectivity.L}
\end{figure}

Figure \ref{Figure.Connectivity.L} shows the connectivity graphs for $L$ estimated by both methods.
The graph estimated by LVGGM is almost fully connected, whereas the proposed method demonstrates the independence among the three sectors (green represents the healthcare sector, pink represents the financial sector, and blue represents the energy sector). Specifically, combining Figures~\ref{Figure.Heatmap.L}(b) and \ref{Figure.Connectivity.L}(b), we find a few weak edges between the energy sector and the financial sector. We speculate that with an increase in sample size, these edges may disappear. In addition, the healthcare sector exhibits almost no edges connecting it to the energy and financial sectors. This indicates that, after removing the influence of the overall market effect, the development of these three sectors is relatively independent. 

Due to the presence of small values in the diagonal blocks of the initial estimator $\widehat{L}^{LVGGM}$, the corresponding diagonal blocks in the estimated latent community graph obtained from the proposed method are compressed to 0 (cf. Figure~\ref{Figure.Heatmap.L}(b)). This explains the absence of the following nodes in Figure~\ref{Figure.Connectivity.L}(b): (1) VLO (Valero Energy Corporation) in the energy sector; (2) V (Visa Inc.), MA (Mastercard Incorporated), and SPGI (S\&P Global Inc.) companies in the financial sector; and (3) UNH (UnitedHealth Group Incorporated), TMO (Thermo Fisher Scientific Inc.), and MDT (Medtronic plc) companies in the healthcare sector. Such absence suggests 
the weak impact of the economic sectors on the corresponding stock prices, and for a finite sample these influences may be compressed to zero.

%that the energy sector has a minimal impact on the stock price of VLO, the financial sector has a weak influence on the stock prices of V, MA, and SPGI, and the healthcare sector has a limited impact on the stock prices of UNH, TMO, and MDT. In a limited sample, these influences are compressed to zero.

\begin{figure}[H]
	\begin{minipage}{\textwidth}
		\centering		
		\begin{subfigure}[b]{0.8\textwidth}
			\includegraphics[width=4.5in]{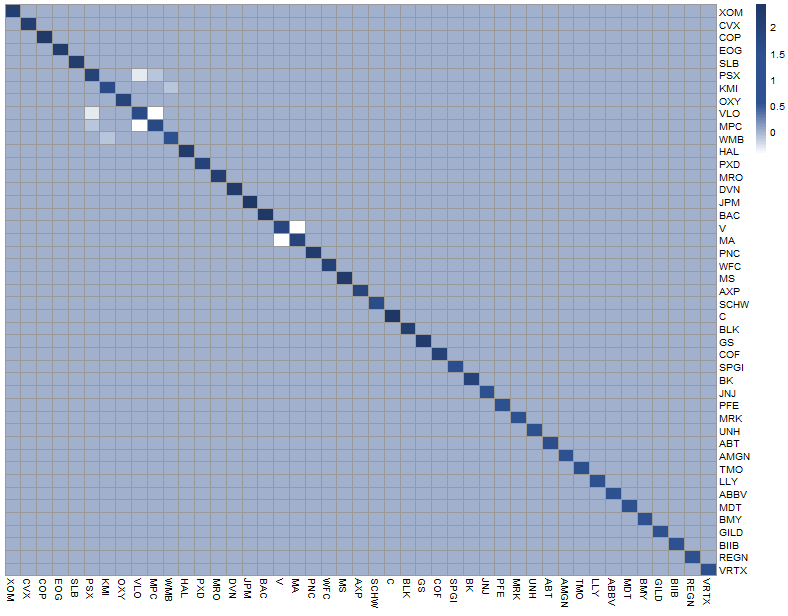}
			\caption{Heatmap of sparse graph estimated by proposed method}
		\end{subfigure}
		
		\vspace{1.5cm} % 调整两个子图之间的垂直间距
		\begin{subfigure}[b]{0.8\textwidth}
			\includegraphics[width=4.5in]{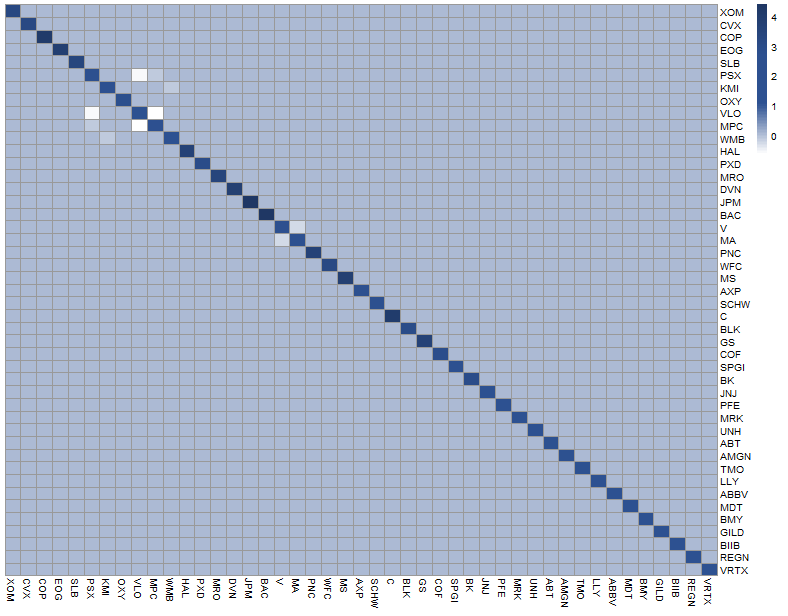}
			\caption{Heatmap of sparse graph estimated by proposed method}
		\end{subfigure}
	\end{minipage}
	
	\caption{Heatmaps of sparse graph estimated by LVGGM method and proposed method respectively. The top 15 companies on the vertical axis are from the energy sector, the middle 15 are from the financial sector, and the bottom 15 are from the healthcare sector.}
		\label{Figure.Heatmap.S}
\end{figure}

\begin{figure}[H]
		\centering
			\includegraphics[width=5.5in]{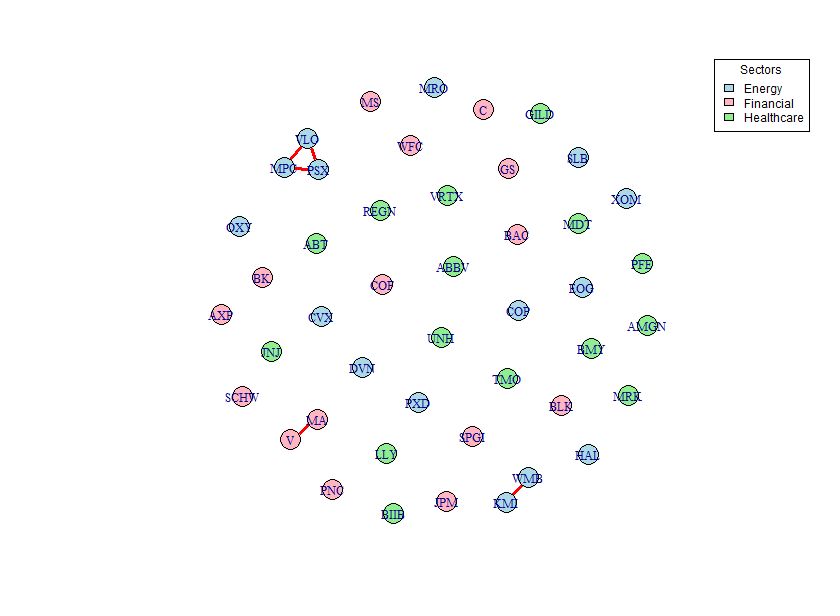}
			\caption{ Sparse graph estimated by both LVGGM and the proposed method. Green represents the healthcare sector, pink represents the financial sector, and blue represents the energy sector.}
	\label{Figure.Connectivity.S}
\end{figure}

With regard to the estimation of the sparse graph $S$, we observe from Figures \ref{Figure.Heatmap.S} and \ref{Figure.Connectivity.S} that both methods identify similar values of correlations, aligning with the results from simulation studies. In particular, Table~\ref{Table.Latentvariable} shows both LVGGM and the proposed method can accurately identify zero and non-zero elements of $S$ when the sample size is around $1000$, and $S$ contains only a small number of edges within the community. For instance, PSX (Phillips 66), VLO, and MPC (Marathon Petroleum Corporation) are among the major companies in the energy and petroleum industry and are involved in the downstream sector of the oil and gas industry, especially for refining and marketing. 
Due to their important roles in the production and distribution of petroleum products,
even after removing the impact of the overall market and sectors, these three companies remain closely interconnected as is shown in Figure~\ref{Figure.Connectivity.S}. 
A similar interpretation applies to 
the relationship between KMI (Kinder Morgan, Inc.) and WMB (Williams Companies, Inc.) as well. Also, as mentioned earlier, the financial services sector has little influence on the stock prices of V and MA. However, as observed from Figures \ref{Figure.Heatmap.S} and \ref{Figure.Connectivity.S}, they still exhibit a strong correlation. This might be attributed to the fact that both companies are engaged in payment services, sharing similar business profiles and investor groups.
%\par \sd{Finally, in the third stage, we perfomr $K$-means clustering based on the proposed method and LVGGM method. Due to the nature of this dataset, we have prior knowledge of the sectors to which each stock belongs. Therefore, we assume that the true labels and the number of communities are known, i.e., $m=3$. Additionally, since the proposed method compresses 7 rows in $\widehat{L}$ entirely to 0, we ultimately perform clustering only on the remaining 38 stocks for both the proposed method and the LVGGM method. When clustering is performed based on $\widehat{L}$, the Hamming error rate for LVGGM is 0.079, and the Hamming error rate for the proposed method is 0.132. Similarly, when clustering is based on $\operatorname{Corr}(\widehat{L})$, both LVGGM and the proposed method have the same Hamming error rate of 0. 
%Comparatively, at this point, $K$-means clustering based on $\operatorname{Corre}(\widehat{L})$ yields better results. In fact, the eigenvalues of $\widehat{L}$ obtained by the proposed method and LVGGM method are $(3.268, 3.206, 1.004), (1.900, 1.672, 0.872)$ respectively. At this moment, all three eigenvalue strengths are relatively large and close. According to the simulation results in Section \ref{Sec.Simu_thirdstage}, it does indeed indicate that in such situations, using $\operatorname{Corre}(\widehat{L})$ yields better results. }

\par In the third stage, we execute $K$-means clustering utilizing the proposed method alongside the LVGGM method. Owing to the specific characteristics of the dataset in question, prior knowledge regarding the sectors each stock pertains to is at our disposal. As such, we operate under the assumption that the true labels and the number of clusters are known a priori, with $m$ equating to 3. Remarkably, due to the compression effect instigated by the proposed method, 7 rows in $\widehat{L}$ are entirely reduced to zero, thus clustering is conducted on the remnant 38 stocks via both the proposed and LVGGM methods. Direct clustering on $\widehat{L}$ yields a Hamming error rate of 0.079 for LVGGM, whereas the proposed method incurs a higher rate of 0.132. Conversely, when the clustering is predicated upon $\operatorname{Cor}(\operatorname{abs}(\widehat{L}))$, the Hamming error rate achieved by both methods is equitably 0. Upon close comparison, it is apparent that $K$-means clustering predicated on $\operatorname{Cor}(\operatorname{abs}(\widehat{L}))$ offers more promising results.

\section{Conclusion}\label{Sec.Conclusion}
In this paper, we introduce a novel graph decomposition approach that combines a sparse component and low-rank diagonal blocks. We then present a three-stage estimation procedure to identify these two structures. Specifically, we posit that, upon mitigating the impact of global factors, the residual factors manifest in distinct, non-overlapped groups (communities). We discuss the importance of this decomposition from two perspectives. Subsequently, we develop an adaptive $\ell_1$ penalized estimation procedure, integrated with a $K$-means clustering algorithm in the second and third stages respectively. Theoretically, we expand upon the concepts of manifolds and tangent spaces about low-rank diagonal blocks. This expansion establishes the conditions necessary for the identifiability of our proposed decomposition.
To ensure model selection consistency in the second stage, we refine the classical irrepresentable condition. In the final stage, we provide an error bound for the clustering accuracy of the $K$-means algorithm. Through simulations and empirical data analysis, we demonstrate the superiority of our method over existing techniques.
Future research could intriguingly explore the extension of our method to scenarios with overlapped community structures, unobserved global factors and discrete data.

\newpage
\bibliographystyle{apalike}
\bibliography{ref}

\newpage
\appendix
\section{Algorithm details}\label{Appen.Algorithm}
\numberwithin{equation}{section}
\setcounter{equation}{0}
\setcounter{theorem}{0}
\setcounter{lemma}{0}
\setcounter{remark}{0}
To ensure convergence of the ADMM algorithm, an equivalent convex minimization problem with two blocks of variables and two separable functions has been proposed in the problem \eqref{blockobj}. Denote $Y_1=(\Theta, S,  L_1, L_2)$, $Y_2=(\widetilde{\Theta}, \widetilde{S},  \widetilde{L}_1,\widetilde{L}_2), $ problem \eqref{blockobj} is 
\begin{equation}\label{Appen.Blockobj}
	\begin{aligned}
		\min\quad &f(Y_1)+\psi(Y_2)\\
		s.t. \quad& Y_1-Y_2 = 0
	\end{aligned}
\end{equation}
where $f(Y_1) \triangleq f_1(\Theta)+f_2(S)+f_3(L_1)+f_4(L_2)$ with
$$
\begin{aligned}
	f_1(\Theta)&=-\log \operatorname{det}(\Theta)+\text{tr}\left(\widehat{\Sigma}\Theta  \right),\\
	f_2(S) &=\gamma_n \cdot\|S\|_1,\\ 
	f_3(L_1)& =\tau_n\sum_{i,j = 1}^pw_{ij}| L_{1,ij}|,\quad f_4(L_2) = \delta_n\Vert L_2\Vert_{*},\\
\end{aligned}
$$ 
and 
$\psi(Y_2) \triangleq \psi_1(\widetilde{\Theta}, \widetilde{S},  \widetilde{L}_1) + \psi_2(\widetilde{L}_1,\widetilde{L}_2)$ with
$$
\begin{aligned}
	\psi_1(\widetilde{\Theta},  \widetilde{L}_1, \widetilde{S}) &= \mathcal{I}\{\widetilde{\Theta}-\widetilde{L}_1-\widetilde{S}=0\},\\
	\psi_2(\widetilde{L}_2) &= \mathcal{I}\{\widetilde{L}_1-\widetilde{L}_2=0\},\\
\end{aligned}
$$ 
The indicator function is defined as 
$$ \mathcal{I}(X\in \mathcal{X})\triangleq \begin{cases}0, & \text { if } X\in \mathcal{X} \\ +\infty, & \text { otherwise }\end{cases}.
$$
Define the augmented Lagrangian function as 
$$
\mathcal{L}_{\mu}(Y_1, Y_2,\Gamma)=f(Y_1)+\psi(Y_2)+\Gamma\cdot (Y_1-Y_2)+\frac{1}{2\mu}\|Y_1-Y_2\|_F^2,
$$
where $\Gamma$ is the Lagrange multiplier and $\mu>0$ is the penalty parameter. Then the problem \eqref{Appen.Blockobj} can be separated as some solvable subproblems through the scaled ADMM algorithm:
$$
\left\{\begin{aligned}
	Y_1^{k+1} & \triangleq \argmin_{Y_1}\mathcal{L}_{\mu}(Y_1, Y_2^k,\Gamma^k) =\argmin_{Y_1} f(Y_1) + \frac{1}{2 \mu}\left\|Y_1-Y_2^k + \Gamma^k\right\|_F^2 \\
	Y_2^{k+1} & \triangleq\argmin_{Y_2}\mathcal{L}_{\mu}(Y_1^{k+1}, Y_2,\Gamma^k) =\argmin_{Y_2} \psi(Y_2) + \frac{1}{2 \mu}\left\|Y_1^{k+1}-Y_2 + \Gamma^k\right\|_F^2, \\
	\Gamma^{k+1} & \triangleq\Gamma^k+\left(Y_1^{k+1}-Y_2^{k+1}\right).
\end{aligned}\right.
$$
We first define the proximal mapping for function $p(X): \mathbb{R}^{p\times q} \to \mathbb{R}$ as
$$
\operatorname{prox}_{\lambda, p(\cdot)}(Z) = \argmin_X p(X) + \frac{1}{2\lambda}\|X-Z\|_F^2.
$$
The proximal mapping for $p(X) = \|X\|_1$ is 
$$
\begin{aligned}
	\operatorname{prox}_{\lambda, \|\cdot\|_1}(Z) &= \argmin_X \|X\|_1 + \frac{1}{2\lambda}\|X-Z\|_F^2\\
	&=\operatorname{sign}(Z)(Z-\lambda\textbf{1}\textbf{1}^T)_+,
\end{aligned}
$$
where $(\cdot)_+ = \max(\cdot, 0)$. 
The proximal mapping for $p(X) = \|X\|_*, X\succeq 0$ is 
$$
\begin{aligned}
	\operatorname{prox}_{\lambda, \|\cdot\|_*}(Z) &= \argmin_{X\succeq 0} \|X\|_* + \frac{1}{2\lambda}\|X-Z\|_F^2\\
	&=U\widetilde{\Sigma} U^T,
\end{aligned}
$$
where $Z = U\Sigma U^T, \Sigma = \operatorname{diag}(\sigma_1, \dots, \sigma_p)$ and $\widetilde{\Sigma} = \operatorname{diag}\left((\sigma_1 -\lambda)_+,\dots, (\sigma_p -\lambda)_+ \right)$. 
The proximal mapping for $p(X) = I(X\in \mathcal{X})$ is the projection mapping
\[
\begin{aligned}
	\operatorname{prox}_{I}(Z) &= \argmin_{X\in \mathcal{X}}  \frac{1}{2}\|X-Z\|_F^2\\
	&= \mathcal{P}_{\mathcal{X}}(Z),
\end{aligned}
\]
where $\mathcal{P}_{ \mathcal{X}}(Z)$ means the projection of $Z$ on set $\mathcal{X}$ with respect to the inner product $"\cdot"$.
\subsection{First subproblem}
We consider the solution for the first subproblem:
$$
Y_1^{k+1} =\argmin_{Y_1} f(Y_1) + \frac{1}{2 \mu}\left\|Y_1-Y_2^k + \Gamma^k\right\|_F^2.
$$
Denote $Y_1^{k+1} = (\Theta^{k+1}, S^{k+1},L_1^{k+1}, L_2^{k+1})$, and partition the matrix $\Gamma^{k} = (\Gamma_{\Theta}^k, \Gamma_{S}^k, \Gamma_{L_1}^k, \Gamma_{L_2}^k)$ into four blocks associated to $Y_2^k$.  This problem can also be reduced to the following four subproblems:

\subsubsection{Update step for $\Theta^{k+1}$}
\begin{equation}\label{Theta}
	\begin{aligned}
		\Theta^{k+1} &=\argmin_{\Theta\succeq 0} f_1(\Theta)+\frac{1}{2 \mu}\left\|\Theta-\widetilde{\Theta}^k + \Gamma_{\Theta}^k\right\|_F^2 \\
		& =\argmin_{\Theta\succeq 0} -\log \operatorname{det}(\Theta)+\text{tr}\left(\widehat{\Sigma}\Theta  \right)  + \frac{1}{2\mu}\left\|\Theta-\widetilde{\Theta}^k + \Gamma_{\Theta}^k\right\|_F^2 ,
	\end{aligned}
\end{equation}
The first-order optimal condition is given by 
$$
-\Theta^{-1} + \widehat{\Sigma} + \frac{1}{\mu}(\Theta - \widetilde{\Theta}^k + \Gamma_{\Theta}^k) = 0.
$$
Denote the eigenvalue decomposition for $\widetilde{\Theta}^k - \mu\widehat{\Sigma} - \Gamma_{\Theta}^k$ by $U\operatorname{diag}(\sigma)U^T$, then it is easy to verify that
$$
\Theta^{k+1} = U\operatorname{diag}(\xi)U^T,\quad \xi_i = \frac{\sigma_i + \sqrt{\sigma_i^2 + 4\mu}}{2}
$$
satisfies the optimal condition and thus is the solution to the problem \eqref{Theta}.  (\cite{ma2013alternating})

\subsubsection{Update step for $S^{k+1}$}
$$
\begin{aligned}
	\left(S^{k+1}\right)_{i j}&=\argmin_{S} f_2(S)+\frac{1}{2 \mu}\left\|S-\widetilde{S}^{k}+ \Gamma_{S}^{k}\right\|_F^2 \\
	& =\argmin_{S} \|S\|_{\text{off},1}+\frac{1}{2 \mu \gamma_n}\left\|S-\widetilde{S}^{k}+ \Gamma_{S}^{k}\right\|_F^2\\
	&= \left\{\begin{array}{l}
		\operatorname{prox}_{\mu\gamma_n ,\|\cdot\|_1}\left(\widetilde{S}^{k}_{ij}- (\Gamma_{S}^{k})_{ij}\right), i \neq j \\
		\widetilde{S}_{i j}^{k}-\left(\Gamma_{S}^{k}\right)_{i j}, \quad i = j .
	\end{array}\right. 
\end{aligned}
$$

\subsubsection{Update step for $L_1^{k+1}$}
Let $\Se =  \{L: L\succeq0, L = L^T \}$ be the semidefinite positive cone,
$$
\begin{aligned}
	L_1^{k+1} &= \argmin_{L_1\succeq 0}f_3(L_1) + +\frac{1}{2\mu}\left(\left\|L_1-\widetilde{L}_1^k + \Gamma_{L_1}^k\right\|_F^2\right)\\
	&=\argmin_{L_1\succeq 0}\sum_{i,j = 1}^pw_{ij}| L_{1,ij}|+\frac{1}{2\mu \tau_n}\left(\left\|L_1-\widetilde{L}_1^k + \Gamma_{L_1}^k\right\|_F^2\right)\\
	& = \mathcal{P}_{\Se}\left( \left(\prox_{\mu \tau_n w_{ij}, \|\cdot\|_1}(\widetilde{L}_1^k - \Gamma_{L_1}^k) \right)_{ij}  \right)
\end{aligned}
$$

\subsubsection{Update step for $L_2^{k+1}$}
$$
\begin{aligned}
	L_2^{k+1} &= \argmin_{L_2\succeq 0}f_4(L_2) + +\frac{1}{2\mu}\left(\left\|L_2-\widetilde{L}_2^k + \Gamma_{L_2}^k\right\|_F^2\right)\\
	&=\argmin_{L_2\succeq 0}\|L_2\|_*+\frac{1}{2\mu \delta_n}\left(\left\|L_2-\widetilde{L}_2^k + \Gamma_{L_2}^k\right\|_F^2\right)\\
	& = \prox_{\mu \delta_n, \|\cdot\|_*}(\widetilde{L}_2^k - \Gamma_{L_2}^k)
\end{aligned}
$$

\subsection{Second subproblem}
Partition the matrix $T^k := Y_1^{k+1} + \Gamma^k = (T^k_{\Theta}, T^k_{S}, T^k_{L_1}, T^k_{L_2})$
	into four blocks in the same form as $Y_2 = (\widetilde{\Theta}, \widetilde{L}_1,\widetilde{L}_2, \widetilde{S})$. The second subproblem can be reduced to 
	\begin{equation}\label{secondsubproblem}
		\begin{aligned}
			\min &\quad \frac{1}{2}\left\|(\widetilde{\Theta}, \widetilde{S}, \widetilde{L}_1, \widetilde{L}_2)-(T^k_{\Theta}, T^k_{S}, T^k_{L_1}, T^k_{L_2})\right\|_F^2 \\
			s.t. &\quad \widetilde{\Theta}-\widetilde{L}_1-\widetilde{S}=0, \quad \widetilde{L}_1 - \widetilde{L}_2 =0.
		\end{aligned}
	\end{equation}
	The Lagrangian function associated with this problem is 
	\[
	\mathcal{L}(Y_2) = \frac{1}{2}\left\|(\widetilde{\Theta}, \widetilde{S}, \widetilde{L}_1, \widetilde{L}_2)-(T^k_{\Theta}, T^k_{S}, T^k_{L_1}, T^k_{L_2})\right\|_F^2 + \Psi_1\cdot (\widetilde{\Theta}-\widetilde{L}_1-\widetilde{S}) + \Psi_2\cdot(\widetilde{L}_1 - \widetilde{L}_2),
	\]
	where $\Psi_1, \Psi_2$ are the corresponding Lagrangian multipliers. 
	The first-order optimal conditions  are given by
	$$
	(\widetilde{\Theta},\widetilde{S},\widetilde{L}_1,\widetilde{L}_2)-(T^k_{\Theta}, T^k_{S}, T^k_{L_1}, T^k_{L_2})+(\Psi_1,-\Psi_1,-\Psi_1 + \Psi_2, -\Psi_2)=0.
	$$
	Then
	$$
	\widetilde{\Theta}=T_{\Theta}^k - \Psi_1, \quad \widetilde{S}=T_{S}^k + \Psi_1, \quad \widetilde{L}_1 =T_{L_1}^k + \Psi_1 - \Psi_2, \quad \widetilde{L}_2 =T_{L_2}^k + \Psi_2 .
	$$
	Substituting them into the equality constraint of the problem \eqref{secondsubproblem}, we get
	$$
	\begin{aligned}
		\Psi_1 &= \frac{2}{5}T^k_{\Theta}-\frac{2}{5}T^k_{S} -\frac{1}{5}T^k_{L_1}-\frac{1}{5}T^k_{L_2}\\
		\Psi_2 &= \frac{1}{5}T^k_{\Theta}-\frac{1}{5}T^k_{S} +\frac{2}{5}T^k_{L_1}-\frac{3}{5}T^k_{L_2}
	\end{aligned}.
	$$
	Finally we get the solution $Y_2 = (\widetilde{\Theta}^{k+1}, \widetilde{S}^{k+1}, \widetilde{L}_1^{k+1}, \widetilde{L}_2^{k+1})$ for the problem \eqref{secondsubproblem},
	$$
	\left\{\begin{array}{l}
		\widetilde{\Theta}^{k+1}=  \frac{3}{5}T^k_{\Theta} + \frac{2}{5}T^k_{S}+ \frac{1}{5}T^k_{L_1} + \frac{1}{5}T^k_{L_2} \\		
		\widetilde{S}^{k+1}= \frac{2}{5}T^k_{\Theta}+\frac{3}{5}T^k_{S}- \frac{1}{5}T^k_{L_1}-\frac{1}{5}T^k_{L_2}\\
		\widetilde{L}_1^{k+1} = \frac{1}{5}T^k_{\Theta}-\frac{1}{5}T^k_{S} + \frac{2}{5}T^k_{L_1}+\frac{2}{5}T^k_{L_2}\\
		\widetilde{L}_2^{k+1}= \frac{1}{5}T^k_{\Theta}-\frac{1}{5}T^k_{S} + \frac{2}{5}T^k_{L_1}+\frac{2}{5}T^k_{L_2}
	\end{array}\right..
	$$
\subsection{Algorithm for graphical models with non-overlapped grouped latent variables}
For the graphical models with non-overlapped grouped latent variables mentioned in Section \ref{Sec.GraphModel}, the algorithm is a bit different from above due to $-1$ in the decomposition. The main difference is reflected in the solution of the second subproblem,
$$
       \left\{\begin{array}{l}
		\widetilde{\Theta}^{k+1}=  \frac{3}{5}T^k_{\Theta} + \frac{2}{5}T^k_{S}- \frac{1}{5}T^k_{L_1} - \frac{1}{5}T^k_{L_2} \\		
		\widetilde{S}^{k+1}= \frac{2}{5}T^k_{\Theta}+\frac{3}{5}T^k_{S}+ \frac{1}{5}T^k_{L_1}+\frac{1}{5}T^k_{L_2}\\
		\widetilde{L}_1^{k+1} = -\frac{1}{5}T^k_{\Theta}+\frac{1}{5}T^k_{S} + \frac{2}{5}T^k_{L_1}+\frac{2}{5}T^k_{L_2}\\
		\widetilde{L}_2^{k+1}= -\frac{1}{5}T^k_{\Theta}+\frac{1}{5}T^k_{S} + \frac{2}{5}T^k_{L_1}+\frac{2}{5}T^k_{L_2}
	\end{array}\right..
$$

\newpage
\section{Technical Details}
\setcounter{equation}{0}
\setcounter{theorem}{0}
\setcounter{lemma}{0}
\setcounter{remark}{0}
\setcounter{proposition}{0}
%\renewcommand{\theequation}{A.\arabic{equation}}
%\subsection{Proof of Theorem \ref{Thm.IdentiforL}}

\subsection{Proof of Proposition}\label{Appen.Propositionproof}
\subsubsection{Proof of Proposition \ref{P.TangentSpace}}\label{Appen.ProProof.Tangent}
(1). Denote $\mathcal{L}(L_i^*) = \{L_i\in\mathbb{S}^{d_i\times d_i}:rank(L_i)= rank(L_i^*)\}$ (they are manifolds, also submanifolds of $ \mathbb{S}^{p\times p}$) and $\mathcal{M}(L^*) = \{L=\operatorname{diag}(L_1,\cdots, L_m):rank(L_i) = rank(L_i^*), L_i\in\mathbb{S}^{d_i\times d_i}\}$. Since $ \mathcal{M}(L^*) \cong \mathcal{L}(L_1^*)\times \cdots\times \mathcal{L}(L_m^*) $ ("$\cong$" means diffeomorphism), according to Example 1.13 in \cite{Lee00}, $\mathcal{M}(L^*)$ is a product manifold. In addition, since $\mathcal{M}'(L^*) = \{L:rank(L) = rank(L^*), \operatorname{Supp}(L) = \operatorname{Supp}(L^*)\}$ is an open subset in $\mathcal{M}(L^*)$, according to Example 1.8 in \cite{Lee00}, $\mathcal{M}'(L^*)$ is also an open submanifold of $\mathcal{M}(L^*)$. All $L\in\mathcal{M}'(L^*)$ form the smooth points in $\mathcal{LS}(L^*)$.\\
(2). Since $\maT_1(L) = \{U_1Y^T + YU_1^T:Y\in\mathbb{R}^{p\times r}\}$ is the tangent space of $L\in\mathcal{M}(L^*)$ \citep{chandrasekaran2010SIAM} and  $\mathcal{M}'(L^*)$ is also an open submanifold of $\mathcal{M}(L^*)$, then $\maT(L) = \maT_1(L)\cap \maT_2^*$. \\
(3). Moreover, note that $\maV(L)$ is orthogonal to $\maT_2^{*\perp}$ and they are both the subspace of $\maT^\perp(L)$. Hence, we only need to prove that $\operatorname{dim}(\maT^\perp(L)) = \operatorname{dim}(\maV(L)) + \operatorname{dim}(\maT_2^{*\perp})$. In fact, from definition, we know
\[
\operatorname{dim}(\maT^\perp(L)) = \operatorname{dim}(\mathbb{S}^{p\times p}) - \operatorname{dim}(\maT(L)) = \operatorname{dim}(\mathbb{S}^{p\times p}) - \sum_{i=1}^m\operatorname{dim}(\maT_1(L_i)) 
\]
\[
\operatorname{dim}(\maV(L)) = \sum_{i = 1}^m[ \operatorname{dim}(\mathbb{S}^{d_i\times d_i}) - \operatorname{dim}(\maT_1(L_i))] 
\]
\[
\operatorname{dim}(\maT_2^{*\perp}) = \sum_{i<j} \operatorname{dim}(\mathbb{R}^{d_i\times d_j}) 
\]
and $ \operatorname{dim}(\mathbb{S}^{p\times p}) = \operatorname{dim}(\mathbb{S}^{d_i\times d_i}) + \sum_{i<j} \operatorname{dim}(\mathbb{R}^{d_i\times d_j}) $, which leads to the conclusion.

\subsection{Proof of Theorem \ref{Thm.IdentiforL}}
According to \cite{chandrasekaran2010SIAM}, for every $i \in [m]$, $(S_i^*, L_i^*)$ is locally identifiable with respect to $\mathcal{S}(s_i)\times\mathcal{L}(r_i)$ under Assumption \ref{A.transver}, where $s_i\triangleq\|\mathbf{O}(S_i^*)\|_0$, i.e., there exists $\varepsilon>0$ such that if it holds $S_i+L_i = S_i^*+L_i^*$ and $\|S_i- S_i^*\|_\infty\leq \varepsilon, \|L_i - L_i^*\|_2\leq \varepsilon $ for $(S_i, L_i)\in\mathcal{S}(s_i)\times\mathcal{L}(r_i) $, then $S_i = S_i^*, L_i = L_i^*$. Hence for any $(S, L)\in \mathcal{S}(s_0)\times\mathcal{LS}(m,r)$ satisfying $\|S- S^*\|_\infty\leq \varepsilon, \|L - L^*\|_2\leq \varepsilon $, if $S+L = S^* + L^*$, then $L = L^*$ (since $L = \operatorname{diag}(L_1,\cdots, L_m)$), which also implies $S = S^*$. This completes the proof.

\subsection{Proof of Theorem \ref{Thm.Main}}
\textbf{Proof Strategy}.  The proof line is similar to \cite{chen2016fused}. We first provide a sketch of the proof for Theorem \ref{Thm.Main}. We introduce several notations and definitions. Let the eigendecomposition of $L^*$ be $L^*=$ $U^* D^* U^{* \top}$, such that $U^*$ is a $p \times p$ orthogonal matrix and $D^*$ is a $p \times p$ diagonal matrix whose first $r$ diagonal elements are strictly positive. We write $U^*=\left[U_1^*, U_2^*\right]$ where $U_1^*$ is the first $r$ columns of $U^*$. Let $D_1^*$ be the $r \times r$ diagonal matrix containing the nonzero diagonal elements of $D^*$. Define the localization set
$$
\begin{aligned}
	\mathcal{M}_1=\left\{(S, L): S\right. &=S^*+\Delta_S, \left\|\Delta_S\right\|_{\infty} \leq \gamma_n^{1-2 \eta}, \Delta_S \text { is symmetric, } \\
	L&=U \Sigma U^{\top},  \left\|U-U^*\right\|_{\infty} \leq \gamma_n^{1-\eta}, U \text { is a } p \times p \text { orthogonal matrix, } \\
     &\left.\left\|\Sigma-\Sigma^*\right\|_{\infty} \leq 	 \gamma_n^{1-2 \eta}, \text { and } \Sigma \text { is a } p\times p\text { diagonal matrix }\right\},
\end{aligned}
$$
and a subset
$$
\begin{aligned}
	\mathcal{M}_2=\left\{(S, L): S=S^*+\Delta_S,\right. & \left\|\Delta_S\right\|_{\infty} \leq \gamma_n^{1-2 \eta}, \Delta_S \in \Omega^*, \\
	L=U_1 \Sigma_1U_1^{\top}, & \left\|U_1 - U_1^*\right\|_{\infty} \leq \gamma_n^{1-\eta}, U_1\in\mathbb{R}^{p \times r}, U_1^{\top} U_1 =I_r, 
(U_1)_{G^*_{ij}} = 0,i\not=j,\\
	& \left.\left\|\Sigma_1-\Sigma_1^*\right\|_{\infty} \leq \gamma_n^{1-2 \eta}, \text { and } \Sigma_1 \text { is a } r \times r \text { diagonal matrix }\right\},
\end{aligned}
$$
%	&U_1= \left[(\widetilde{U}_1^{T},\mathbf{ 0}_{ r_1\times(p-d_1)})^T,\cdots, (\mathbf{0}_{ r_m\times(p-d_m)},\widetilde{U}_m^{T} )^T\right]
where as discussed under Assumption \ref{A.non-overlapped}, we choose $U_1$ owning the same zero-block pattern as  $(U_1)_{G^*_{ij}} = 0,i\not=j$, and $\eta$ being a positive constant that is sufficiently small. For each pair of $(S, L) \in \mathcal{M}_1, S$ is close to $S^*$. Moreover, the eigendecomposition of $L$ and $L^*$ are close to each other. As the sample size $n$ grows large, the set $\mathcal{M}_1$ will tend to $\left\{\left(S^*, L^*\right)\right\}$. The set $\mathcal{M}_2$ is a subset of $\mathcal{M}_1$. For each pair of $(S, L) \in \mathcal{M}_2, S$ has the same sparsity pattern as $S^*$, and $L$ has the same rank and block-sparsity pattern as $L^*$ for sufficiently large $n$.\\
The proof consists of three steps.
\begin{enumerate}
	\item To be ready, we present some technical lemmas in Section \ref{Appen.Lemmas}. In addition, we also provide some other useful lemmas interspersed with the proof of the main theorem. We later give the proof of all the lemmas in Section \ref{Appen.Lemmaproof}.
	\item We prove that with a probability converging to 1 the optimization problem (\ref{penobj}) restricted to the subset $\mathcal{M}_2$ has a unique solution, which does not lie on the manifold boundary of $\mathcal{M}_2$. This part of the proof is presented in Step 1 of Section \ref{Appen.TheoremMainproof}.
	\item We then show the unique solution restricted to $\mathcal{M}_2$ is also a solution to (\ref{penobj}) on $\mathcal{M}_1$. It is further shown that with probability converging to 1, this solution is the unique solution to (\ref{penobj}) restricted to $\mathcal{M}_1$. This part of the proof is presented in  Step 2 of Section \ref{Appen.TheoremMainproof}.
\end{enumerate}

\par The previous three steps together imply that the convex optimization problem (\ref{penobj}) with the constraint $(S, L) \in \mathcal{M}_1$ has a unique solution that belongs to $\mathcal{M}_2$. Furthermore, this solution $(\widehat{S}, \widehat{L})$ is an interior point of $\mathcal{M}_1$ and thus is also the unique solution to the optimization problem (\ref{penobj}) thanks to the convexity of the objective function. We conclude the proof for Theorem \ref{Thm.Main} by noticing that all $(S, L) \in \mathcal{M}_2$ converge to the true parameter $\left(S^*, L^*\right)$ as $ n \rightarrow \infty$.

\subsubsection{Lemmas}\label{Appen.Lemmas}
\numberwithin{equation}{subsection}
%Consider the following the problem,
%\begin{equation}\label{P1}
%	\begin{aligned}
%		( \widehat{S}, \widehat{L}) &= \argmin_{(S, L)} \ell(S,L) + \gamma \|S\|_{\text{off},1}  + \delta\|L\|_*+ \tau\|W\odot L\|_1\\
%		s.t.\quad & L + S \succ 0.
%	\end{aligned} \tag{P1}
%\end{equation} 
\par We first establish some useful lemmas.
\begin{lemma}\label{L.WeightOrder}
	Let $\mathbf{s}_{n 1}=W_n\odot \operatorname{sign}\left(L^*\right)$. Suppose Assumption \ref{A.Wn} holds. Then,
	$$
	\left\|\mathbf{s}_{n 1}\right\|=\left(1+o_P(1)\right) M_{n 1} = o_P(r_n^{1-\alpha}).
	$$
\end{lemma}

\begin{lemma}\label{L.Lipschitz}
	Denote the eigendecomposition $L=U_1 \Sigma_1 U_1^{\top}$, and define the corresponding linear spaces $\maT(L)$ and $\mathcal{D}$ as
	$$
	\mathcal{D}=\left\{U_1 \Sigma_1^{\prime} U_1^{\top}: \Sigma_1^{\prime} \text { is a } r \times r\text { diagonal matrix }\right\},
	$$
	and
	$$
	\maT(L)=\left\{U_1 Y+Y^{\top} U_1^{\top}: Y \text { is a } r \times p \text { matrix }\right\} .
	$$
	Then, the mappings $U_1 \rightarrow \mathcal{P}_{\maT(L)}, U_1 \rightarrow \mathcal{P}_{\maT^\perp(L)}$ and $U_1 \rightarrow \mathcal{P}_{\mathcal{D}}$ are Lipschitz in $U_1$. That is, for all $r \times r$ symmetric matrix $M$, there exists a constant $\kappa$ such that
	$$
	\begin{aligned}
		& \quad \max \left\{\left\|\mathcal{P}_{\maT(L)} M-\mathcal{P}_{\maT^*} M\right\|_{\infty},\left\|\mathcal{P}_{\maT^\perp(L)} M-\mathcal{P}_{T^{*\perp}} M\right\|_{\infty},\left\|\mathcal{P}_{\mathcal{D}} M-\mathcal{P}_{\mathcal{D}^*} M\right\|_{\infty}\right\} \\
		& \leq \kappa\left\|U_1-U_1^*\right\|_{\infty}\|M\|_{\infty} .
	\end{aligned}
	$$
\end{lemma}

Denote
\[ g_{\rho_1,\rho_2,L,\widetilde{W}_n}(L') = \max\left\{ \frac{\|L_1\|_2}{\rho_1}, \frac{r_n^{1-\alpha}\|\widetilde{W}_n\odot L_2\|_{\infty} }{\rho_2}  \right\}  ,\]
where $L' = L_1 \oplus L_2$ is the unique orthogonal direct sum decomposition of $L$ restricted to $  \maV(L)\oplus \maT_2^{*\perp}= \maT^\perp(L) $, i.e., $L_1 = \mathcal{P}_{\maV(L)}(L'), L_2 = \mathcal{P}_{\maT_2^{*\perp}}(L')$. For convenience, without confusion sometimes we will abbreviate $\maV(L)$ as $\maV$. The following lemma states that this norm is well-defined. 

\begin{lemma}\label{L.DirectDecomp}
	Given any smooth point  $L$ of $\mathcal{LS}(L^*)$ with  $\maT^\perp(L)= \maV(L) \oplus \maT_2^{*\perp} $, 
	\begin{enumerate}
		\item $ g_{\rho_1,\rho_2, L, \widetilde{W}_n}(L) $ is a norm for fixed $\widetilde{W}_n, L$.
		\item $\|(\mathcal{P}_{\maV(L)} - \mathcal{P}_{\maV^*})(M)\|_2 \lesssim\|L-L^*\|_2\|M\|$.
	\end{enumerate}
\end{lemma}

\begin{lemma}\label{L.LSubgradient}
The subgradient for $\partial(\delta_n\|L\|_* +\tau_n\|W_n\odot L\|_1)$ has the following form 
$$
\begin{aligned}
    \partial(\delta_n\|L\|_* +\tau_n\|W_n\odot L\|_1) = &\{ \delta_n UU^T + \tau_nW_n\odot\operatorname{sign}(L^*) + \delta_n F_1 + \tau_n W_n\odot F_2:\\
    & F_1\in \maT^\perp_1(L), F_2\in \maT_2^{*\perp}, \|F_1\|_\infty\leq 1, \|F_2\|_2\leq 1  \} . 
\end{aligned}
$$
For any $Z\in \mathbb{R}^{p\times p}$, let $L = U\Sigma U^T$ be the eigendecomposition  of $L$. If 
\[ \mathcal{P}_{\maT(L)}(Z) =  \delta_nUU^T+   \tau_n\mathcal{P}_{\maT(L)}\left(W_n\odot\operatorname{sign}(L) \right) , \]
and
\[ \quad g_{\rho_1,\rho_2,L,\widetilde{W}_n}\left(\mathcal{P}_{\maT^\perp(L)}(Z) -\tau_n\mathcal{P}_{\maT^\perp(L)}\left(W_n\odot\operatorname{sign}(L) \right)\right) \leq \gamma_n
,\]
then $Z\in \partial\left(  \delta_n\|L\|_*+\tau_n\|W_n\odot L\|_1 \right)$.
\end{lemma}
\begin{lemma}\label{L.FInvertible}
	Under Assumption \ref{A.transver}, the linear operator $\mathbf{G}$ is invertible over $\Omega^* \times \maT^*$.
\end{lemma}

\begin{lemma}\label{L.AngleL}
Given any smooth point  $L$ of $\mathcal{LS}(L^*)$ with $\|L-L^*\|_2\leq d_0$ for sufficiently small $d_0>0$. For any   $L'\in \maT^\perp(L) = \maV(L)\oplus \maT_2^{*\perp}$, it can be uniquely decomposed as $L' = L_1 \oplus L_2$.  Then there exists some constants $\varepsilon_0>0, \varepsilon_1>0$  such that
	\[  \|L'\|_F^2  = \|L_1\|_F^2 + \|L_2\|_F^2. \]
	Moreover, 
	\[
   \max\big\{\|S\|_{\infty},l_n\|L'\|_2 \big\}	\leq h_{\rho_1, \rho_2,L, \Lambda_n}(S,L') \leq \max\big\{\|S\|_{\infty},u_n\|L'\|_2 \big\},
	\]
	where $l_n, u_n$  are bounded as $n$ tending to infinite under Assumption \ref{A.Wn}. (All the constants are specified in the proof.)
\end{lemma}
\begin{remark}
	 This lemma states that the norm depending on $L$ can be dominated by the norm $\max\{\|\cdot\|_{\infty}, \|\cdot\|_2\}$ uniformly.
\end{remark}

\subsubsection{Main Steps}\label{Appen.TheoremMainproof}
\subsubsection{Step 1}
Denote by $\left(\widehat{S}_{\mathcal{M}_2}, \widehat{L}_{\mathcal{M}_2}\right)$ a solution to the optimization problem
\begin{equation}\label{eq.objconstriantM2}
\begin{aligned}
	& \min \left\{\ell_n(S+L)+\gamma_n \|S\|_{\text{off},1}  + \delta_n\|L\|_*+ \tau_n\|W_n\odot L\|_1\right\} \\
	& \text { subject to }\quad L\succeq 0, \quad S = S^T,\quad ( S,L) \in \mathcal{M}_2 .
\end{aligned}	
\end{equation}
 We write the eigendecomposition $\widehat{L}_{\mathcal{M}_2}=\widehat{U}_{1, \mathcal{M}_2} \widehat{\Sigma}_{1, \mathcal{M}_2} \widehat{U}_{1, \mathcal{M}_2}^{\top}$, where $\widehat{U}_{1, \mathcal{M}_2}^{\top} \widehat{U}_{1, \mathcal{M}_2}=I_r$, $ \widehat{U}_{1, \mathcal{M}_2}$ has the same zero-block pattern as $U_1^*$ and $\widehat{\Sigma}_{1, \mathcal{M}_2}$ is a $r \times r$ diagonal matrix. To establish that $\left(\widehat{S}_{\mathcal{M}_2}, \widehat{L}_{\mathcal{M}_2}\right)$ does not lie on the manifold boundary of $\mathcal{M}_2$, it is sufficient to show that
\begin{align}
	\label{eq.Dconstraint.S}\left\|\widehat{S}_{\mathcal{M}_2}-S^*\right\|_{\infty} & <_P  \gamma_n^{1-2 \eta}, \\
	\label{eq.Dconstraint.D}\left\|\widehat{\Sigma}_{1, \mathcal{M}_2}-\Sigma_1^*\right\|_{\infty} & <_P  \gamma_n^{1-2 \eta}, \\
    \label{eq.Dconstraint.U}	\left\|\widehat{U}_{1, \mathcal{M}_2}-U_1^*\right\|_{\infty} & <_P  \gamma_n^{1-\eta}.
\end{align}	

\begin{lemma}\label{L.Dconstraint}
	For $ \|\widehat{U}_{1, \mathcal{M}_2} - U_1^*\|_{\infty}\leq \gamma_n^{1-\eta}$,  let
	$$
	\widehat{\mathcal{D}}=\left\{\widehat{U}_{1, \mathcal{M}_2} \Sigma_1^{\prime} \widehat{U}_{1, \mathcal{M}_2}^{\top}: \Sigma_1^{\prime} \text { is a } r \times r \text { diagonal matrix }\right\}
	$$
	Consider the convex optimization problem
	\begin{equation}\label{eq.objconstraintD}
		\min _{S \in \Omega^*, L \in \widehat{\mathcal{D}}}\left\{\ell_n(S+L)+\gamma_n \|S\|_{\text{off},1}  + \delta_n\|L\|_*+ \tau_n\|W_n\odot L\|_1\right\} .
	\end{equation}
	Then (\ref{eq.objconstraintD}) has a unique solution with probability converging to 1. Denote the solution by $\left(\widehat{S}_{\widehat{\mathcal{D}}}, \widehat{L}_{\widehat{\mathcal{D}}}\right)$ and $\widehat{L}_{\widehat{\mathcal{D}}}=\widehat{U}_{1, \mathcal{M}_2} \widehat{\Sigma}_{1, \widehat{\mathcal{D}}} \widehat{U}_{1, \mathcal{M}_2}^{\top}$. In addition, there exists a constant $\kappa>0$ such that $\left\|\widehat{S}_{\widehat{\mathcal{D}}}-S^*\right\|_{\infty} \leq_P \kappa \gamma_n^{1-\eta}$ and $\left\|\widehat{\Sigma}_{1, \widehat{\mathcal{D}}}-\Sigma_1^*\right\|_{\infty} \leq_P \kappa \gamma_n^{1-\eta}$.
\end{lemma}
According to the convexity of the objective function, a direct application of Lemma \ref{L.Dconstraint} is
$$
\left(\widehat{S}_{\mathcal{M}_2}, \widehat{L}_{\mathcal{M}_2}\right)={ }_P\left(\widehat{S}_{\widehat{\mathcal{D}}}, \widehat{L}_{\widehat{\mathcal{D}}}\right) \text { and } \widehat{\Sigma}_{1, \mathcal{M}_2}={ }_P \widehat{\Sigma}_{1, \widehat{\mathcal{D}}}
$$
Thus, (\ref{eq.Dconstraint.S}) and (\ref{eq.Dconstraint.D}) are proved. We show (\ref{eq.Dconstraint.U}) by contradiction. If on the contrary $\| \widehat{U}_{1, \mathcal{M}_2}-$ $U_1^* \|_{\infty}=\gamma_n^{1-\eta}$, then in what follows we will show that
\begin{equation}\label{eq.Contrary}
	\begin{aligned}
	&	\ell_n\left(\widehat{S}_{\widehat{\mathcal{D}}}+\widehat{L}_{\widehat{\mathcal{D}}}\right)+\gamma_n\left\|\widehat{S}_{\widehat{\mathcal{D}}}\right\|_{\text{off},1}+\delta_n\left\|\widehat{L}_{\widehat{\mathcal{D}}}\right\|_* + \tau_n\|W_n\odot \widehat{L}_{\widehat{\mathcal{D}}}\|_1  \\
	>_P& \ell_n\left(S^*+L^*\right)+\gamma_n\left\|S^*\right\|_{\text{off},1}+\delta_n\left\|L^*\right\|_* + \tau_n\|W_n\odot L^*\|_1,
	\end{aligned}
\end{equation}
and thus a contradiction is reached. We start with the Taylor expansion of $\ell_n(S+L)$ around $S^*$ and $L^*$.  Since $\widehat{\Sigma} = \frac{1}{n}\widehat{\boldsymbol{R}}^T\widehat{\boldsymbol{R}}= \frac{1}{n}\boldsymbol{X}^T(\boldsymbol{I}_n - \boldsymbol{\widetilde{C}}(\boldsymbol{\widetilde{C}}^T\boldsymbol{\widetilde{C}})^{-1}\boldsymbol{\widetilde{C}}^T)\boldsymbol{\widetilde{X}}$ satisfying that $\|\widehat{\Sigma} - \Sigma^*\|_{\infty} = O_P(\frac{1}{\sqrt{n}}) $ for fixed $\boldsymbol{C}$ and $\Sigma^* = (S^* + L^*)^{-1}$, we let $\ell(\Theta)=\operatorname{lim}\mathbb{E} \ell_n(\Theta) = -\log\operatorname{det}(\Theta) + \operatorname{tr}(\Sigma^*\Theta)$. Then we have
\begin{equation}\label{eq.taylorexpansion}
	\ell_n\left(\Theta^*+\Delta\right)=\ell\left(\Theta^*\right)+\frac{1}{2} v(\Delta)^{\top} \mathcal{I}^* v(\Delta)+R_n(\Delta),
\end{equation}
where $\Theta^*=S^*+L^*$, and the function $v: \mathbb{R}^p \times \mathbb{R}^p \rightarrow \mathbb{R}^{p^2} \times 1$ is a map that vectorizes a matrix. Moreover, $R_n(\Delta)$ is the remainder term satisfying
\begin{equation}\label{eq.Rn}
	R_n(\Delta)=R_n\left(\mathbf{0}_{p \times p}\right)+O_P\left(\|\Delta\|_{\infty}^3\right)+O_P\left(\frac{\|\Delta\|_{\infty}}{\sqrt{n}}\right) \text { as } \Delta \rightarrow 0,n\rightarrow \infty,
\end{equation}
where $R_n\left(\mathbf{0}_{p\times p}\right)=\ell_n\left(\Theta^*\right)-\ell\left(\Theta^*\right), O_P\left(\|\Delta\|_{\infty} / \sqrt{n}\right)$ term corresponds to $v\left(\nabla \ell_n\left(\Theta^*\right)\right)^{\top} v(\Delta)$, and $O_P\left(\|\Delta\|_{\infty}^3\right)$ characterizes the remainder. Furthermore, from (\ref{eq.taylorexpansion}), as $\Delta \rightarrow 0$,  the first derivative of $R_n(\Delta)$ satisfies
\begin{equation}\label{eq.FirstderivativeRn}
	\begin{aligned}
		\nabla R_n(\Delta)	& = \nabla\ell_n(\Theta^* + \Delta) - \mathcal{I}^*v(\Delta)\\
		& = \nabla\ell_n(\Theta^* + \Delta) - \nabla\ell(\Theta^* + \Delta) +\nabla\ell(\Theta^* + \Delta) -\nabla\ell(\Theta^*) +\nabla\ell(\Theta^*)    -  \mathcal{I}^*v(\Delta)\\
		& = O_P(\frac{1}{\sqrt{n}}) + O_P(\|\Delta\|_\infty^2)           \text { as } \Delta \rightarrow 0,
	\end{aligned}
\end{equation}
where we used the fact $  \nabla\ell(\Theta^*) = 0  $. In addition, the second derivative satisfies
\begin{equation}\label{eq.SeconderivativeRn}
	\begin{aligned}
		\nabla^2 R_n(\Delta) & = \nabla^2 \ell_n\left(\Theta^*+\Delta\right)-\mathcal{I}^*=  \nabla^2 \ell\left(\Theta^*+\Delta\right)-\mathcal{I}^*\\
		&=O_P\left(\|\Delta\|_{\infty}\right).*
	\end{aligned}
\end{equation}
 We plug $\Delta=\widehat{S}_{\widehat{\mathcal{D}}}+\widehat{L}_{\widehat{\mathcal{D}}}-S^*-L^*$ into (\ref{eq.taylorexpansion}), then
\begin{equation}\label{eq.PluginTaylorExpansion}
	\begin{aligned} \ell_n\left(\widehat{S}_{\widehat{\mathcal{D}}}+\widehat{L}_{\widehat{\mathcal{D}}}\right)-\ell_n\left(S^*+L^*\right)  =& \frac{1}{2} v\left(\widehat{S}_{\widehat{\mathcal{D}}}+\widehat{L}_{\widehat{\mathcal{D}}}-\left(S^*+L^*\right)\right)^{\top} \mathcal{I}^* v\left(\widehat{S}_{\widehat{\mathcal{D}}}+\widehat{L}_{\widehat{\mathcal{D}}}-\left(S^*+L^*\right)\right)\\
	&+R_n\left(\widehat{S}_{\widehat{\mathcal{D}}}+\widehat{L}_{\widehat{\mathcal{D}}}-\left(S^*+L^*\right)\right)-R_n\left(\mathbf{0}_{p\times p}\right) .
	\end{aligned}
\end{equation}
We first establish a lower bound for $R_n\left(\widehat{S}_{\widehat{\mathcal{D}}}+\widehat{L}_{\widehat{\mathcal{D}}}-\left(S^*+L^*\right)\right)-R_n\left(\mathbf{0}_{p\times p}\right)$. According to Lemma \ref{L.Dconstraint}, we have
$$
\left\|\widehat{S}_{\widehat{\mathcal{D}}}+\widehat{L}_{\widehat{\mathcal{D}}}-\left(S^*+L^*\right)\right\|_{\infty} \leq_P \kappa \gamma_n^{1-\eta},
$$
with a possibly different $\kappa$. The above display and (\ref{eq.Rn}) yield
\begin{equation}\label{eq.PluginRnError}
	R_n\left(\widehat{S}_{\widehat{\mathcal{D}}}+\widehat{L}_{\widehat{\mathcal{D}}}-\left(S^*+L^*\right)\right)-R_n\left(\mathbf{0}_{p \times p}\right)=O_P\left(\gamma_n^{3(1-\eta)}\right)+O_P\left(\frac{\gamma_n^{1-\eta}}{\sqrt{n}}\right) .
\end{equation}
We proceed to a lower bound for the term $\frac{1}{2} v\left(\widehat{S}_{\widehat{\mathcal{D}}}+\widehat{L}_{\widehat{\mathcal{D}}}-\left(S^*+L^*\right)\right)^{\top} \mathcal{I}^* v\left(\widehat{S}_{\widehat{\mathcal{D}}}+\widehat{L}_{\widehat{\mathcal{D}}}-\left(S^*+L^*\right)\right)$ on the right-hand side of (\ref{eq.PluginTaylorExpansion}) with the aid of Lemma \ref{L.SLseparate} and Lemma \ref{L.DeltaLError}.
\begin{lemma}\label{L.SLseparate} 
	Under Assumption \ref{A.transver}, there exists a positive constant $\varepsilon$ such that
	$$
	\|S+L\|_{\infty} \geq \varepsilon\|L\|_{\infty} \text { for all }(S, L) \in \Omega^* \times \maT^* .
	$$
\end{lemma} 
\begin{lemma}\label{L.DeltaLError}
	Let
	$$
	\Delta_L=U_1^* \Sigma_1^*\left(\widehat{U}_{1, \mathcal{M}_2}-U_1^*\right)^{\top}+\left(\widehat{U}_{1, \mathcal{M}_2}-U_1^*\right) \Sigma_1^* U_1^{* \top}+U_1^*\left(\widehat{\Sigma}_{1, \widehat{\mathcal{D}}}-\Sigma_1^*\right) U_1^{* \top},
	$$
	then under Assumption \ref{A.non-overlapped}  we have
	\begin{enumerate}
		\item $\Delta_L \in \maT^*$.
		\item  There exists positive constant $\varepsilon$ such that $\left\|\Delta_L\right\|_{\infty}>_P \varepsilon \gamma_n^{1-\eta}$, for all $\left\|\widehat{U}_{1, \mathcal{M}_2}-U_1^*\right\|_{\infty}=$ $\gamma_n^{1-\eta}$.
		\item  $\left\|\widehat{L}_{\widehat{\mathcal{D}}}-L^*-\Delta_L\right\|_{\infty} \leq_P \kappa \gamma_n^{2(1-\eta)}$.
	\end{enumerate}
\end{lemma}

According to Lemma  \ref{L.SLseparate}  and Lemma \ref{L.DeltaLError} (i) (iii) and noticing that $\widehat{S}_{\widehat{\mathcal{D}}}-S^* \in \Omega^*$, we have
$$
\left\|\widehat{S}_{\widehat{\mathcal{D}}}+\widehat{L}_{\widehat{\mathcal{D}}}-\left(S^*+L^*\right)\right\|_{\infty} \geq_P\left\|\widehat{S}_{\widehat{\mathcal{D}}}-S^*+\Delta_L\right\|_{\infty}-\kappa \gamma_n^{2(1-\eta)} \geq_P \varepsilon\left\|\Delta_L\right\|_{\infty}-\kappa \gamma_n^{2(1-\eta)} .
$$
According to Lemma \ref{L.DeltaLError} (ii), the above display further implies that
$$
\left\|\widehat{S}_{\widehat{\mathcal{D}}}+\widehat{L}_{\widehat{\mathcal{D}}}-\left(S^*+L^*\right)\right\|_{\infty}>_P \varepsilon \gamma_n^{1-\eta},
$$
with a possibly different $\varepsilon$. Since $\mathcal{I}^*$ is positive definite,
\begin{equation}\label{eq.PluginFisherLowerbound}
	v\left(\widehat{S}_{\widehat{\mathcal{D}}}+\widehat{L}_{\widehat{\mathcal{D}}}-\left(S^*+L^*\right)\right)^{\top} \mathcal{I}^* v\left(\widehat{S}_{\widehat{\mathcal{D}}}+\widehat{L}_{\widehat{\mathcal{D}}}-\left(S^*+L^*\right)\right)>\varepsilon\left\|\widehat{S}_{\widehat{\mathcal{D}}}+\widehat{L}_{\widehat{\mathcal{D}}}-\left(S^*+L^*\right)\right\|_{\infty}^2 \geq_P \varepsilon^2 \gamma_n^{2(1-\eta)}.
\end{equation}
Hence, combing (\ref{eq.PluginTaylorExpansion}), (\ref{eq.PluginRnError}) and (\ref{eq.PluginFisherLowerbound}) give
\begin{equation}\label{eq.PluginLossLowerbound}
     \ell_n\left(\widehat{S}_{\widehat{\mathcal{D}}}+\widehat{L}_{\widehat{\mathcal{D}}}\right)-\ell_n\left(S^*+L^*\right)>_P \frac{\varepsilon^2}{2} \gamma_n^{2(1-\eta)} .
\end{equation}
We proceed to the regularization terms in (\ref{eq.Contrary}). For the $\ell_1$ penalty term for $S$, we have
\begin{equation*}
	\left\|\left(\widehat{S}_{\widehat{\mathcal{D}}}\right)\right\|_{\text{off},1}-\left\|\left(S^*\right)\right\|_{\text{off},1}=\operatorname{sign}\left(\mathbf{O}\left(S^*\right)\right) \cdot\left(\widehat{S}_{\widehat{\mathcal{D}}}-S^*\right)= O_P\left(\gamma_n^{1-\eta}\right) .
\end{equation*}
The second equality in the above display is due to Lemma \ref{L.Dconstraint}. For the nuclear norm term and adaptive $\ell_1$ for $L$, we have also from Lemma \ref{L.Dconstraint},
$$
\left\|\widehat{L}_{\widehat{\mathcal{D}}}\right\|_*-\left\|L^*\right\|_* \geq-\left\|\widehat{L}_{\widehat{\mathcal{D}}}-L^*\right\|_* \geq-\kappa \gamma_n^{1-\eta},
$$
and from Lemma \ref{L.WeightOrder} and Lemma \ref{L.Dconstraint},
$$
\left\|W_n\odot\widehat{L}_{\widehat{\mathcal{D}}}\right\|_1-\left\|W_n\odot L^*\right\|_1 = (W_n\odot\operatorname{sign}(L^*))\cdot\left(\widehat{L}_{\widehat{\mathcal{D}}} - L^*\right)= o_P(r_n^{1-\alpha}\gamma_n^{1-\eta}).
$$
 Notice that $\delta_n=\rho_1 \gamma_n$ and $\tau_n = \rho_2\frac{\gamma_n}{r_n^{1-\alpha}}$. Equations (\ref{eq.PluginFisherLowerbound}), (\ref{eq.PluginLossLowerbound}) and the above three inequalities imply
$$
\begin{aligned}
	& \ell_n\left(\widehat{S}_{\widehat{\mathcal{D}}}+\widehat{L}_{\widehat{\mathcal{D}}}\right)-\ell_n\left(S^*+L^*\right)+\gamma_n\left(\left\|\left(\widehat{S}_{\widehat{\mathcal{D}}}\right)\right\|_{\text{off},1}-\left\|\left(S^*\right)\right\|_{\text{off},1}\right)+\delta_n\left(\left\|\widehat{L}_{\widehat{\mathcal{D}}}\right\|_*-\left\|L^*\right\|_*\right)\\ 
	&  + \tau_n\left(\left\|W_n\odot\widehat{L}_{\widehat{\mathcal{D}}}\right\|_1-\left\|W_n\odot L^*\right\|_1\right)>_P\varepsilon \gamma_n^{2(1-\eta)}>_P 0,
\end{aligned}
$$
with a possibly different $\varepsilon$. Notice that from Lemma \ref{L.Dconstraint}, $\left(\widehat{S}_{\widehat{\mathcal{D}}}, \widehat{L}_{\widehat{\mathcal{D}}}\right)={}_P\left(\widehat{S}_{\mathcal{M}_2}, \widehat{L}_{\mathcal{M}_2}\right)$, so we obtain (\ref{eq.Contrary}) by rearranging terms in the above inequality, and this contradicts the definition of $\widehat{S}_{\mathcal{M}_2}$ and $\widehat{L}_{\mathcal{M}_2}$. This completes the proof for (\ref{eq.Dconstraint.U}). Thus, $\left(\widehat{S}_{\mathcal{M}_2}, \widehat{L}_{\mathcal{M}_2}\right)$ is an interior point of $\mathcal{M}_2$. The uniqueness of the solution is obtained according to the following Lemma \ref{L.SolutionError}.

\begin{lemma}\label{L.SolutionError}
	The solution to the optimization problem (\ref{eq.objconstriantM2}) is unique with a probability converging to 1. In addition,
	$$
	\left(\widehat{S}_{\mathcal{M}_2}, \widehat{L}_{\mathcal{M}_2}\right)=\left(S^*, L^*\right)+\mathbf{G}^{-1}\left(\gamma_n \operatorname{sign}\left(\mathbf{O}\left(S^*\right)\right), \delta_n U_1^* U_1^{* \top} \right)+o_P\left(\gamma_n\right), 
	$$
	as $n \to \infty.$
\end{lemma}

\subsubsection{Step 2}
In this section, we first show that $\left(\widehat{S}_{\mathcal{M}_2}, \widehat{L}_{\mathcal{M}_2}\right)$ is a solution of the optimization problem
\begin{equation}\label{eq.objconstraintM1}
	\begin{aligned}
		& \min \left\{\ell_n(S+L)+\gamma_n \|S\|_{\text{off},1}  + \delta_n\|L\|_*+ \tau_n\|W_n\odot L\|_1\right\} \\
		& \text { subject to }\quad L\succeq 0, \quad S = S^T,\quad ( S,L) \in \mathcal{M}_1 .
	\end{aligned}
\end{equation}
To prove this, we will show that $\left(\widehat{S}_{\mathcal{M}_2}, \widehat{L}_{\mathcal{M}_2}\right)$ satisfies the first order condition
$$
\left.\mathbf{0}_{p \times p} \in \partial_S H\right|_{\left(\widehat{S}_{\mathcal{M}_2}, \widehat{L}_{\mathcal{M}_2}\right)} \text { and }\left.\mathbf{0}_{p \times p} \in \partial_L H\right|_{\left(\widehat{S}_{\mathcal{M}_2}, \widehat{L}_{\mathcal{M}_2}\right)},
$$
where the function $H$ is the objective function
$$
H(S, L)=\ell_n(S+L)+\gamma_n\|S\|_{\text{off},1}+\delta_n\|L\|_* +\tau_n\|W_n\odot L\|_1
$$
and $\partial_S H$ and $\partial_L H$ denotes the sub-differentials of $H$. We first derive an explicit expression of the first-order condition. The sub-differential concerning $S$ is defined as
$$
\left.\partial_S H\right|_{\left(\widehat{S}_{\mathcal{M}_2}, \widehat{L}_{\left.\mathcal{M}_2\right)}\right.}=\left\{\nabla \ell_n\left(\widehat{S}_{\mathcal{M}_2}+\widehat{L}_{\mathcal{M}_2}\right)+\gamma_n\left(\operatorname{sign}\left(\mathbf{O}\left(S^*\right)\right)\right)+\gamma_n M:\|M\|_{\infty} \leq 1 \text { and } M \in \Omega^{* \perp}\right\},
$$
where $\Omega^{* \perp}$ is the orthogonal complement space of $\Omega^*$ in the space of symmetric matrices. According to Example 2 of \cite{watson1992characterization}, the sub-differential with respect to $L$ is
$$
\begin{aligned}
\left.\partial_L H\right|_{\left(\widehat{S}_{\mathcal{M}_2}, \widehat{L}_{\mathcal{M}_2}\right)} 	=&\left\{\nabla \ell_n\left(\widehat{S}_{\mathcal{M}_2}+\widehat{L}_{\mathcal{M}_2}\right)+\delta_n \widehat{U}_{1, \mathcal{M}_2} \widehat{U}_{1, \mathcal{M}_2}^{\top}+ \tau_n W_n\odot \operatorname{sign}(L^*) \right.\\
&\left.	+ \delta_n \widehat{U}_{2, \mathcal{M}_2} M_1 \widehat{U}_{2, \mathcal{M}_2}^{\top} +\tau_n W_n\odot M_2:\|M_1\|_2\leq 1, \|M_2\|_{\infty}\leq 1, M_2\in \maT_2^{*\perp}\right\}
\end{aligned}
$$
where  $\widehat{U}_{1, \mathcal{M}_2}$ has the same zero-block as $ U_1^*$, $\widehat{U}_{2, \mathcal{M}_2}$ is a $p \times(p-r)$ matrix satisfying $\widehat{U}_{1, \mathcal{M}_2}^{\top} \widehat{U}_{2, \mathcal{M}_2}=\mathbf{0}_{r \times(p-r)}, \widehat{U}_{2, \mathcal{M}_2}^{\top} \widehat{U}_{2, \mathcal{M}_2}=I_{p-r}$ and $\left\|\widehat{U}_{2, \mathcal{M}_2}-U_2^*\right\|_{\infty} \leq \gamma_n^{1-\eta}$. For some $(S, L)$, if
\begin{equation}\label{eq.KKT}
	\mathcal{P}_{\Omega^*}(S)=\mathbf{0}_{p \times p}, \mathcal{P}_{\Omega^{* \perp}}(S)=\mathbf{0}_{p \times p}, \mathcal{P}_{T({\widehat{L}_{\mathcal{M}_2}})} (L)=\mathbf{0}_{p \times p} \text { and } \mathcal{P}_{\maT^\perp({\widehat{L}_{\mathcal{M}_2}})}(L)=\mathbf{0}_{p \times p},
\end{equation}
then $S=\mathbf{0}_{p \times p}$ and $L=\mathbf{0}_{p \times p}$. Consequently, to prove (\ref{eq.KKT}), it suffices to show that
\begin{equation}\label{eq.KKTprojection}
	\left.\mathcal{P}_{\Omega^*} \partial_S H\right|_{\left(\widehat{S}_{\mathcal{M}_2}, \widehat{L}_{\mathcal{M}_2}\right)}=\mathbf{0}_{p \times p} \text { and }\left.\mathcal{P}_{\maT(\widehat{L}_{\mathcal{M}_2})} \partial_L H\right|_{\left(\widehat{S}_{\mathcal{M}_2}, \widehat{L}_{\mathcal{M}_2}\right)}=\mathbf{0}_{p \times p}.
\end{equation}
and
\begin{equation}\label{eq.KKTOrthogprojection}
	\left.\left.\mathbf{0}_{p \times p} \in \mathcal{P}_{\Omega^{* \perp}} \partial_S H\right|_{\left(\widehat{S}_{\mathcal{M}_2}, \widehat{L}_{\mathcal{M}_2}\right)} \text { and } \mathbf{0}_{p\times p} \in \mathcal{P}_{\maT^\perp(\widehat{L}_{\mathcal{M}_2})} \partial_L H\right|_{\left(\widehat{S}_{\mathcal{M}_2}, \widehat{L}_{\mathcal{M}_2}\right)} \text {. }
\end{equation}
According to the definition of $\left(\widehat{S}_{\mathcal{M}_2}, \widehat{L}_{\mathcal{M}_2}\right)$, it is the solution to the optimization (\ref{eq.objconstriantM2}). In addition, according to the discussion in Step 1, $\left(\widehat{S}_{\mathcal{M}_2}, \widehat{L}_{\mathcal{M}_2}\right)$ does not lie on the boundary of $\mathcal{M}_2$. Therefore, it satisfies the first order condition of (\ref{eq.objconstriantM2}), which is equivalent to (\ref{eq.KKTprojection}). Thus, to prove \ref{eq.KKT}, it is sufficient to show (\ref{eq.KKTOrthogprojection}). Lemma \ref{L.LSubgradient} establishes a sufficient expression for the above equation.
 Take gradient on both side of (\ref{eq.taylorexpansion}) to obtain
$$
\nabla \ell_n\left(S^*+L^*+\Delta\right)=\mathcal{I}^* v(\Delta)+\nabla R_n(\Delta).
$$
We plug $\Delta=\widehat{S}_{\mathcal{M}_2}+\widehat{L}_{\mathcal{M}_2}-S^*-L^*$ into the above equation to get
\begin{equation}\label{eq.PluginGradient}
	\nabla \ell_n\left(\widehat{S}_{\mathcal{M}_2}+\widehat{L}_{\mathcal{M}_2}\right)=\mathcal{I}^* v\left(\widehat{S}_{\mathcal{M}_2}+\widehat{L}_{\mathcal{M}_2}-S^*-L^*\right)+\nabla R_n\left(\widehat{S}_{\mathcal{M}_2}+\widehat{L}_{\mathcal{M}_2}-S^*-L^*\right) .
\end{equation}
According to Lemma \ref{L.SolutionError},
$$
\widehat{S}_{\mathcal{M}_2}+\widehat{L}_{\mathcal{M}_2}-S^*-L^*=\mathcal{A} \mathbf{G}^{-1}\left(\gamma_n \operatorname{sign}\left(\mathbf{O}\left(S^*\right)\right), \delta_n U_1^* U_1^{* \top}\right)+o_P\left(\gamma_n\right),
$$
where $\mathcal{A}$ is the adding operator of two matrices $\mathcal{A}(A, B)=A+B$. Combining this with (\ref{eq.FirstderivativeRn}), (\ref{eq.PluginGradient}), and notice that $\delta_n=\rho_1 \gamma_n$, we have
$$
\nabla \ell_n\left(\widehat{S}_{\mathcal{M}_2}+\widehat{L}_{\mathcal{M}_2}\right)=\gamma_n \mathcal{I}^* \mathcal{A} \mathbf{G}^{-1}\left(\operatorname{sign}\left(\mathbf{O}\left(S^*\right)\right), \rho_1 U_1^* U_1^{* \top}\right)+o_P\left(\gamma_n\right),
$$
and consequently,
\begin{equation}\label{eq.DualOmega}
	\mathcal{P}_{\Omega^{* \perp}} \nabla \ell_n\left(\widehat{S}_{\mathcal{M}_2}+\widehat{L}_{\mathcal{M}_2}\right)=\gamma_n \mathcal{P}_{\Omega^{* \perp}} \mathcal{I}^* \mathcal{A} \mathbf{G}^{-1}\left(\operatorname{sign}\left(\mathbf{O}\left(S^*\right)\right), \rho_1 U_1^* U_1^{* \top}\right)+o_P\left(\gamma_n\right),
\end{equation}
\begin{equation}\label{eq.DualV}
	\mathcal{P}_{\maV(\widehat{L}_{\mathcal{M}_2})} \nabla \ell_n\left(\widehat{S}_{\mathcal{M}_2}+\widehat{L}_{\mathcal{M}_2}\right)=\gamma_n \mathcal{P}_{\maV^*} \mathcal{I}^* \mathcal{A} \mathbf{G}^{-1}\left(\operatorname{sign}\left(\mathbf{O}\left(S^*\right)\right), \rho_1 U_1^* U_1^{* \top}\right)+o_P\left(\gamma_n\right).
\end{equation}
\begin{equation}\label{eq.DualT2}
	\mathcal{P}_{\maT_2^{*\perp}} \nabla \ell_n\left(\widehat{S}_{\mathcal{M}_2}+\widehat{L}_{\mathcal{M}_2}\right)=\gamma_n \mathcal{P}_{\maT_2^{*\perp}} \mathcal{I}^* \mathcal{A} \mathbf{G}^{-1}\left(\operatorname{sign}\left(\mathbf{O}\left(S^*\right)\right), \rho_1 U_1^* U_1^{* \top}\right)+o_P\left(\gamma_n\right).
\end{equation}
\begin{equation}\label{eq.DualT}
	\mathcal{P}_{\maT^\perp(\widehat{L}_{\mathcal{M}_2})} \nabla \ell_n\left(\widehat{S}_{\mathcal{M}_2}+\widehat{L}_{\mathcal{M}_2}\right)=\gamma_n \mathcal{P}_{\maT^{*\perp}} \mathcal{I}^* \mathcal{A} \mathbf{G}^{-1}\left(\operatorname{sign}\left(\mathbf{O}\left(S^*\right)\right), \rho_1 U_1^* U_1^{* \top}\right)+o_P\left(\gamma_n\right).
\end{equation}
According to Lemma \ref{L.LSubgradient}, it suffices to show that 
\begin{equation}\label{eq.SufficientOrthogOmega}
\|\mathcal{P}_{\Omega^{* \perp}} \nabla \ell_n\left(\widehat{S}_{\mathcal{M}_2}+\widehat{L}_{\mathcal{M}_2}\right)\|_{\infty}\leq \gamma_n, \quad 
\end{equation}
\begin{equation}\label{eq.SufficientOrthogT}
g_{\rho_1, \rho_2,\widehat{L}_{\mathcal{M}_2},\widetilde{W}_n}\left(\mathcal{P}_{\maT^\perp(\widehat{L}_{\mathcal{M}_2})} \nabla \ell_n\left(\widehat{S}_{\mathcal{M}_2}+\widehat{L}_{\mathcal{M}_2}\right) - \tau_n\mathcal{P}_{\maT^\perp(\widehat{L}_{\mathcal{M}_2})}\left(W_n\odot\operatorname{sign}(L) \right)\right)\leq \gamma_n.
\end{equation}
The first inequality is easy to check according to (\ref{eq.DualOmega}) and Assumption \ref{A.irrepresentable}. We emphasize the second part. Notice that from \eqref{eq.DualV}, \eqref{eq.DualT2}, \eqref{eq.DualT} and Lemma \ref{L.WeightOrder}, we have
\begin{equation}\label{eq.ProjV}
	\begin{aligned}
		&\mathcal{P}_{\maV(\widehat{L}_{\mathcal{M}_2})} \left(\nabla \ell_n\left(\widehat{S}_{\mathcal{M}_2}+\widehat{L}_{\mathcal{M}_2}\right) - \tau_n\mathcal{P}_{\maT^\perp(\widehat{L}_{\mathcal{M}_2})}\left(W_n\odot\operatorname{sign}(\widehat{L}_{\mathcal{M}_2}) \right)\right)\\
		= &\gamma_n \mathcal{P}_{\maV^*} \mathcal{I}^* \mathcal{A} \mathbf{G}^{-1}\left(\operatorname{sign}\left(\mathbf{O}\left(S^*\right)\right), \rho_1 U_1^* U_1^{* \top}\right)+o_P\left(\gamma_n\right) + o_P(r_n^{1-\alpha} \frac{\gamma_n}{r^{1-\alpha}_n})\\
		=&\gamma_n \mathcal{P}_{\maV^*} \mathcal{I}^* \mathcal{A} \mathbf{G}^{-1}\left(\operatorname{sign}\left(\mathbf{O}\left(S^*\right)\right), \rho_1 U_1^* U_1^{* \top}\right)+o_P(\gamma_n),
	\end{aligned}
\end{equation}
and
\begin{equation}\label{eq.ProjT2}
	\begin{aligned}
		&\mathcal{P}_{\maT_2^{*\perp}} \left(\nabla \ell_n\left(\widehat{S}_{\mathcal{M}_2}+\widehat{L}_{\mathcal{M}_2}\right) - \tau_n\mathcal{P}_{\maT^\perp(\widehat{L}_{\mathcal{M}_2})}\left(W_n\odot\operatorname{sign}(\widehat{L}_{\mathcal{M}_2}) \right)\right)\\
		= &\gamma_n \mathcal{P}_{\maT_2^{*\perp}} \mathcal{I}^* \mathcal{A} \mathbf{G}^{-1}\left(\operatorname{sign}\left(\mathbf{O}\left(S^*\right)\right), \rho_1 U_1^* U_1^{* \top}\right)+o_P\left(\gamma_n\right) + o_P(r_n^{1-\alpha} \frac{\gamma_n}{r^{1-\alpha}_n})\\
		=&\gamma_n \mathcal{P}_{\maT_2^{*\perp}} \mathcal{I}^* \mathcal{A} \mathbf{G}^{-1}\left(\operatorname{sign}\left(\mathbf{O}\left(S^*\right)\right), \rho_1 U_1^* U_1^{* \top}\right)+o_P(\gamma_n).
	\end{aligned}
\end{equation}
and
\begin{equation}\label{eq.ProjT}
	\begin{aligned}
		&\mathcal{P}_{\maT^\perp(\widehat{L}_{\mathcal{M}_2})} \left(\nabla \ell_n\left(\widehat{S}_{\mathcal{M}_2}+\widehat{L}_{\mathcal{M}_2}\right) - \tau_n\mathcal{P}_{\maT^\perp(\widehat{L}_{\mathcal{M}_2})}\left(W_n\odot\operatorname{sign}(\widehat{L}_{\mathcal{M}_2}) \right)\right)\\
		= &\gamma_n \mathcal{P}_{T^{*\perp}} \mathcal{I}^* \mathcal{A} \mathbf{G}^{-1}\left(\operatorname{sign}\left(\mathbf{O}\left(S^*\right)\right), \rho_1 U_1^* U_1^{* \top}\right)+o_P\left(\gamma_n\right) + o_P(r_n^{1-\alpha} \frac{\gamma_n}{r^{1-\alpha}_n})\\
		=&\gamma_n \mathcal{P}_{T^{*\perp}} \mathcal{I}^* \mathcal{A} \mathbf{G}^{-1}\left(\operatorname{sign}\left(\mathbf{O}\left(S^*\right)\right), \rho_1 U_1^* U_1^{* \top}\right)+o_P(\gamma_n).
	\end{aligned}
\end{equation}
Then according to $\max\{a, b+c\}\leq \max\{a, b\} +c $,
$$
\begin{aligned}
&	g_{\rho_1, \rho_2,\widehat{L}_{\mathcal{M}_2},\widetilde{W}_n}\left(\mathcal{P}_{\maT^\perp(\widehat{L}_{\mathcal{M}_2})} \nabla \ell_n\left(\widehat{S}_{\mathcal{M}_2}+\widehat{L}_{\mathcal{M}_2}\right) - \tau_n\mathcal{P}_{\maT^\perp(\widehat{L}_{\mathcal{M}_2})}\left(W_n\odot\operatorname{sign}(\widehat{L}_{\mathcal{M}_2}) \right)\right)\\
\leq &g_{\rho_1, \rho_2,\widehat{L}_{\mathcal{M}_2},\Lambda_n}\left(\mathcal{P}_{\maT^\perp(\widehat{L}_{\mathcal{M}_2})} \nabla \ell_n\left(\widehat{S}_{\mathcal{M}_2}+\widehat{L}_{\mathcal{M}_2}\right) - \tau_n\mathcal{P}_{\maT^\perp(\widehat{L}_{\mathcal{M}_2})}\left(W_n\odot\operatorname{sign}(\widehat{L}_{\mathcal{M}_2}) \right)	\right)\\
& + \frac{r_n^{1-\alpha}\max_{(i,j)\not\in J_1}|(\widetilde{W}_n)_{ij} - (\Lambda_n)_{ij}|}{\rho_2}\left\|\mathcal{P}_{\maT^\perp(\widehat{L}_{\mathcal{M}_2})} \nabla \ell_n\left(\widehat{S}_{\mathcal{M}_2}+\widehat{L}_{\mathcal{M}_2}\right) - \tau_n\mathcal{P}_{\maT^\perp(\widehat{L}_{\mathcal{M}_2})}\left(W_n\odot\operatorname{sign}(\widehat{L}_{\mathcal{M}_2}) \right)	\right\|_F\\
\triangleq & \uppercase\expandafter{\romannumeral1} + \uppercase\expandafter{\romannumeral2}.
\end{aligned}
$$
For $\uppercase\expandafter{\romannumeral1}$, by \eqref{eq.ProjV}, \eqref{eq.ProjT2}, Assumption \ref{A.irrepresentable} and Lemma \ref{L.AngleL}, we have
$$
\begin{aligned}
	\uppercase\expandafter{\romannumeral1}& \leq \gamma_ng_{\rho_1,\rho_2,L^*,\Lambda_n}\left( \mathcal{P}_{\maT^{*\perp}} \mathcal{I}^* \mathcal{A} \mathbf{G}^{-1}\left(\operatorname{sign}\left(\mathbf{O}\left(S^*\right)\right), \rho_1 U_1^* U_1^{* \top}\right)\right) + \left(\frac{1}{\rho_1} + \frac{r_n^{1-\alpha}M_{n2}}{\rho_2} \right)o_P\left(\gamma_n\right)\\
	&\leq  \gamma_ng_{\rho_1,\rho_2,L^*,\Lambda_n}\left(\mathcal{P}_{\maT^{*\perp}} \mathcal{I}^* \mathcal{A} \mathbf{G}^{-1}\left(\operatorname{sign}\left(\mathbf{O}\left(S^*\right)\right), \rho_1 U_1^* U_1^{* \top}\right) \right) + o_P(\gamma_n)\\
	&<\gamma_n + o_P(\gamma_n), 
\end{aligned}
$$
For $\uppercase\expandafter{\romannumeral2}$, from Assumption \ref{A.Wn} and \eqref{eq.ProjT}, we have
$$
\begin{aligned}
	\uppercase\expandafter{\romannumeral2} & =\frac{r_n^{1-\alpha}\max_{(i,j)\not\in J_1}|(\widetilde{W}_n)_{ij} - (\Lambda_n)_{ij}|}{\rho_2}\left\|\gamma_n \mathcal{P}_{\maT^{*\perp}} \mathcal{I}^* \mathcal{A} \mathbf{G}^{-1}\left(\operatorname{sign}\left(\mathbf{O}\left(S^*\right)\right), \rho_1 U_1^* U_1^{* \top}\right)+o_P\left(\gamma_n\right)	\right\|_F\\
	& \leq r_n^{1-\alpha}O_P(r_n^{-1})O_P(\gamma_n) = o_P(\gamma_n),
\end{aligned}
$$
which completes the proof for equation (\ref{eq.SufficientOrthogT}). Then with probability tending to 1 as $n\to \infty$, 
\[\uppercase\expandafter{\romannumeral1}+ \uppercase\expandafter{\romannumeral2}\leq \gamma_n.   \]
Thus far we have proved that with probability tending to 1, $\left(\widehat{S}_{\mathcal{M}_2}, \widehat{L}_{\mathcal{M}_2}\right)$ is a solution to the optimization problem (\ref{eq.objconstraintM1}).
 We proceed to the proof of the uniqueness of the solution to (\ref{eq.objconstraintM1}). Because the objective function $H(S, L)$ is a convex function, it is sufficient to show the uniqueness of the solution in a neighborhood of $\left(\widehat{S}_{\mathcal{M}_2}, \widehat{L}_{\mathcal{M}_2}\right)$. Let $d_n = e^{-n}\ll\gamma_n$,  we  choose a small neighborhood as follows:
$$
\begin{aligned}
	\mathcal{N}=&\left\{(S, L): S = S_{\Omega^*} + S_{\Omega^{*\perp}},  \left\|S_{\Omega^*}-\widehat{S}_{\mathcal{M}_2}\right\|_{\infty}<d_n,\left\|S_{\Omega^{*\perp}}\right\|_{\infty}<d_n, S_{\Omega^*}\in\Omega^*, S_{\Omega^{*\perp}}\in\Omega^{*\perp},\right.\\
	& L \text{ has the following eigendecomposition }:\\
    & L=\left[U_1, U_2\right]\left[\begin{array}{cc}\Sigma_1 & \mathbf{0}_{p\times(p-r)} \\ \mathbf{0}_{(p-r) \times p} & \Sigma_2\end{array}\right]\left[U_1, U_2\right]^{\top}, U_1 = U_{11} + U_{11}^{\perp},\\
    & (U_{11})_{G_{ij}^*} = 0, \text{ for } i\not= j,\quad (U_{11}^\perp)_{G_{ij}^*} = 0, \text{ for } i= j, \quad i,j\in[m],\\
	&\left\|U_{11}-\widehat{U}_{1, \mathcal{M}_2}\right\|_{\infty}<d_n,\left\|U_{11}^\perp\right\|_{\infty}<d_n,\left\|U_2-\widehat{U}_{2, \mathcal{M}_2}\right\|_{\infty}<d_n, \\
	&\left.
	\left\|\Sigma_1-\widehat{\Sigma}_{1, \mathcal{M}_2}\right\|_{\infty}<d_n,\left\|\Sigma_2\right\|_{\infty}<d_n.
	\right\}.
\end{aligned}
$$

%    &U_{11}= \left[(V_{11}^{T},\mathbf{ 0}_{ d_1\times(p-d_1)})^T,\cdots, (\mathbf{0}_{ d_m\times(p-d_m)},V_{m1}^{T} )^T\right], V_{i1}\in\mathbb{R}^{d_1\times d_1},\\
%&U_{11}^\perp= \left[(\mathbf{ 0}_{ d_1\times d_1},V_{12}^{T})^T,\cdots, (V_{m2}^{T},\mathbf{0}_{ d_m\times d_m} )^T\right],V_{i2}\in\mathbb{R}^{d_m\times d_m},\\

The next lemma, together with the uniqueness of the solution to (\ref{eq.objconstriantM2}) established in Lemma \ref{L.SolutionError}, guarantees that $\left(\widehat{S}_{\mathcal{M}_2}, \widehat{L}_{\mathcal{M}_2}\right)$ is the unique solution in $\mathcal{N}$.

\begin{lemma}\label{L.UniqueSolution}
	 For all $\left(\tilde{S}, \tilde{L}\right) \in \mathcal{N}$, if $\left(\tilde{S}, \tilde{L}\right)$ is a solution to \ref{eq.objconstraintM1}, then $\left(\tilde{S}, \tilde{L}\right) =(\widehat{S}_{\mathcal{M}_2}, \widehat{L}_{\mathcal{M}_2})$  with probability tending to 1.
\end{lemma}

%\subsection{Proof of Theorem \ref{Thm.LatentA}}
%\begin{proof}
%  Note that $L^* = A^*A^{*T} = U_1^*\Sigma_1^* U_1^*$, then in general, there exists $O^*\in\mathcal{O}^{r\times r}$ such that $A^*O^* = U_1^*\Sigma_1^{*\frac{1}{2}}$. In addition, from the proof of Theorem \ref{Thm.Main}, $(\widehat{S}, \widehat{L})\in\mathcal{M}_2$, then
%  \begin{equation*}
%  	\begin{aligned}
%  		  \|\widehat{U}_1\widehat{\Sigma}_1^{\frac{1}{2}} - U_1^*\Sigma_1^{*\frac{1}{2}}\|_{\infty} &\leq (\widehat{U}_1 - \widehat{U}_1^*)\widehat{\Sigma}_1^{\frac{1}{2}}+ \widehat{U}_1^*(\widehat{\Sigma}_1^{\frac{1}{2}} - \Sigma_1^{*\frac{1}{2}})  \\
%  		  & =(\widehat{U}_1 - \widehat{U}_1^*)\widehat{\Sigma}_1^{\frac{1}{2}}+ \widehat{U}_1^* (\widehat{\Sigma}_1- \Sigma_1^*)(\widehat{\Sigma}_1^{\frac{1}{2}}+ \Sigma_1^{*\frac{1}{2}})^{-1} \\
%  		  &\lesssim_P \gamma_n^{1-\eta} + \gamma_n^{1-2\eta}\lesssim_P \gamma_n^{1-2\eta},
%  	\end{aligned}  
%  \end{equation*}
%  where the second inequality follows from the fact that $\widehat{\Sigma}_1$, $\Sigma^*_1$ are diagonal matrices. Moreover, according to proof of Theorem \ref{Thm.Main}, $\widehat{A}_{G_{ij}^{*\pi}} =(\widehat{U}_1)_{ G^*_{ij}} =0 $ for $i\not=j$, which concludes the proof.
%\end{proof}

\subsection{Proof of Theorem \ref{Thm.Hammingerror}}
This proof is similar to Proposition 19 in \cite{ma2020universal}.
Denote $
\hat{\mu}_j = \frac{1}{|\{i:\hat{\bt}(i) = j\}|}\sum_{i:\hat{\bt}(i) = j}\hat{l}_i
$, then from defintion of equation \eqref{eq.cluster},
$$
\sum_{j = 1}^m\sum_{i:\hat{\bt}(i) = j}\|\hat{l}_i-\hat{\mu}_j\|^2 \leq  \sum_{j = 1}^m\sum_{i:\hat{\bt}(i) = j}\|\hat{l}_i-\mu^*_j\|^2.
$$
Let $S = \{i: \|\hat{\mu}_{\hat{\bt}(i)} - \mu^*_{\bt(i)} \|\ge c\} $, then 
\begin{align*}
	\frac{1}{p}|S|  & \leq \frac{1}{pc^2}\sum_{i\in S} \|\hat{\mu}_{\hat{\bt}(i)} - \mu^*_{\bt(i)} \|^2\\
	& \leq \frac{2}{pc^2}\left(\sum_{i=1}^p\|\hat{l}_i - \hat{\mu}_{\hat{\bt}(i)}\|^2 +  \|\hat{l}_i - \mu^*_{\bt(i)}\|^2 \right)\\
	& \leq \frac{4}{pc^2}\sum_{i=1}^p \|\hat{l}_i - \mu^*_{\bt(i)}\|^2 \\
	& \leq \frac{4}{pc^2}\sum_{i=1}^p \left( \|\hat{l}_i - l_i^*\|^2 + \| \hat{l}_i - \mu^*_{\bt(i)}\|^2   \right)\\
	& = \frac{4}{pc^2} \left( \|\widehat{L} - L^*\|_F^2 + \sum_{i=1}^p\| \hat{l}_i - \mu^*_{\bt(i)}\|^2   \right)
\end{align*}
In addition, similar to the proof of Proposition 19 in \cite{ma2020universal}, we have $H(\hat{\boldsymbol{t}}, \pi(\boldsymbol{t}))\leq \frac{C\omega}{p}|S|$, which completes the proof.

\subsection{Proof of Corollary \ref{Cor.CorHamming}}
From definition of correlation, for $k = l$, 
$$
\operatorname{cor}(\operatorname{abs}(l_i^*), \operatorname{abs}(l_j^*))  = \frac{\frac{1}{d_k}\sum_{h = 1}^{d_k}|u_{ki}u_{kj}||u_{kh}^2| - |u_{ki}u_{kj}|\bar{u}_k^2}{|u_{ki}u_{kj}|\|U_{1i}\|^2} = \frac{1}{d_k} - \bar{u}_k^2.
$$
For $k \not= l$, it follows from the orthogonal property of $L_k^*, L^*_l$. Furthermore, according to Theorem \ref{Thm.Main} and continuity of $\operatorname{Cor}(\operatorname{abs}(\cdot))$, $\|\operatorname{Cor}(\operatorname{abs}(\widehat{L}))-\operatorname{Cor}(\operatorname{abs}(L^*))\|_F = o_P(1) $. Combing the above equality and Theorem \ref{Thm.Hammingerror} completes the proof.

\subsection{Proof of Lemmas}\label{Appen.Lemmaproof}
\subsubsection{Proof of Lemma \ref{L.WeightOrder}}
\begin{proof}
	The proof is similar to Lemma 2 in \cite{huang2008adaptive}. Since $M_{n 1}=o\left(r_n^{1-\alpha}\right)$, and $\max _{(i,j) \in J_{1}}|(\widetilde{W}_n)_{ij}|/| (\Lambda_{n})_{ij}|-1| \leq M_{1 n} O_P\left(1 / r_n\right)=o_P(1)$ by the $r_n$-consistency of $(\widetilde{W}_n)_{ij}$. Thus, $\left\|\mathbf{s}_{n 1}\right\|=\left(1+o_P(1)\right) M_{n 1} = o_P(r_n^{1-\alpha})$. 
\end{proof}

\subsubsection{Proof of Lemma \ref{L.DirectDecomp}}
\begin{proof}
	(1). The first part is obvious by triangular inequality and the uniqueness decomposition of the direct sum.\\
	(2). According to Lemma \ref{L.Lipschitz} and Proposition \ref{P.TangentSpace}, we have
	\[  \|(\mathcal{P}_{\maV(L)} - \mathcal{P}_{\maV^*})(M)\|_2 \lesssim\|L-L^*\|_2\|M\|.\]
\end{proof}

\subsubsection{Proof of Lemma \ref{L.LSubgradient}}
\begin{proof}
	The fist part follows from expressions of subgradients for $\|\cdot\|_1, \|\cdot\|_*$ in \citep{candes2013simple} and  the additivity of the subgradient. Second,
	from the condition, we  can rewrite 
	$ Z = \delta_n UV^T + \tau_nW_n\odot\operatorname{sign}(L) + F$, where $F = F_1 + W_n\odot F_2, F_1\in V\subseteq \maT_1(^{\perp}(L), F_2 \in \maT_2^\perp(L)$  and $g^L_{\rho_1,\rho_2,\widetilde{W}_n}(F)\leq \gamma_n$. It implies $\|F_1\|_{2}\leq\delta_n$, $\|F_2\|_{\infty}\leq \tau_n$. Then $Z\in \partial(\delta_n\|L\|_* + \tau_n\|W_n\odot L\|_1)$.
\end{proof}

\subsubsection{Proof of Lemma \ref{L.FInvertible}}
\begin{proof}
	The proof is omitted here by referring to Lemma 1 in \cite{chen2016fused}. 
%	In addition, it can also be viewed as a corollary of the conclusion (1) of the Theorem \ref{Appen.Theorem} in Appendix \ref{Appen.Conditioncheck}. 
\end{proof}

\subsubsection{Proof of Lemma \ref{L.AngleL}}\label{Appen.ProProof.Angle}
\begin{proof}
	The constants specified in this lemma are as follows,
	$$l_n = \frac{1}{\rho_1 + \rho_2 \frac{p}{m_{n2}r_n^{1-\alpha}}},\quad  u_n = \frac{1}{\rho_1} + \frac{r_n^{1-\alpha}M_{n2}}{\rho_2},$$
	where  $m_{2n} \triangleq \min_{(i,j)\notin J_1}(\Lambda_n)_{ij}$ satisfying $ m_{n2}r_n^{1-\alpha} \leq M_{n2}r_n^{1-\alpha} = O_P(1) $ as $n\to\infty$.
	\par First, note that for any given smooth point $L\in\mathcal{LS}(m,r)$ with $\maT^\perp(L) = \maV(L)\oplus \maT_2^{*\perp}$, $L'\in \maT^\perp(L) $ has the orthogonal direct sum decomposition $L' = L_1 \oplus L_2$ with $L_1 = \mathcal{P}_{\maV(L)}(L'), L_2 =\mathcal{P}_{\maT_2^{*\perp}}(L') $. Then
	$$
	 \|L'\|_F^2 =	\|L_1 + L_2\|_F^2 = \|L_1\|_F^2 + \|L_2\|_F^2.
	$$
	\par Second, we prove the norm inequality. The right part of the inequality is a corollary of the above inequality for $L$. For the left part, note that for any  $a\in [0,1]$,
	$$
	\begin{aligned}
		g_{\rho_1, \rho_2, L,\Lambda_n}(L') &\ge \frac{a}{\rho_1}\|L_1\|_2 + \frac{(1-a)r_n^{1-\alpha}}{\rho_2}\|\Lambda_n\odot L_2\|_\infty\\
		& \ge  \frac{a}{\rho_1}\|L_1\|_2 + \frac{(1-a)r_n^{1-\alpha}m_{n2}}{\rho_2p}\|L_2\|_2.
	\end{aligned}
	$$
	Take $ a = \frac{m_{n2}r_n^{1-\alpha}\rho_1}{p\rho_2+m_{n2}r_n^{1-\alpha}\rho_1}
	$, then 
	\[
	g^L_{\rho_1, \rho_2, \Lambda_n}(L') \ge l_n\|L\|_2.
	\]
	This concludes the proof.
	
\end{proof}

\subsubsection{Proof of Lemma \ref{L.Dconstraint}}
\begin{proof}
	We consider the first-order condition for the optimization problem (\ref{eq.objconstraintD}). Notice that $\Omega^*$ and $\widehat{\mathcal{D}}$ are linear spaces, so the first order condition becomes
	$$
	\left.\mathbf{0}_{p \times p} \in \mathcal{P}_{\Omega^*} \partial_S H\right|_{(S, L)} \text { and }\left.\mathbf{0}_{p \times p} \in \mathcal{P}_{\widehat{\mathcal{D}}} \partial_L H\right|_{(S, L)},
	$$
	where $H$ is defined as
	$$
	H(S, L)=\ell_n(S+L)+\gamma_n \|S\|_{\text{off},1}  + \delta_n\|L\|_*+ \tau_n\|W_n\odot L\|_1.
	$$
	We will show that there is a unique $(S, L) \in \Omega^* \times \widehat{\mathcal{D}}$ satisfying the first order condition. Because of the convexity of the optimization problem (\ref{eq.objconstraintD}), it suffices to show that with a probability converging to 1, there is a unique $(S, L) \in \mathcal{B}$ satisfying the first order condition, where
	$$
	\mathcal{B}=\left\{(S, L) \in \Omega^* \times \widehat{\mathcal{D}}:\left\|S-S^*\right\|_{\infty} \leq \gamma_n^{1-\eta} \text {, and }\left\|L-L^*\right\|_{\infty} \leq \gamma_n^{1-\eta}\right\} .
	$$
	We simplify the first order condition for $(S, L) \in \mathcal{B}$. For the $\ell_1$ penalty term for $S$, if $\left\|S-S^*\right\|_{\infty} \leq$ $\gamma_n^{1-\eta}$ and $S \in \Omega^*$, then $\|S\|_{\text{off},1}$ is smooth on $\Omega^*$ and
	$$
	\mathcal{P}_{\Omega^*} \partial_S\|S\|_{\text{off},1}=\operatorname{sign}\left(\mathbf{O}\left(S^*\right)\right) \text { for } S \in \Omega^* .
	$$
	Similarly, for $L \in \widehat{\mathcal{D}}$ and $\left\|L-L^*\right\|_{\infty} \leq \gamma_n^{1-\eta},\|L\|_*, \|W_n\odot L\|_1$ is smooth over the linear space $\widehat{\mathcal{D}}$ and
	$$
	\mathcal{P}_{\widehat{\mathcal{D}}} \partial_L\|L\|_*= \widehat{U}_{1, \mathcal{M}_2} \widehat{U}_{1, \mathcal{M}_2}^{\top},  \text { for } L \in \widehat{\mathcal{D}} \text {, }
	$$
	and
	$$
	\mathcal{P}_{\widehat{\mathcal{D}}} \partial_L\|W_n\odot L\|_1=\mathcal{P}_{\widehat{\mathcal{D}}}( W_n\odot \operatorname{sign}(L^*)). 
	$$
	Combining the above two equations with the $\nabla \ell_n$ term, we arrive at an equivalent form of the first order condition, that is, there exists $(S, L) \in \mathcal{B}$ satisfying
	$$
	\begin{gathered}
		\mathcal{P}_{\Omega^*} \nabla \ell_n(S+L)+\gamma_n \operatorname{sign}\left(\mathbf{O}\left(S^*\right)\right)=\mathbf{0}_{p \times p}, \\
		\mathcal{P}_{\widehat{\mathcal{D}}} \nabla \ell_n(S+L)+\delta_n \widehat{U}_{1, \mathcal{M}_2} \widehat{U}_{1, \mathcal{M}_2}^{\top} + \tau_n\mathcal{P}_{\widehat{\mathcal{D}}}( W_n\odot \operatorname{sign}(L^*)) =\mathbf{0}_{p \times p} .
	\end{gathered}
	$$
	We will show the existence and uniqueness of the solution to the above equations using the contraction mapping theorem. We first construct the contraction operator. Let $(S, L)=$ $\left(S^*+\Delta_S, L^*+\Delta_L\right)$. We plug (\ref{eq.PluginGradient}) into the above equations, and arrive at their equivalent ones
	\begin{equation}\label{eq.PluginDconstraint}
			\begin{aligned}
			& \mathcal{P}_{\Omega^*} \mathcal{I}^*\left(\Delta_S+\Delta_L\right)+\mathcal{P}_{\Omega^*} \nabla R_n\left(\Delta_S+\Delta_L\right)+\gamma_n \operatorname{sign}\left(\mathbf{O}\left(S^*\right)\right)=\mathbf{0}_{p \times p}, \\
			& \mathcal{P}_{\widehat{\mathcal{D}}} \mathcal{I}^*\left(\Delta_S+\Delta_L\right)+\mathcal{P}_{\widehat{\mathcal{D}}} \nabla R_n\left(\Delta_S+\Delta_L\right)+\delta_n \widehat{U}_{1, \mathcal{M}_2} \widehat{U}_{1, \mathcal{M}_2}^{\top} +\tau_n\mathcal{P}_{\widehat{\mathcal{D}}}( W_n\odot \operatorname{sign}(L^*))   =\mathbf{0}_{p \times p} . 
		\end{aligned}
	\end{equation}
	We define an operator $\tilde{\mathbf{G}}_{\widehat{\mathcal{D}}}: \Omega^* \times\left(\widehat{\mathcal{D}}-L^*\right) \rightarrow \Omega^* \times \widehat{\mathcal{D}}$,
	$$
	\tilde{\mathbf{G}}_{\widehat{\mathcal{D}}}\left(\Delta_S, \Delta_L\right)=\left(\mathcal{P}_{\Omega^*} \mathcal{I}^*\left(\Delta_S+\Delta_L\right), \mathcal{P}_{\widehat{\mathcal{D}}} \mathcal{I}^*\left(\Delta_S+\Delta_L\right)\right),
	$$
	where the set $\widehat{\mathcal{D}}-L^*=\left\{L-L^*: L \in \widehat{\mathcal{D}}\right\}$. We further transform equation (\ref{eq.PluginDconstraint}) to
	\begin{equation}\label{eq.PluginDconstraintTransform}
		\begin{aligned}
			& \tilde{\mathbf{G}}_{\widehat{\mathcal{D}}}\left(\Delta_S, \Delta_L\right)+\left(\mathcal{P}_{\Omega^*} \nabla R_n\left(\Delta_S+\Delta_L\right), \mathcal{P}_{\widehat{\mathcal{D}}} \nabla R_n\left(\Delta_S+\Delta_L\right)\right)+\left(\gamma_n \operatorname{sign}\left(\mathbf{O}\left(S^*\right)\right), \delta_n \widehat{U}_{1, \mathcal{M}_2} \widehat{U}_{1, \mathcal{M}_2}^{\top} \right.\\
			&\left. +\tau_n\mathcal{P}_{\widehat{\mathcal{D}}}( W_n\odot \operatorname{sign}(L^*))\right) =\left(\mathbf{0}_{p \times p}, \mathbf{0}_{p \times p}\right) .
		\end{aligned}
	\end{equation}
	Notice that the projection $\mathcal{P}_{\widehat{\mathcal{D}}}$ is uniquely determined by the matrix $\widehat{U}_{1, \mathcal{M}_2}$. Lemma \ref{L.Lipschitz} states that the mapping $\widehat{U}_{1, \mathcal{M}_2} \rightarrow \mathcal{P}_{\widehat{\mathcal{D}}}$ is Lipschitz.
	Under Assumption \ref{A.transver}, similar to  Lemma \ref{L.FInvertible}, we have that $\tilde{\mathbf{G}}_{\mathcal{D}^*}$ is invertible, where we define $\tilde{\mathbf{G}}_{\mathcal{D}^*}: \Omega^* \times \mathcal{D}^* \rightarrow \Omega^* \times \mathcal{D}^*$,
	$$
	\tilde{\mathbf{G}}_{\mathcal{D}^*}\left(S^{\prime}, L^{\prime}\right)=\left(\mathcal{P}_{\Omega^*}\left\{\mathcal{I}^*\left(S^{\prime}+L^{\prime}\right)\right\}, \mathcal{P}_{\mathcal{D}^*}\left\{\mathcal{I}^*\left(S^{\prime}+L^{\prime}\right)\right\}\right) \text {, for } S^{\prime} \in \Omega^* \text { and } L^{\prime} \in \mathcal{D}^*,
	$$
	and $\mathcal{D}^*=\left\{U_1^* \Sigma_1^{\prime} U_1^{* \top}: \Sigma_1^{\prime}\right.$ is a $r \times r$ diagonal matrix $\}$ ($\mathcal{D}^*- L^* =\mathcal{D}^*$). According to the invertibility of $\tilde{\mathbf{G}}_{\mathcal{D}^*}$, Lemma \ref{L.Lipschitz} and the fact that $\left\|\widehat{U}_{1, \mathcal{M}_2}-U_1^*\right\|_{\infty} \leq \gamma_n^{1-\eta}$, we know that $\tilde{\mathbf{G}}_{\widehat{\mathcal{D}}}$ is also invertible over $\Omega^* \times\left(\widehat{\mathcal{D}}-L^*\right)$ and is Lipschitz in $\widehat{U}_{1, \mathcal{M}_2}$ for sufficiently large $n$. We apply $\tilde{\mathbf{G}}_{\widehat{\mathcal{D}}}^{-1}$ on both sides of (\ref{eq.PluginDconstraintTransform}) and transform it to a fixed point problem,
	$$
	\left(\Delta_S, \Delta_L\right)=\mathbf{C}\left(\Delta_S, \Delta_L\right),
	$$
	where the operator $\mathbf{C}$ is defined by
	\begin{equation}\label{eq.C}
		\begin{aligned}	
		\mathbf{C}\left(\Delta_S, \Delta_L\right)\triangleq	&  -\tilde{\mathbf{G}}_{\widehat{\mathcal{D}}}^{-1}\left(\left(\mathcal{P}_{\Omega^*} \nabla R_n\left(\Delta_S+\Delta_L\right), \mathcal{P}_{\widehat{\mathcal{D}}} \nabla R_n\left(\Delta_S+\Delta_L\right)\right)\right.\\
		& \left.+\left(\gamma_n \operatorname{sign}\left(\mathbf{O}\left(S^*\right)\right), \tau_n\mathcal{P}_{\widehat{\mathcal{D}}}( W_n\odot \operatorname{sign}(L^*))+ \delta_n \widehat{U}_{1, \mathcal{M}_2} \widehat{U}_{1, \mathcal{M}_2}^{\top}\right)\right) .
	\end{aligned}
	\end{equation}
	
	Define the set $\mathcal{B}^*=\mathcal{B}-\left(S^*, L^*\right)=\left\{\left(S-S^*, L-L^*\right):(S, L) \in \mathcal{B}\right\}$. We will show that with a probability converging to $1, \mathbf{C}$ is a contraction mapping over $\mathcal{B}^*$. 
	\par First, according to (\ref{eq.FirstderivativeRn}), $\delta_n = \rho_1\gamma_n$ and the definition of set $\mathcal{B}$,  for sufficiently large $n$, we have
	\begin{equation}
		\begin{aligned}
			&\left\|\left(\mathcal{P}_{\Omega^*} \nabla R_n\left(\Delta_S+\Delta_L\right) + \gamma_n \operatorname{sign}\left(\mathbf{O}\left(S^*\right)\right), \mathcal{P}_{\widehat{\mathcal{D}}} \nabla R_n\left(\Delta_S+\Delta_L\right) + \delta_n \widehat{U}_{1, \mathcal{M}_2} \widehat{U}_{1, \mathcal{M}_2}^{\top}\right)\right\|_{\infty}\\
			\leq & \kappa \left(\|\Delta_S\| + \|\Delta_L\| + \frac{1}{\sqrt{n}} + \gamma_n \right) \leq O_P(\gamma_n),
		\end{aligned}	
	\end{equation}
	with possibly a different $\kappa$. In addtion, according to  $\tau_n = \rho_2\frac{\gamma_n}{r_n^{1-\alpha}}$ and Lemma \ref{L.WeightOrder},
	\begin{equation*}
		\begin{aligned}
			\tau_n	\|\mathcal{P}_{\widehat{\mathcal{D}}}( W_n\odot \operatorname{sign}(L^*))\|_{\infty}& \leq\tau_n \|(\mathcal{P}_{\widehat{\mathcal{D}}}- \mathcal{P}_{\mathcal{D}^*})( W_n\odot \operatorname{sign}(L^*))\|_{\infty}+\tau_n \|\mathcal{P}_{\mathcal{D}^*}( W_n\odot \operatorname{sign}(L^*))\|_{\infty}\\
			&\leq \kappa\left(\tau_n\|\widehat{U}_1 - U_1^*\|_{\infty}\| W_n\odot \operatorname{sign}(L^*)\|_{\infty} + \tau_n\| W_n\odot \operatorname{sign}(L^*)\|_{\infty}\right) \\
			& = o_P((\gamma_n + 1 ) r_n^{1-\alpha}\tau_n) = o_P(\gamma_n).
		\end{aligned}
	\end{equation*}
  Then it is easy to check that with probability converging to $1, \mathbf{C}\left(\Delta_S, \Delta_L\right) \in \mathcal{B}^*$ for all $(S, L) \in \mathcal{B}$, i.e., $\mathbf{C}\left(\mathcal{B}^*\right) \subset \mathcal{B}^*$. Next, according to (\ref{eq.SeconderivativeRn}) and the boundedness of $\tilde{\mathbf{G}}_{\widehat{\mathcal{D}}}^{-1}$, we know that $\mathbf{C}\left(\Delta_S, \Delta_L\right)$ is Lipschitz in $\left(\Delta_S, \Delta_L\right)$ with a probability converging to 1. To see the size of the Lipschitz constant, according to (\ref{eq.SeconderivativeRn}) we know that $\nabla R_n\left(\Delta_S+\Delta_L\right)$ is Lipschitz with respect to $\left(\Delta_S, \Delta_L\right)$ with the Lipschitz constant of order $O_P\left(\gamma_n^{1-\eta}\right)$. Therefore, the Lipschitz constant for $\mathbf{C}$ is also of order $O_P\left(\gamma_n^{1-\eta}\right)$. Consequently, $\mathbf{C}$ is a contraction mapping over the complete metric space $\mathcal{B}^*$ with a probability converging to 1. According to the Banach fixed point theorem \citep{debnath2005introduction}, (\ref{eq.objconstraintD}) has a unique solution in $\mathcal{B}^*$ with a probability converging to 1, which concludes our proof.
\end{proof}

\subsubsection{Proof of Lemma \ref{L.Lipschitz}, Lemma \ref{L.SLseparate} and Lemma \ref{L.DeltaLError}}
The proof is similar to Lemma 8, Lemma 3 and Lemma 4 in \cite{chen2016fused} and thus ommitted here.

\subsubsection{Proof of Lemma \ref{L.SolutionError}}
\begin{proof}
	Assume that on the contrary, (\ref{eq.objconstriantM2}) has two solutions $\left(\widehat{S}_{\mathcal{M}_2}, \widehat{L}_{\mathcal{M}_2}\right)$ and $\left(\tilde{S}_{\mathcal{M}_2}, \tilde{L}_{\mathcal{M}_2}\right)$. Similar to $\left(\widehat{S}_{\mathcal{M}_2}, \widehat{L}_{\mathcal{M}_2}\right),\left(\tilde{S}_{\mathcal{M}_2}, \tilde{L}_{\mathcal{M}_2}\right)$ also satisfy (\ref{eq.Dconstraint.S}), (\ref{eq.Dconstraint.D})  and (\ref{eq.Dconstraint.U}) if we replace $\left(\tilde{S}_{\mathcal{M}_2}, \tilde{L}_{\mathcal{M}_2}\right)$ by $\left(\widehat{S}_{\mathcal{M}_2}, \widehat{L}_{\mathcal{M}_2}\right)$, and it is also an interior point of $\mathcal{M}_2$. Thus, it satisfies the first order condition of (\ref{eq.objconstriantM2}). That is,
	\begin{equation}\label{eq.PluginOptimalEq}
		\begin{aligned}
			\mathcal{P}_{\Omega^*} \nabla \ell_n\left(\widehat{S}_{\mathcal{M}_2}+\widehat{L}_{\mathcal{M}_2}\right)+\gamma_n \operatorname{sign}\left(\mathbf{O}\left(S^*\right)\right)&=\mathbf{0}_{p \times p} \\
			\mathcal{P}_{\maT(\widehat{L}_{\mathcal{M}_2})} \nabla \ell_n\left(\widehat{S}_{\mathcal{M}_2}+\widehat{L}_{\mathcal{M}_2}\right)+\delta_n \widehat{U}_{1, \mathcal{M}_2} \widehat{U}_{1, \mathcal{M}_2}^{\top} + \tau_nW_n\odot\operatorname{sign}(L^*)&=\mathbf{0}_{p \times p} .
		\end{aligned}
	\end{equation}
	We define an operator $\mathbf{G}_L: \Omega^* \times \maT(L) \rightarrow \Omega^* \times \maT(L)$ in a similar way as that of $\mathbf{G}$,
	$$
	\mathbf{G}_L\left(S, L^{\prime}\right)=\left(\mathcal{P}_{\Omega^*}\left\{\mathcal{I}^*\left(S+L^{\prime}\right)\right\}, \mathcal{P}_{\maT(L)}\left\{\mathcal{I}^*\left(S+L^{\prime}\right)\right\}\right) .
	$$
	With similar arguments as that in Step 1, we know that $\mathbf{G}_{\widehat{L}_{\mathcal{M}_2}}$ is invertible with the aid of Lemmas \ref{L.Lipschitz} and Lemma \ref{L.FInvertible}, and (\ref{eq.PluginOptimalEq}) is transformed to
	\begin{equation}\label{eq.TransFhat}
		\begin{aligned}
			\left(\Delta_{\widehat{S}_{\mathcal{M}_2}}, \Delta_{\widehat{L}_{\mathcal{M}_2}}\right)=\mathbf{G}_{\widehat{L}_{\mathcal{M}_2}}^{-1}&\left(\left(\mathcal{P}_{\Omega^*} \nabla R_n\left(\Delta_{\widehat{S}_{\mathcal{M}_2}}+\Delta_{\widehat{L}_{\mathcal{M}_2}}\right)\right.\right.\left.\mathcal{P}_{\maT(\widehat{L}_{\mathcal{M}_2})}  \nabla R_n\left(\Delta_{\widehat{S}_{\mathcal{M}_2}}+\Delta_{\widehat{L}_{\mathcal{M}_2}}\right)\right)\\
			& \left.+\left(\gamma_n \operatorname{sign}\left(\mathbf{O}\left(S^*\right)\right), \delta_n \widehat{U}_{1, \mathcal{M}_2} \widehat{U}_{1, \mathcal{M}_2}^{\top} + \tau_nW_n\odot\operatorname{sign}(L^*)\right)\right), 
		\end{aligned}
	\end{equation}
	where $\Delta_{\widehat{S}_{\mathcal{M}_2}}=\widehat{S}_{\mathcal{M}_2}-S^*$ and $\Delta_{\widehat{L}_{\mathcal{M}_2}}=\widehat{L}_{\mathcal{M}_2}-L^*$. Similarly for $\left(\tilde{S}_{\mathcal{M}_2}, \tilde{L}_{\mathcal{M}_2}\right)$, we have
	\begin{equation}\label{eq.TransFtilde}
		\begin{aligned}
		\left(\Delta_{\tilde{S}_{\mathcal{M}_2}}, \Delta_{\tilde{L}_{\mathcal{M}_2}}\right)=\mathbf{G}_{\tilde{L}_{\mathcal{M}_2}}^{-1}&\left(\left(\mathcal{P}_{\Omega^*} \nabla R_n\left(\Delta_{\tilde{S}_{\mathcal{M}_2}}+\Delta_{\tilde{L}_{\mathcal{M}_2}}\right)\right.\right.\left.\mathcal{P}_{\maT(\tilde{L}_{\mathcal{M}_2})}  \nabla R_n\left(\Delta_{\tilde{S}_{\mathcal{M}_2}}+\Delta_{\tilde{L}_{\mathcal{M}_2}}\right)\right)\\
		& \left.+\left(\gamma_n \operatorname{sign}\left(\mathbf{O}\left(S^*\right)\right), \delta_n \widehat{U}_{1, \mathcal{M}_2}  \widehat{U}_{1, \mathcal{M}_2} ^{\top} + \tau_nW_n\odot\operatorname{sign}(L^*)\right)\right) .
	\end{aligned}	
	\end{equation}
	Similar to the definition (\ref{eq.C}), for $(S, L) \in \mathcal{M}_2$ we define
	$$
	\begin{aligned}
	 \mathbf{C}_{(S, L)}\left(\Delta_S, \Delta_L\right) 
	   =&-\mathbf{G}_L^{-1}\left(\left(\mathcal{P}_{\Omega^*} \nabla R_n\left(\Delta_S+\Delta_L\right), \mathcal{P}_{\maT(L)} \nabla R_n\left(\Delta_S+\Delta_L\right)\right)\right.\\ 
		& \left.+\left(\gamma_n \operatorname{sign}\left(\mathbf{O}\left(S^*\right)\right), \delta_n U_1 U_1^{\top} + \tau_nW_n\odot\operatorname{sign}(L^*)\right),\right.
	\end{aligned}
	$$
	where $L$ has the eigendecomposition $L=U_1 \Sigma_1 U_1^{\top},\left\|U_1-U_1^*\right\|_{\infty} \leq \gamma_n^{1-\eta}$ and $\left\|\Sigma_1-\Sigma_1^*\right\|_{\infty} \leq$ $\gamma_n^{1-2 \eta}$. The operator $\mathbf{C}_{(S, L)}$ is well defined, because for $L \in \mathcal{M}_2$ the eigendeposition of $L$ is 	uniquely determined given $\left(U_1, \Sigma_1\right)$ is in the set $\left\{\left(U_1, \Sigma_1\right):\left\|U_1-U_1^*\right\|_{\infty} \leq \gamma_n^{1-\eta}\right.$ and $\| \Sigma_1-$ $\left.\Sigma_1^* \|_{\infty} \leq \gamma_n^{1-2 \eta}\right\}$. Now, we take difference between (\ref{eq.TransFhat}) and (\ref{eq.TransFtilde}),
	\begin{equation}\label{eq.DiffUnique}		\left(\tilde{S}_{\mathcal{M}_2}-\widehat{S}_{\mathcal{M}_2}, \tilde{L}_{\mathcal{M}_2}-\widehat{L}_{\mathcal{M}_2}\right)=\mathbf{C}_{\tilde{S}_{\mathcal{M}_2}, \tilde{L}_{\mathcal{M}_2}}\left(\Delta_{\tilde{S}_{\mathcal{M}_2}}, \Delta_{\tilde{L}_{\mathcal{M}_2}}\right)-\mathbf{C}_{\widehat{S}_{\mathcal{M}_2}, \widehat{L}_{\mathcal{M}_2}}\left(\Delta_{\widehat{S}_{\mathcal{M}_2}}, \Delta_{\widehat{L}_{\mathcal{M}_2}}\right).
	\end{equation}
	We provide an upper bound for the norm of the right-hand side of the above equation. We split the right-hand side of the above display into two terms to get
	$$
	\begin{aligned}
		& \left\|\mathbf{C}_{(\tilde{S}_{\mathcal{M}_2}, \tilde{L}_{\mathcal{M}_2})}\left(\Delta_{\tilde{S}_{\mathcal{M}_2}}, \Delta_{\tilde{L}_{\mathcal{M}_2}}\right)-\mathbf{C}_{(\widehat{S}_{\mathcal{M}_2}, \widehat{L}_{\mathcal{M}_2})}\left(\Delta_{\widehat{S}_{\mathcal{M}_2}}, \Delta_{\widehat{L}_{\mathcal{M}_2}}\right)\right\|_{\infty} \\
		\leq& \left\|\mathbf{C}_{(\tilde{S}_{\mathcal{M}_2}, \tilde{L}_{\mathcal{M}_2})}\left(\Delta_{\tilde{S}_{\mathcal{M}_2}}, \Delta_{\tilde{L}_{\mathcal{M}_2}}\right)-\mathbf{C}_{(\tilde{S}_{\mathcal{M}_2}, \tilde{L}_{\mathcal{M}_2})}\left(\Delta_{\widehat{S}_{\mathcal{M}_2}}, \Delta_{\widehat{L}_{\mathcal{M}_2}}\right)\right\|_{\infty} \\
		 &+\left\|\mathbf{C}_{(\tilde{S}_{\mathcal{M}_2}, \tilde{L}_{\mathcal{M}_2})}\left(\Delta_{\widehat{S}_{\mathcal{M}_2}}, \Delta_{\widehat{L}_{\mathcal{M}_2}}\right)-\mathbf{C}_{(\widehat{S}_{\mathcal{M}_2}, \widehat{L}_{\mathcal{M}_2})}\left(\Delta_{\widehat{S}_{\mathcal{M}_2}}, \Delta_{\widehat{L}_{\mathcal{M}_2}}\right)\right\|_{\infty} \\
		\triangleq & \uppercase\expandafter{\romannumeral1} + \uppercase\expandafter{\romannumeral2}.
	\end{aligned}
	$$
	We present upper bounds for $\uppercase\expandafter{\romannumeral1}$ and $\uppercase\expandafter{\romannumeral2}$ separately. For $\uppercase\expandafter{\romannumeral1}$, using similar arguments as those in the proof of Lemma \ref{L.Dconstraint}, we have that with probability converging to $1, \mathbf{C}_{\tilde{S}_{\mathcal{M}_2}, \tilde{L}_{\mathcal{M}_2}}(\cdot, \cdot)$ is a Lipschitz operator with an $O\left(\gamma_n^{1-\eta}\right)$ Lipschitz constant, that is,
	$$\uppercase\expandafter{\romannumeral1} \leq_P \kappa \gamma_n^{1-\eta} \times \max \left(\left\|\widehat{S}_{\mathcal{M}_2}-\tilde{S}_{\mathcal{M}_2}\right\|_{\infty},\left\|\widehat{L}_{\mathcal{M}_2}-\tilde{L}_{\mathcal{M}_2}\right\|_{\infty}\right)$$
	We proceed to an upper bound of $\uppercase\expandafter{\romannumeral2}$. Thanks to the Lipschitz property of $\mathbf{G}_L$ and $\mathcal{P}_{\maT(L)}$ and the invertibility of $\mathbf{G}_{L^*}$, with a probability converging to $1, \mathbf{C}_{(S, L)}\left(\Delta_{\widehat{S}_{\mathcal{M}_2}}, \Delta_{\widehat{L}_{\mathcal{M}_2}}\right)$ is Lipschitz in $(S, L)$ when $\left(\Delta_{\widehat{S}_{\mathcal{M}_2}}, \Delta_{\widehat{L}_{\mathcal{M}_2}}\right)$ is fixed, i.e., we used the following factorizing
	\[
	\mathbf{C}_{(S_1,L_1)}(\cdot,\cdot) - \mathbf{C}_{(S_2, L_2)}(\cdot, \cdot) = \mathbf{G}_{L_2}^{-1}(\cdot,\cdot) -\mathbf{G}_{L_1}^{-1}(\cdot,\cdot) = \left(\mathbf{G}_{L_1}^{-1}\left( \mathbf{G}_{L_1} -\mathbf{G}_{L_2}\right)\mathbf{G}_{L_2}^{-1}\right)(\cdot, \cdot).
	\]
	Moreover, according to (\ref{eq.FirstderivativeRn}), $\left\|\Delta_{\widehat{S}_{\mathcal{M}_2}}\right\|_{\infty} \leq \gamma_n^{1-2 \eta}$ and $\left\|\Delta_{\widehat{L}_{\mathcal{M}_2}}\right\|_{\infty} \leq \gamma_n^{1-\eta}$, we have
	\begin{equation}\label{eq.GradientHat}
	     \left\|\nabla R_n\left(\Delta_{\widehat{S}_{\mathcal{M}_2}}+\Delta_{\widehat{L}_{\mathcal{M}_2}}\right)\right\|_{\infty} \leq O_P\left(\frac{1}{\sqrt{n}}\right)\leq o_P\left(\gamma_n\right)  .
	\end{equation}
	Combining the above inequality with the fact that $\mathbf{G}_L$ and $\mathcal{P}_{\maT(L)}$ are locally Lipschitz in $L$ according to Theorem 2 in \cite{Yu2014} under Assumption \ref{A.non-overlapped}, we have that
	$$
	\uppercase\expandafter{\romannumeral2} \leq_P \kappa \gamma_n \times \max \left(\left\|\widehat{S}_{\mathcal{M}_2}-\tilde{S}_{\mathcal{M}_2}\right\|_{\infty},\left\|\widehat{L}_{\mathcal{M}_2}-\tilde{L}_{\mathcal{M}_2}\right\|_{\infty}\right) .
	$$
	We combine the upper bounds for $\uppercase\expandafter{\romannumeral1}$ and $\uppercase\expandafter{\romannumeral2}$ with the equation (\ref{eq.DiffUnique}) to get
	$$
	\max \left(\left\|\widehat{S}_{\mathcal{M}_2}-\tilde{S}_{\mathcal{M}_2}\right\|_{\infty},\left\|\widehat{L}_{\mathcal{M}_2}-\tilde{L}_{\mathcal{M}_2}\right\|_{\infty}\right) \leq_P 2 \kappa \gamma_n^{1-\eta} \times \max \left(\left\|\widehat{S}_{\mathcal{M}_2}-\tilde{S}_{\mathcal{M}_2}\right\|_{\infty},\left\|\widehat{L}_{\mathcal{M}_2}-\tilde{L}_{\mathcal{M}_2}\right\|_{\infty}\right)
	$$
	Consequently, $\widehat{S}_{\mathcal{M}_2}={ }_P \tilde{S}_{\mathcal{M}_2}$ and $\widehat{L}_{\mathcal{M}_2}={ }_P \tilde{L}_{\mathcal{M}_2}$. We proceed to prove the equality in Lemma \ref{L.SolutionError}. According to (\ref{eq.TransFhat}) and (\ref{eq.GradientHat}), we have
	$$
	\left(\widehat{S}_{\mathcal{M}_2}-S^*, \widehat{L}_{\mathcal{M}_2}-L^*\right)=\gamma_n \mathbf{G}_{\widehat{L}_{\mathcal{M}_2}}^{-1}\left(\mathbf{q}_{\widehat{U}_{1, \mathcal{M}_2}}\right)+o_P\left(\gamma_n\right),
	$$
	where $\mathbf{q}_{\widehat{U}_{1, \mathcal{M}_2}}=\left(\operatorname{sign}\left(\mathbf{O}\left(S^*\right)\right), \rho_1 \widehat{U}_{1, \mathcal{M}_2} \widehat{U}_{1, \mathcal{M}_2}^{\top}+ \frac{\rho_2}{r_n^{1-\alpha}}W_n\odot\operatorname{sign}(L^*)\right)$. Because both $\mathbf{G}_{\widehat{L}_{\mathcal{M}_2}}$ and $\mathbf{q}_{\widehat{U}_{1, \mathcal{M}_2}}$ are Lipschitz continuous in $\widehat{U}_{1, \mathcal{M}_2}$,  $\left\|\widehat{U}_{1, \mathcal{M}_2}-U_1^*\right\| \leq \gamma_n^{1-\eta}$, and $ \frac{\rho_2}{r_n^{1-\alpha}}\|W_n\odot \operatorname{sign}(L^*)\|_{\infty} = o_P(1) $ (Lemma \ref{L.WeightOrder}) and note that 
	$$
	\begin{aligned}
		\mathbf{G}^{-1}_{\widehat{L}_{\mathcal{M}_2}}(\mathbf{q}_{\widehat{U}_{1, \mathcal{M}_2}}) &= (\mathbf{G}^{-1}_{\widehat{L}_{\mathcal{M}_2}} -\mathbf{G}^{-1}_{L^*})(\mathbf{q}_{\widehat{U}_{1, \mathcal{M}_2}}) + \mathbf{G}^{-1}_{L^*}(\mathbf{q}_{\widehat{U}_{1, \mathcal{M}_2}} - \mathbf{q}_{U_1^*}) + \mathbf{G}^{-1}_{L^*}(\mathbf{q}_{U_1^*})\\
		&= \left(\mathbf{G}^{-1}_{L^*}\left(\mathbf{G}_{L^*}-\mathbf{G}_{\widehat{L}_{\mathcal{M}_2}}\right) \mathbf{G}^{-1}_{\widehat{L}_{\mathcal{M}_2}}\right)(\mathbf{q}_{\widehat{U}_{1, \mathcal{M}_2}}) + \mathbf{G}^{-1}_{L^*}(\mathbf{q}_{\widehat{U}_{1, \mathcal{M}_2}} - \mathbf{q}_{U_1^*}) + \mathbf{G}^{-1}_{L^*}(\mathbf{q}_{U_1^*}),
	\end{aligned}  
	$$
	we have
	$$
	\left(\widehat{S}_{\mathcal{M}_2}-S^*, \widehat{L}_{\mathcal{M}_2}-L^*\right)=\gamma_n \mathbf{G}_{L^*}^{-1} (\mathbf{q}_{U_1^*})+o_P\left(\gamma_n\right),
	$$
	where $\mathbf{q}_{\widehat{U}^*_{1}}=\left(\operatorname{sign}\left(\mathbf{O}\left(S^*\right)\right), \rho_1 U_1^*U_1^{*T}\right)$.
\end{proof}

\subsubsection{Proof of Lemma \ref{L.UniqueSolution}}
\begin{proof}
	For any solution $(\tilde{S}, \tilde{L})\in \mathcal{N}$, from definition of $\mathcal{N}$, we can rewrite 
	$$
	\tilde{S}=\tilde{S}_{\mathcal{M}_2}+\tilde{S}_{\Omega^{* \perp}},\quad \tilde{L}=\tilde{L}_{\mathcal{M}_2}+	\tilde{L}_{\maT^\perp(\tilde{L}_{\mathcal{M}_2})},
	$$ 
	where
	$$
	\tilde{S}_{\mathcal{M}_2} \in \Omega^*,\quad \tilde{S}_{\Omega^{* \perp}} \in \Omega^{* \perp},\quad \tilde{L}_{\mathcal{M}_2}=\tilde{U}_{1, \mathcal{M}_2} \tilde{\Sigma}_{1, \mathcal{M}_2} \tilde{U}_{1, \mathcal{M}_2}^{T} \in \maT(\tilde{L}_{\mathcal{M}_2}), \quad \tilde{L}_{\maT^\perp(\tilde{L}_{\mathcal{M}_2})}   \in \maT^\perp(\tilde{L}_{\mathcal{M}_2}),
	$$
and
	$$  	
	\|\tilde{S}_{\mathcal{M}_2}\|_{\infty} \leq O(d_n) ,\quad	\|\tilde{L}_{\maT^\perp(\tilde{L}_{\mathcal{M}_2})}\|_{\infty} \leq O(d_n).
	$$
	Notice that $\left(\tilde{S}_{\mathcal{M}_2}, \tilde{L}_{\mathcal{M}_2}\right) \in \mathcal{M}_2$, we have the sub-differentials of $H(S, L)$ at $\left(\tilde{S}_{\mathcal{M}_2}, \tilde{L}_{\mathcal{M}_2}\right)$,
	$$
	\left.\partial_S H\right|_{\left(\tilde{S}_{\mathcal{M}_2}, \tilde{L}_{\mathcal{M}_2}\right)}=\left\{\nabla \ell_n\left(\tilde{S}_{\mathcal{M}_2}+\tilde{L}_{\mathcal{M}_2}\right)+\gamma_n \operatorname{sign}\left(\mathbf{O}\left(S^*\right)\right)+\gamma_n M_1:\left\|M_1\right\|_{\infty} \leq 1 \text {, and } M_1 \in \Omega^{* \perp}\right\}
	$$
	and
	$$
	\begin{aligned}
		\left.\partial_L H\right|_{\left(\tilde{S}_{\mathcal{M}_2}, \tilde{L}_{\mathcal{M}_2}\right)} 	=&\left\{\nabla \ell_n\left(\tilde{S}_{\mathcal{M}_2}+\tilde{L}_{\mathcal{M}_2}\right)+\delta_n \tilde{U}_{1, \mathcal{M}_2} \tilde{U}_{1, \mathcal{M}_2}^{\top}+ \tau_n W_n\odot \operatorname{sign}(L^*) \right.\\
		&\left.	+ \delta_nM_2 +\tau_n W_n\odot M_3:\|M_2\|_2\leq 1, M_2\in \maT_1^{\perp}(\tilde{L}_{\mathcal{M}_2}), \|M_3\|_{\infty}\leq 1, M_3\in \maT_2^{*\perp}\right\}
	\end{aligned}	$$
	According to the definition of sub-differential, we have
	$$
	\begin{aligned}
		& H(\tilde{S}, \tilde{L})-H\left(\tilde{S}_{\mathcal{M}_2}, \tilde{L}_{\mathcal{M}_2}\right) \\
		\geq & \nabla \ell_n\left(\tilde{S}_{\mathcal{M}_2}+\tilde{L}_{\mathcal{M}_2}\right) \cdot\left(\tilde{S}_{\Omega^{* \perp}}+\tilde{L}_{\maT^\perp(\tilde{L}_{\mathcal{M}_2})}\right)  + \left(\gamma_n \operatorname{sign}\left(\mathbf{O}\left(S^*\right)\right)+\gamma_n M_1\right)\cdot\tilde{S}_{\Omega^{* \perp}}   \\
		&  + \tilde{L}_{\maT^\perp(\tilde{L}_{\mathcal{M}_2})}\cdot\left( \delta_n \widehat{U}_{1, \mathcal{M}_2} \widehat{U}_{1, \mathcal{M}_2}^{\top}+ \tau_n W_n\odot \operatorname{sign}(L^*) 	+ \delta_n M_2 +\tau_n W_n\odot M_3 \right) \\
	\end{aligned}
	$$
	Because $\tilde{S}_{\Omega^{* \perp}} \in \Omega^{* \perp}$ and $\tilde{L}_{\maT^\perp(\tilde{L}_{\mathcal{M}_2})}\in \maT^\perp(\tilde{L}_{\mathcal{M}_2})$, we further expand the above inequality,
	\begin{equation}\label{eq.DifferH1}
		\begin{aligned}
			& H(\tilde{S}, \tilde{L})-H\left(\tilde{S}_{\mathcal{M}_2}, \tilde{L}_{\mathcal{M}_2}\right) \\
			\ge	& \mathcal{P}_{\Omega^{* \perp}}\left[\nabla \ell_n\left(\tilde{S}_{\mathcal{M}_2}+\tilde{L}_{\mathcal{M}_2}\right)\right] \cdot \tilde{S}_{\Omega^{* \perp}}+\mathcal{P}_{\maT^\perp(\tilde{L}_{\mathcal{M}_2})}\left[\nabla \ell_n\left(\tilde{S}_{\mathcal{M}_2}+\tilde{L}_{\maT^\perp(\tilde{L}_{\mathcal{M}_2})}\right)\right] \cdot \tilde{L}_{\maT^\perp(\tilde{L}_{\mathcal{M}_2})} \\
			&   + \left(\gamma_n \operatorname{sign}\left(\mathbf{O}\left(S^*\right)\right)+\gamma_n M_1\right)\cdot\tilde{S}_{\Omega^{* \perp}} +\tilde{L}_{\maT^\perp(\tilde{L}_{\mathcal{M}_2})}\cdot\left( \delta_n \widehat{U}_{1, \mathcal{M}_2}  \widehat{U}_{1, \mathcal{M}_2} ^{\top}+ \tau_n W_n\odot \operatorname{sign}(L^*) 	+ \delta_n M_2 +\tau_n W_n\odot M_3 \right)\\
			\ge	& \mathcal{P}_{\Omega^{* \perp}}\left[\nabla \ell_n\left(\tilde{S}_{\mathcal{M}_2}+\tilde{L}_{\mathcal{M}_2}\right)\right] \cdot \tilde{S}_{\Omega^{* \perp}} +  \mathcal{P}_{\maT^\perp(\tilde{L}_{\mathcal{M}_2})} \left[\nabla \ell_n\left(\tilde{S}_{\mathcal{M}_2}+\tilde{L}_{\mathcal{M}_2}\right)\right]\cdot \tilde{L}_{\maT^\perp(\tilde{L}_{\mathcal{M}_2})}   \\
			&  + \left(\gamma_n \operatorname{sign}\left(\mathbf{O}\left(S^*\right)\right)+\gamma_n M_1\right)\cdot\tilde{S}_{\Omega^{* \perp}}  +\tilde{L}_{\maT^\perp(\tilde{L}_{\mathcal{M}_2})}\cdot\left( \delta_n \widehat{U}_{1, \mathcal{M}_2}  \widehat{U}_{1, \mathcal{M}_2} ^{\top}+ \tau_n W_n\odot \operatorname{sign}(L^*) 	+ \delta_n M_2 +\tau_n W_n\odot M_3 \right) .\\
		\end{aligned}
	\end{equation}
    Denote 
	\[
	N_1 \triangleq \mathcal{P}_{\Omega^{* \perp}} \mathcal{I}^* \mathcal{A} \mathbf{G}^{-1}\left(\operatorname{sign}\left(\mathbf{O}\left(S^*\right)\right), \rho_1 U_1^* U_1^{* \top}\right)
	\]
	\[
	N_2\triangleq \mathcal{P}_{\maT^{*\perp}} \mathcal{I}^* \mathcal{A} \mathbf{G}^{-1}\left(\operatorname{sign}\left(\mathbf{O}\left(S^*\right)\right), \rho_1 U_1^* U_1^{* \top}\right),
	\]
	where we can decompose $N_2 = P \oplus Q\in \maT^{*\perp}  = \maV^{*}\oplus \maT_2^{*\perp}, P\in \maV^{*}, Q\in \maT_2^{*\perp} $.
Then, plug them into (\ref{eq.DifferH1}) we have 
	\begin{equation}\label{eq.DifferH2}
		\begin{aligned}
			& H(\tilde{S}, \tilde{L})-H\left(\tilde{S}_{\mathcal{M}_2}, \tilde{L}_{\mathcal{M}_2}\right) \\
			\ge	& \left(  \gamma_n N_1+o_P\left(\gamma_n\right) \right)\cdot \tilde{S}_{\Omega^{* \perp}} + \left(\gamma_n (P +Q) + o_P(\gamma_n)  \right)\cdot \tilde{L}_{\maT^\perp(\tilde{L}_{\mathcal{M}_2})} +\left(\gamma_n \operatorname{sign}\left(\mathbf{O}\left(S^*\right)\right)+\gamma_n M_1\right)\cdot \tilde{S}_{\Omega^{* \perp}} \\
			& + \tilde{L}_{\maT^\perp(\tilde{L}_{\mathcal{M}_2})}\cdot\left( \delta_n \tilde{U}_{1, \mathcal{M}_2}  \tilde{U}_{1, \mathcal{M}_2} ^{\top}+ \tau_n W_n\odot \operatorname{sign}(L^*) 	+ \delta_n M_2 +\tau_n W_n\odot M_3 \right) \\
			= & \gamma_n N_1\cdot \tilde{S}_{\Omega^{* \perp}} + \gamma_n (P +Q)  \cdot  \tilde{L}_{\maT^\perp(\tilde{L}_{\mathcal{M}_2})} \\
			&+ \gamma_n M_1\cdot \tilde{S}_{\Omega^{* \perp}} +  \tilde{L}_{\maT^\perp(\tilde{L}_{\mathcal{M}_2})}\cdot\left(\delta_n \tilde{U}_{1, \mathcal{M}_2}  \tilde{U}_{1, \mathcal{M}_2} ^{\top}+ \tau_n W_n\odot \operatorname{sign}(L^*) 	+ \delta_n M_2 +\tau_n W_n\odot M_3 \right)+ o_P(\gamma_nD_n).
		\end{aligned}
	\end{equation}
	where  we use the fact that $\operatorname{sign}\left(\mathbf{O}\left(S^*\right)\right)\cdot \tilde{S}_{\Omega^{* \perp}} = 0$ and define $D_n = \max\{\|\tilde{S}_{\Omega^{* \perp}}\|_{\infty}, \|\tilde{L}_{\maT^\perp(\tilde{L}_{\mathcal{M}_2})}\|_{\infty} \}$. 
	\par Denote $A = \tilde{S}_{\Omega^{* \perp}}, B= \tilde{L}_{\maT^\perp(\tilde{L}_{\mathcal{M}_2})}$, and
	 let  $\mathcal{P}_\maV B =\tilde{U}\tilde{D}\tilde{U}^T $, where $\tilde{U}$ is the orthogonal basis in linear  space $V$ with $\maT^\perp(\tilde{L}_{\mathcal{M}_2}) = \maV \oplus \maT_2^{*\perp}$, $V$ is a subspace of $\maT_1^{\perp}(\tilde{L}_{\mathcal{M}_2})$.	Take $M_1 = \operatorname{sign}(\tilde{S}_{\Omega^{* \perp}})$, $M_2 =\tilde{U}\operatorname{sign}(\tilde{D})\tilde{U}^T $, $M_3 = \operatorname{sign}(\mathcal{P}_{\maT_2^{*\perp}}B)$. Denote $  C = \delta_n \tilde{U}_{1, \mathcal{M}_2}  \tilde{U}_{1, \mathcal{M}_2} ^{\top}, D = \tau_n W_n\odot \operatorname{sign}(L^*), E= \delta_n M_2, F=\tau_n W_n\odot M_3  $, and  note that 
	$$
	P\cdot B = P\cdot (\mathcal{P}_{\maV^*}B),\quad Q\cdot B = Q\cdot(\mathcal{P}_{\maT_2^{*\perp}}B).
	$$
   Now we compute that
	\[
  M_1\cdot A = \|A\|_1, \quad B\cdot C = 0,\quad B\cdot D = 0,\quad B\cdot E = \delta_n \| \mathcal{P}_\maV B\|_*,\quad B\cdot F = \tau_n \|W_n\odot \mathcal{P}_{\maT_2^{*\perp}}B\|_1.
	\]
	On the other hand, according to Assumption \ref{A.Wn}, Assumption 
	\ref{A.irrepresentable}, the above expression and dual inequality, we have
	$$
	\begin{aligned}
		& \gamma_n\left|N_1\cdot A\right| \leq \gamma_n\|N_1\|_{\infty}\left\|A\right\|_{1} < \gamma_n\|A\|_1, \\
		& \gamma_n\left|P\cdot B\right| = \gamma_n\left|P\cdot \mathcal{P}_{\maV^*}B\right|\leq \gamma_n \|P\|_2\|\mathcal{P}_{\maV^*}B\|_*< \gamma_n\rho_1\|\mathcal{P}_{\maV^*}B\|_*=\delta_n\left\|\mathcal{P}_{\maV^*}B\right\|_* ,
	\end{aligned}
	$$
 and
	$$
	\begin{aligned}
		\gamma_n|Q\cdot B|& = \gamma_n |(\widetilde{W}_n\odot Q)\cdot(W_n\odot \mathcal{P}_{\maT_2^{*\perp}}B)|\\
		&\leq \gamma_n\|\widetilde{W}_n\odot Q\|_\infty \|W_n\odot \mathcal{P}_{\maT_2^{*\perp}}B\|_1\\
		&\leq \gamma_n\left(\|(\widetilde{W}_n - \Lambda_n )\odot Q\|_\infty + \|\Lambda_n\odot Q\|_\infty\right)\|W_n\odot \mathcal{P}_{\maT_2^{*\perp}}B\|_1\\
		&< \gamma_n\left( O_P(\frac{1}{r_n})  + \frac{\rho_2}{r_n^{1-\alpha}} \right)\|W_n\odot \mathcal{P}_{\maT_2^{*\perp}}B\|_1\\
		& < (\tau_n + o_P(\tau_n))\|W_n\odot\mathcal{P}_{\maT_2^{*\perp}}B\|_1  \\
		&<_P \tau_n\|W_n\odot \mathcal{P}_{\maT_2^{*\perp}}B\|_1.
	\end{aligned}
	$$
    i.e., there exists $0<c_1,c_2,c_3<1$ such that $  \gamma_n\left|N_1\cdot A\right|\leq c_1\gamma_n\left\| A\right\|_1, \gamma_n\left|P\cdot B\right| \leq c_2 \delta_n\left\|\mathcal{P}_{\maV^*}B\right\|_*, 	\gamma_n|Q\cdot B|\leq c_3\tau_n\|W_n\odot \mathcal{P}_{\maT_2^{*\perp}}B\|_1$. Therefore, combining above inequalities with  (\ref{eq.DifferH2}) we get 
	$$
	\begin{aligned}
		H(\tilde{S}, \tilde{L})- H\left(\tilde{S}_{\mathcal{M}_2}, \tilde{L}_{\mathcal{M}_2}\right)\ge_P &  \gamma_n(1-c_1)\|A\|_1 + \delta_n(\left\|\mathcal{P}_{V}B\right\|_*-c_2\left\|\mathcal{P}_{\maV^*}B\right\|_*)\\
		& + (1 - c_3)\tau_n\|W_n\odot \mathcal{P}_{\maT_2^{*\perp}}B\|_1 + o_P(D_n\gamma_n)\\
		\ge_P &\gamma_n(1-c_1)\|A\|_1  +  (1 - c_3)\tau_n\|W_n\odot \mathcal{P}_{\maT_2^{*\perp}}B\|_1\\
		&
		+\delta_n\left((1 -c_2)\left\|\mathcal{P}_{\maV^*}B\right\|_*-c_2\left\|(\mathcal{P}_{V} -\mathcal{P}_{\maV^*} )B\right\|_*\right) +  o_P(D_n\gamma_n)\\
		\ge_P&  \gamma_n(1-c_1)\|A\|_1 +\delta_n(1 -c_2)\left\|\mathcal{P}_{\maV^*}B\right\|_* +(1 - c_3)\tau_n\|W_n\odot \mathcal{P}_{\maT_2^{*\perp}}B\|_1 + o_P(D_n\gamma_n)\\
		>_P& 0,	
	\end{aligned}
	$$
	provided $\left\|A\right\|_{1}>0$ or $\left\|\mathcal{P}_{\maV^*}B\right\|_*>0$ or $\|W_n\odot \mathcal{P}_{\maT_2^{*\perp}}B\|_1 >0$, where the third inequality is according to Lemma \ref{L.DirectDecomp}, 
	$$ \delta_n\left\|(\mathcal{P}_{\maV} -\mathcal{P}_{\maV^*} )B\right\|_*\lesssim  \kappa\gamma_n\|\tilde{L}_{\mathcal{M}_2} - L^*\|\|B\| = o_P(D_n\gamma_n) $$
	with  possibly different $\kappa$.
	Moreover, according to Lemma \ref{L.SolutionError}
	$$
	H(\tilde{S}, \tilde{L})>_P H\left(\tilde{S}_{\mathcal{M}_2}, \tilde{L}_{\mathcal{M}_2}\right)>_PH\left(\widehat{S}_{\mathcal{M}_2}, \widehat{L}_{\mathcal{M}_2}\right)
	$$
 provided $\left\|A\right\|_{1}>0$ or $\left\|\mathcal{P}_{\maV^*}B\right\|_*>0$ or $\|W_n\odot \mathcal{P}_{\maT_2^{*\perp}}B\|_1 >0$. Since $(\tilde{S}, \tilde{L})$ is a solution to (\ref{eq.objconstraintM1}), the above statement implies $A=B = \boldsymbol{0}_{p \times p}$, i.e.,  $\tilde{S}_{\Omega^{* \perp}}=\tilde{L}_{\maT^\perp(\tilde{L}_{\mathcal{M}_2})}=\mathbf{0}_{p \times p}$. Therefore, $\tilde{S}=_P\widehat{S}_{\mathcal{M}_2}, \tilde{L}=_P\widehat{L}_{\mathcal{M}_2}$, and $(\tilde{S}, \tilde{L}) \in \mathcal{M}_2$, which completes the proof.
\end{proof}

\end{spacing}
\end{document}